%% file: main.tex
\documentclass[10pt,twocolumn,letterpaper]{article}

\usepackage{iccv}              % To produce the CAMERA-READY version
\input{preamble}

\def\paperID{11343} % *** Enter the Paper ID here
\def\confName{ICCV}
\def\confYear{2025}

\title{Adversarial Exploitation of Data Diversity Improves Visual Localization}
\author{
  Sihang Li\textsuperscript{*} \quad 
  Siqi Tan\textsuperscript{*} \quad 
  Bowen Chang \quad 
  Jing Zhang \quad 
  Chen Feng\textsuperscript{\ding{41}} \quad 
  Yiming Li\textsuperscript{\ding{41}} \vspace{5pt}\\
  New York University \\
  \tt{\small \{sihangli, tan.k, cfeng, yimingli\}@nyu.edu} \\
  {\small \url{https://ai4ce.github.io/RAP}}
}

\begin{document}

\maketitle

\renewcommand{\thefootnote}{\fnsymbol{footnote}} 
\setcounter{footnote}{0} 
\footnotetext{\textsuperscript{*}Equal contribution. \quad \textsuperscript{\ding{41}}Corresponding author.}

\input{sec/0_abstract}    
\input{sec/1_intro}

\input{sec/2_Related_Works}
\input{sec/3_Pre_Processing}

\input{sec/4_Training}

\input{sec/5_Experiment}
\input{sec/6_Conclusion}

\noindent\textbf{Acknowledgment.}
This work was supported in part through NSF grants 2238968, 2121391, and 2024882, and the NYU IT High Performance Computing resources, services, and staff expertise. Yiming Li is supported by NVIDIA Graduate Fellowship (2024-2025).

\clearpage
\clearpage

{
    \small
    \bibliographystyle{ieeenat_fullname}
    \bibliography{main}
}

\input{sec/X_suppl}

\end{document}

%% file: preamble.tex
\usepackage[dvipsnames]{xcolor}
\usepackage{amsmath}
\usepackage{array}
\usepackage{color}
\usepackage{epsfig}
\usepackage{graphicx}
\usepackage{textcomp}
\usepackage{gensymb}
\usepackage{booktabs}
\usepackage{makecell}
\usepackage{enumitem}
\usepackage[ruled,vlined]{algorithm2e}
\usepackage[normalem]{ulem}
\usepackage{multirow}
\usepackage{wrapfig}
\usepackage{xspace}
\usepackage[accsupp]{axessibility} 
\usepackage{booktabs} 
\usepackage{soul} 
\usepackage{xcolor} 
\usepackage{colortbl}
\sethlcolor{orange!30}
\usepackage{graphicx} 
\usepackage{multirow} 
\usepackage{multicol}
\usepackage{array}
\usepackage{bm}
\let\ul\underline
\let\bs\boldsymbol
\let\sub\textsubscript
\DeclareMathOperator*{\argmin}{\operatorname{argmin}}
\definecolor{iccvblue}{rgb}{0.21,0.49,0.74}
\definecolor{red}{RGB}{192,0,0}
\definecolor{green}{RGB}{0,176,80}
\definecolor{blue}{RGB}{0,112,192}
\usepackage[pagebackref,breaklinks,colorlinks,allcolors=iccvblue]{hyperref}
\hypersetup{
bookmarksnumbered=true,
}
\usepackage{balance}
\usepackage{bookmark}
\usepackage[accsupp]{axessibility}  % Improves PDF readability for those with disabilities.
\usepackage{pifont}

%% file: sec/0_abstract.tex
\begin{abstract}
Visual localization, which estimates a camera's pose within a known scene, is a fundamental capability for autonomous systems. While absolute pose regression (APR) methods have shown promise for efficient inference, they often struggle with generalization. Recent approaches attempt to address this through data augmentation with varied viewpoints, yet they overlook a critical factor: appearance diversity.
In this work, we identify appearance variation as the key to robust localization. Specifically, we first lift real 2D images into 3D Gaussian Splats with varying appearance and deblurring ability, enabling the synthesis of diverse training data that varies not just in poses but also in environmental conditions such as lighting and weather. To fully unleash the potential of the appearance-diverse data, we build a two-branch joint training pipeline with an adversarial discriminator to bridge the syn-to-real gap.
Extensive experiments demonstrate that our approach significantly outperforms state-of-the-art methods, reducing translation and rotation errors by 50\% and 41\% on indoor datasets, and 38\% and 44\% on outdoor datasets. Most notably, our method shows remarkable robustness in dynamic driving scenarios under varying weather conditions and in day-to-night scenarios, where previous APR methods fail.
\end{abstract}

%% file: sec/1_intro.tex
\vspace{-20pt}
\section{Introduction}
\label{sec:intro}

Visual localization, the task of calculating a 6-DoF camera pose—its translation and rotation—based on a query image within a given environment, is essential for various applications, including robotics~\cite{biswas2012depth}, autonomous vehicles~\cite{Geiger2012KITTI}, and virtual reality~\cite{chessa2023detection}. Besides traditional geometry-based approaches, recent learning-based visual localization methods adopt absolute pose regression (APR)~\cite{kendall2015posenet, chen2022dfnet, brahmbhatt2018geometry, shavit2021learning}, scene coordinate regression (SCR)~\cite{brachmann2023accelerated, wang2024glace, brachmann2017dsac, moreau2023crossfire}, or post pose refinement (PPR)~\cite{germain2022feature, moreau2023crossfire, zhao2024pnerfloc, chen2024neural, liu2024hr, zhou2024nerfect, trivigno2024unreasonable}. SCR methods focus on learning-based 2D-3D correspondences followed by subsequent Perspective-n-Point (PnP) for pose estimation. PPR methods heavily rely on a pose prior, usually obtained from image retrieval, followed by iterative refinement.
In contrast, APR methods employ a supervised framework to train a regression neural network on image-pose pairs, enabling direct pose estimation during inference. APR offers faster runtime and lower error in challenging scenes with sparse views, significant lighting variations, or numerous dynamic objects, making it a promising method for ensuring robustness in real-world applications.

\begin{figure}
    \centering
    \includegraphics[width=1\linewidth]{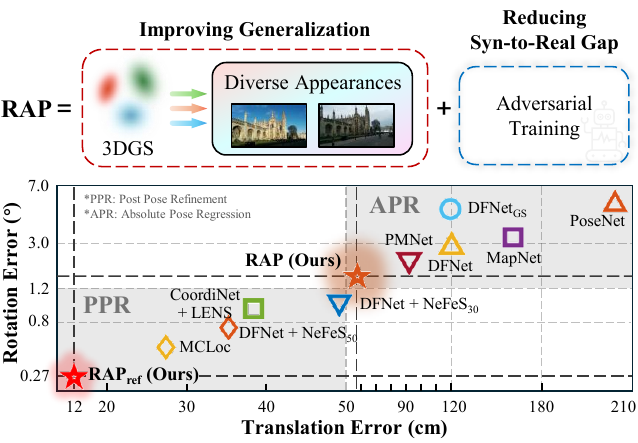}
    \vspace{-15pt}
    \caption{We propose RAP, a novel pipeline to train robust APR models. We lift real-world 2D images into 3D Gaussian Splats~\cite{kerbl20233d} to synthesize images with \textit{diverse appearances and poses}, improving model generalizability. We also introduce an adversarial discriminator, mitigating the syn-to-real gap to learn robust features. Together, we achieve state-of-the-art performance.}
    % We use tailored 3D Gaussian Splats~\cite{kerbl20233d} to synthesize diverse appearance images and propose a novel training paradigm for robust pose regression, achieving state-of-the-art performance. 
    %(b) Enhancing an APR method with appearance-diverse images, beyond spatial augmentation, significantly reduces rotation and translation errors under severe lighting variations.}
        % (a) To achieve robust and generalizable pose regression, we use tailored 3D Gaussian Splats~\cite{kerbl20233d} as the data engine to synthesize images with more appearance diversity, and propose a novel training paradigm for robust pose regression. Experiments validate the state-of-the-art performance of our method. (b) Adding images with more appearance diversity to an existing APR method that only performs spatial augmentation can substantially reduce both rotation and translation errors in scenes with severe lighting variations.}
    \label{fig:teaser}
    \vspace{-10pt}
\end{figure}

Despite promises, there is a performance gap in localization accuracy between APR and other methods. A well-known pivotal work~\cite{sattler2019understanding} attributes this to APR performing image-based memorization, i.e., retrieving poses seen during training. Driven by this crucial finding, to improve such memorization while avoiding the need for denser real-world training samples, recent methods leverage Neural Radiance Fields (NeRF)~\cite{mildenhall2020nerf} to synthesize additional posed images for APR training~\cite{moreau2022lens, chen2022dfnet, lin2024learning}.
LENS~\cite{moreau2022lens} tried to employ appearance perturbation using NeRF-W~\cite{martinbrualla2020nerfw}, but found the improvements minimal~\cite{moreau2022lens}. 
As appearance augmentation is common in other learning tasks, why does it fail in APR?
% However, these approaches primarily address spatial diversity while largely overlooking appearance variations between training and test images caused by changing weather and lighting conditions. This naturally raises two critical questions: (1) \textit{Beyond additional viewpoints, could more diverse appearances enhance APR's robustness?} (2) \textit{How can we better learn robust pose features from such diverse data?}

% We hypothesize a learning gap: Limitations in previous training pipelines prevented the effective use of diverse data to boost performance. 

% However, these approaches primarily address spatial diversity while largely overlooking appearance variations between training and test images caused by changing weather and lighting conditions. 

We hypothesize a learning gap: Limitations in previous training pipelines prevented the effective use of diverse data to boost performance. Artifacts always exist in images rendered by common novel view synthesis (NVS) methods, which might disturb the model feature space. Inspired by generative adversarial networks (GAN)~\cite{goodfellow2014generative}, where a discriminator is trained to distinguish between real and generated samples, we propose adversarial training for APR, designing a discriminator to align the features of synthetic and real images, thereby reducing the syn-to-real domain gap and mitigating the impact of rendering artifacts. Augmentation quality also matters. To efficiently synthesize diverse high-quality images with \textit{controllable} varying appearance, we extend the vanilla 3D Gaussian Splatting (3DGS)~\cite{kerbl20233d} to appearance-varying 3DGS with deblurring ability.

% To answer the two questions, we propose a novel \textbf{r}obust training framework for \textbf{a}bsolute \textbf{p}ose regression (\textbf{RAP}), as shown in Fig.~\ref{fig:teaser}.
% Firstly, we extend the vanilla 3D Gaussian Splatting (3DGS)~\cite{kerbl20233d} to appearance-varying 3DGS, allowing \textit{efficient} and \textit{diverse} data synthesis with \textit{controllable} varying appearance, which cannot be achieved by existing methods~\cite{chen2022dfnet, moreau2022lens}. To validate our hypothesis, we conduct preliminary experiments following the DFNet~\cite{chen2022dfnet} framework and settings. During pose augmentation, we utilize our appearance-varying 3DGS to randomly adjust image appearances, thereby increasing the training set's appearance diversity. In \textit{Court} and \textit{College} that are scenes with severe lighting variations from the Cambridge Landmarks dataset~\cite{kendall2015posenet}, this appearance augmentation reduces rotation and translation errors by 32.8\% / 18.5\% and 23.4\% / 16.7\% respectively, as shown in Fig.~\ref{fig:teaser}. These results strongly demonstrate the critical importance of appearance diversity in visual localization tasks.

These form our two-branch joint training framework for \textbf{r}obust \textbf{a}bsolute \textbf{p}ose regression (\textbf{RAP}). The first branch coarsely trains our Transformer-based pose regressor with both real data and data synthesized at the original real pose, together with an adversarial discriminator to reduce the syn-to-real domain gap. The second branch progressively generates randomly perturbed poses and appearances, providing additional supervision to the same APR Transformer. Through extensive experiments, we demonstrate that exploiting data diversity using adversarial training significantly increases localization accuracy in APR. Meanwhile, our results indicate that APR consistently benefits from more diverse visual data, and we observe clear signs of a more generalizable APR emerging with its localization performance cannot be explained merely by memorization.

Our contributions are summarized as follows:
\begin{itemize}
    \item We identify the crucial role of appearance diversity for APR, and develop a 3DGS-based appearance-varying data augmentation framework to efficiently generate diverse synthetic data with controllable lighting conditions.
    
    \item We propose an adversarial discriminator to reduce the syn-to-real gap. Together with progressive data synthesis, we form a robust two-branch joint training pipeline that fully unleashes the power of data diversity.
    
    \item We conduct extensive experiments showing our method outperforms state-of-the-art approaches on challenging datasets with significant appearance change. Ablation studies further analyze key factors affecting performance.
\end{itemize}

% \begin{itemize}
%     \item We identify research gaps in studying APR model generalizability and demonstrate through comprehensive experiments that large-scale data synthesis with effective learning strategies makes APR more generalizable.
%     \item We leverage 3D Gaussian Splats as our data engine to efficiently synthesize novel views with new appearances.
%     \item We propose a robust two-branch joint training pipeline with an adversarial discriminator to reduce the syn-to-real gap and fully unleash the power of synthetic data.
%     \item Our method sets a new state-of-the-art in visual localization on real-world indoor, outdoor, and driving datasets with significant appearance change.

% \end{itemize}

%% file: sec/2_Related_Works.tex
\begin{figure*}[ht]
    \centering
    \includegraphics[width=0.99\linewidth]{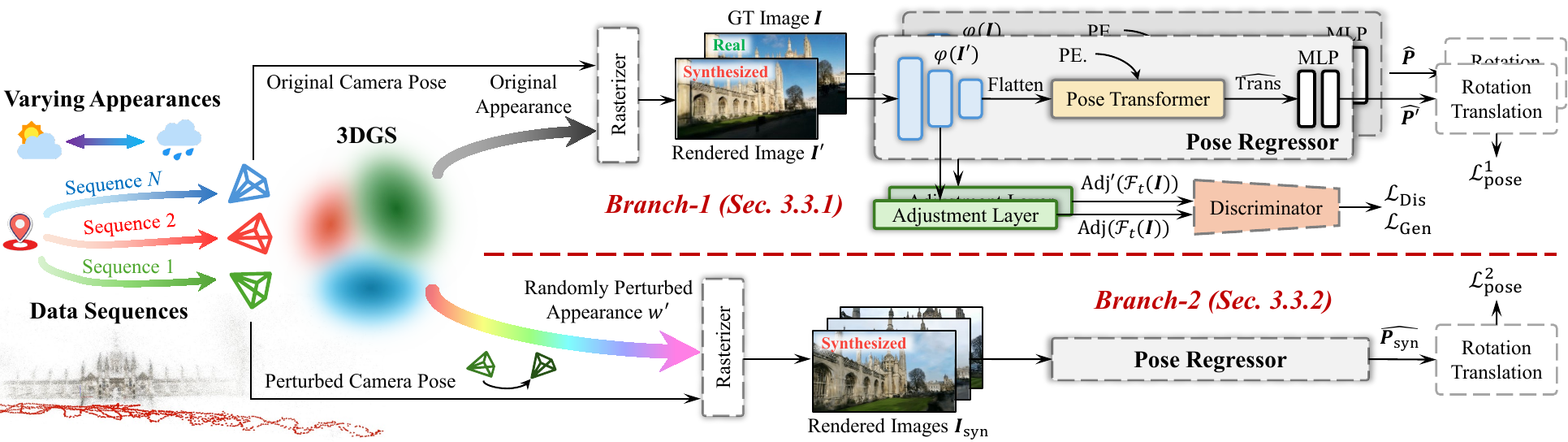}
    \caption{\textbf{Pipeline of RAP.} We lift multiple RGB video sequences into 3D Gaussian Splats, which serve as our data engine. The \textbf{branch-1} (see Sec.~\ref{branch-1}) inputs paired real and synthetic images to regress poses, with a discriminator to bridge the syn-to-real gap. The \textbf{branch-2} (see Sec.~\ref{branch-2}) generates views with novel poses and appearances, which are fed into the same pose regressor as additional supervision.}
    \vspace{-5pt}
    \label{fig:main}
\end{figure*}
\vspace{-5pt}
\section{Related Works}
\vspace{-5pt}
\noindent\textbf{Visual Localization.} Visual localization aims to estimate a camera's translation and rotation within a 3D scene. Traditional geometry-based methods~\cite{Chen_Baatz_Koser, Liu_Li_Dai_2017, dusmanu2019d2, sarlin2019coarse, taira2018inloc, noh2017large, sattler2016efficient, sarlin2020superglue, lindenberger2023lightglue} accomplish this by using point clouds and a reference image database, relying on stored descriptors and image retrieval to establish 2D-3D correspondences. In contrast, scene coordinate regression (SCR) methods~\cite{brachmann2017dsac, brachmann2023accelerated, wang2024glace, brachmann2021visual} embed map information within neural networks to directly predict 2D-3D correspondences. Both approaches generally require PnP~\cite{gao2003complete} and RANSAC~\cite{fischler1981random} to output camera poses at test time, which adds additional computation cost. Alternatively, absolute pose regression (APR)~\cite{kendall2015posenet, brahmbhatt2018geometry, Kendall_Cipolla_2017, moreau2022coordinet, Shavit_Ferens_Keller_2021, Chen_Wang_Prisacariu_2021} aims to directly regress the camera pose from a query image using neural networks. Although the performance is suboptimal compared with geometry-based methods, APR remains a promising approach due to its fast inference.

\noindent\textbf{Data Augmentation for Pose Regression.} End-to-end pose regression methods rely heavily on the amount and diversity of training data. Previous work~\cite{sattler2019understanding} shows that APR implicitly learns image \textit{retrieval} in the given environment. Therefore, the following works LENS~\cite{moreau2022lens}, DFNet~\cite{chen2022dfnet} and PMNet~\cite{lin2024learning} enhance APR performance by spatially enriching training views with NeRF.
However, these approaches fail to address the generalizability of APR models and exhibit several limitations: (1) The efficiency of training and novel view synthesis (NVS) in NeRF is severely restricted, hindering scalability. (2) They limit NVS to geometric (pose) transformations while neglecting photometric (appearance) variations, thereby decreasing APR robustness to changes in visual appearance. (3) The augmented data is underutilized in their learning frameworks, leaving its potential for improving APR largely untapped. Differently, our framework switches to 3DGS~\cite{kerbl20233d} as the scene representation to efficiently generate novel posed images with \textit{controllable} appearances and introduce adversarial training to unleash the power of such diverse data.

\noindent\textbf{Handling and Synthesizing Challenging Scenarios.} \label{related works: 2}
Visual localization often encounters unstructured photo collections~\cite{snavely2006photo}, where visual appearance varies due to moving objects, lighting changes, and inconsistent camera exposure settings. To tackle these in-the-wild challenges, NeRF-W~\cite{Martin-Brualla_Radwan_Sajjadi_Barron_Dosovitskiy_Duckworth_2021} uses per-image transient and appearance embeddings. In 3DGS~\cite{kerbl20233d}, VastGaussian~\cite{Lin2024VastGaussianV3} applies a CNN to 3DGS outputs but still struggles with significant appearance variations. SWAG~\cite{dahmani2024swag} mitigates this issue by storing appearance information in an external hash-grid-based implicit field, while GS-W~\cite{zhang2024gaussianwild3dgaussian} enhances flexibility by separating intrinsic and dynamic appearance features for each Gaussian point. 3DGM~\cite{li2024memorize} leverages consensus across multiple sequences as the self-supervision signal to remove transient and moving objects without human annotations. Deblur-GS~\cite{Chen_deblurgs2024} addresses motion blur---another challenge in localization datasets---by modeling camera motion to yield sharper edges in rendered scenes. Our method incorporates appearance modeling and edge refinement to handle and synthesize diverse indoor, outdoor, and driving scenes.

%% file: sec/3_Pre_Processing.tex
\vspace{-20pt}
\section{Method}
\vspace{-5pt}
\subsection{Pre-Processing with 3DGS}
\vspace{-5pt}
\label{Pre-Processing with 3DGS}
A robust pose regressor should focus on intrinsic scene attributes, not appearance variations. Therefore, we first synthesize diverse visual data for training. We leverage 3DGS~\cite{kerbl20233d}, representing scenes with explicit ellipsoids, to model diverse appearances.  % (see Sec.~\ref{supp: 3D Gaussian Splatting Preliminary} for more on 3DGS). 
Following GS-W~\cite{zhang2024gaussianwild3dgaussian}, we assume the scene contains \( K \) Gaussians and represent the independent intrinsic material attributes using positions \(\bs{\mu} \in \mathbb{R}^{K \times 3}\), spherical harmonics \(\bs{\mathcal{Y}} \in \mathbb{R}^{K \times 16 \times 3}\), and other parameters $\bs{\Theta}$ including rotation $\bs{q} \in \mathbb{R} ^ {K \times 4}$, scaling $\bs{s} \in \mathbb{R} ^ {K \times 3}$, and opacity $\bs{\alpha} \in \mathbb{R} ^ K$. To capture the dynamic appearance influenced by environmental factors, we extract features from the input image and assign each Gaussian its own feature using a learnable sampler $\bs{\mathcal{S}} \in \mathbb{R} ^ {K \times 2}$, forming features $\bs{\mathcal{E}} \in \mathbb{R} ^ {K \times 16 \times 3}$. We also incorporate the camera's view direction $\theta$ to account for viewpoint-dependent effects. The final color of Gaussians $\bs{\mathcal{C}} \in \mathbb{R} ^ {K \times 3}$ is:  % computed as follows:
\begin{equation}
    \bs{\mathcal{C}} = \operatorname{MLP}(\bs{\mu}, \bs{\mathcal{Y}}, \omega \bs{\mathcal{E}}, \theta),
\end{equation}
where $\omega$ is the blending weight that controls the dynamic appearance of the rendered image.
 
Another significant challenge in visual localization is motion blur, often caused by slow shutter speeds during video capture, leading to pose ambiguity and degraded rendering quality, further decreasing localization accuracy. Inspired by Deblur-GS~\cite{Chen_deblurgs2024}, we model camera motion blur as the inverse of scene motion, i.e., the transformation in Gaussian position denoted by $\bs{\mathcal{T}} \in \mathrm{SE}(3)$. 
For each training image, we sample a certain time step along a linear trajectory with a sampling weight $\phi \in \mathbb{R} ^ n$ and blend them to compute loss $\mathcal{L}$ with the original blurry image $\bs{\mathcal{I}}_b \in \mathbb{R} ^ {H \times W \times 3}$, optimizing $\bs{\mathcal{T}}$, $\phi$, $\bs{\mathcal{C}}$ and other 3DGS parameters $\bs{\Theta}$:
\begin{gather}
\argmin_{\phi, \bs{\mathcal{T}}, \bs{\mu}, \bs{\mathcal{Y}}, \bs{\mathcal{E}}, \bs{\Theta}} \mathcal{L} \left( \bs{\mathcal{I}}_b,~\sum_{i=1}^n \phi_i \operatorname{Render}(\bs{\mathcal{T}}_i(\bs{\mu}), \bs{\mathcal{C}}, \bs{\Theta}) \right),
\end{gather}
where the details of $\mathcal{L}$ are in supplemental materials.  % (Eq.~\ref{loss: 3dgs}).
After training, our 3DGS can efficiently render posed images given $\theta$ and $\omega$.

%% file: sec/4_Training.tex
\subsection{Architecture of Pose Regressor}
\label{Architecture of Pose Regressor}
Given a set of images and their associated camera poses $\{(\bs{I}_i, \bs{P}_i)\}_{i = 1}^n$, our goal is to train a neural network to directly output a homogeneous camera pose $\bs{P} \in \mathbb{R}^{3 \times 4}$ for a query image $\bs{I} \in \mathbb{R}^{H \times W \times C}$. Our network architecture is shown as the pose regressor in Fig.~\ref{fig:main}.

\noindent\textbf{Feature Extractor.} Pose regression networks typically extract features using a common backbone \(\varphi\), such as VGG~\cite{simonyan2014very} or EfficientNet~\cite{tan2019efficientnet}, leveraging multiple deeper layers for translation and rotation regression:
\begin{equation}
    \varphi(\bs{I})=\{\mathcal{F}_{0}(\bs{I}),..., \mathcal{F}_{N-1}(\bs{I}), \mathcal{F}_{N}(\bs{I})\},
\end{equation}
\(\mathcal{F}_{*}(\cdot)\) denotes features extracted from the \( * \)-th layer of a backbone with \( N \) layers. \(\mathcal{F}_{t}(\bs{I})\) and \(\mathcal{F}_{r}(\bs{I})\) denote features for translation and rotation regression, respectively.

\noindent\textbf{Pose Transformer.}  Unlike CNN-based regression models~\cite{chen2022dfnet,lin2024learning}, where fine-grained local features can introduce noise and harm performance, we propose \textit{Pose Transformer} to leverage the strong ability of Vision Transformer (ViT)~\cite{dosovitskiy2020image} for modeling long-range dependencies. Each Transformer generates a global token (\texttt{Trans} for translation and \texttt{Rot} for rotation) to provide a comprehensive context for pose regression, inspired by the \texttt{CLS} token in ViT. Given $\mathcal{F}_{r}(\bs{I})$ and $\mathcal{F}_{t}(\bs{I})$, the translation token is then concatenated with the flattened input features\footnote{We only present the translation regression for simplicity.}:
\begin{equation}
    \widetilde{\mathcal{F}_{t}}(\bs{I}) = \mathrm{Cat}(\mathrm{Flatten}(\mathcal{F}_{t}(\bs{I})), \texttt{Trans}) \in \mathbb{R}^{(H_t W_t + 1) \times C_t}.
\end{equation}
The positional encodings are then added to the flattened feature $( \texttt{PE} + \widetilde{\mathcal{F}_{t}}(\bs{I}) ) \in \mathbb{R}^{(H_t W_t + 1) \times C_t }$. Multi-head Self-Attention (MSA) is then conducted through a stack of multiple layers with the post-processing as follows:
\begin{equation}
\begin{aligned}
    \widehat{\mathcal{F}'_{t}}(\bs{I}) &= \mathrm{MSA}(\texttt{PE} + \widetilde{\mathcal{F}_{t}}(\bs{I})) + {\texttt{PE} + \widetilde{\mathcal{F}_{t}}(\bs{I})},\\
    \widehat{\mathcal{F}_{t}}(\bs{I}) &= \mathrm{LN}(\mathrm{FFN}(\mathrm{LN}(\widehat{\mathcal{F}'_{t}}(\bs{I}))) + \widehat{\mathcal{F}'_{t}}(\bs{I})),
\end{aligned}
\end{equation}
where $\mathrm{LN}$ indicates layer normalization and $\mathrm{FFN}$ denotes the fully connected feed-forward network, consisting of two linear layers with a ReLU. The final output is flattened back to $(H_tW_t + 1)\times c_t$. See supplementary for more details.  % Sec.~\ref{supp:PoseTransformer}.

\noindent\textbf{Regression Head.} Only the processed translation token, $\widehat{\texttt{Trans}}$, capturing global features for regression, is fed into the regression head.
This regression head consists of two MLPs, each with a hidden layer and GeLU activation:
\begin{equation}
    \hat{\bs{t}} = \mathrm{Linear}(\mathrm{GeLU(\mathrm{Linear}(\widehat{\texttt{Trans}}))}).
\end{equation}
The $\hat{\bs{t}}$ represents the final prediction for translation. Similarly, we obtain the rotation prediction denoted by $\hat{\bs{r}}$.

\subsection{Two-Branch Joint Training Paradigm}
% \vspace{-3pt}
\subsubsection{Branch-1: Aligning Features via Discriminator}
% \vspace{-6pt}
\label{branch-1}
Synthetic images from 3DGS provide novel viewpoints and appearances but often contain artifacts, leading to a syn-to-real domain gap. To align features from rendered and real images of the same pose, we introduce an adversarial training mechanism besides the basic pose regression training. 

\noindent\textbf{Pose Regression Loss.} For basic training, we render the synthetic image $\bs{I'}$ with the same pose label $\bs{P}$ as the real image $\bs{I}$, both used as supervision for the pose regressor. The training objective consists of translation loss $\mathcal{L}_t$ and the rotation loss $\mathcal{L}_r$, which are measured by the Euclidean distance between the ground truth pose $\bs{P} = \{\bs{t}, \bs{r}\}$ and the estimated pose $\widehat{\bs{P}} = \{\hat{\bs{t}}, \hat{\bs{r}}\}$:

\begin{gather}
    \mathcal{L}_t = \| \bs{t} - \hat{\bs{t}} \|_2, \\
    \mathcal{L}_r = \left\| \bs{r} - \frac{\hat{\bs{r}}}{\|\hat{\bs{r}}\|} \right\|_2, \\
    \mathcal{L}_\text{pose}^1 = \mathcal{L}_t \exp(-s_t) + s_t + \mathcal{L}_r \exp(-s_r) + s_r,
\end{gather}
where $s_t$ and $s_r$ are learned parameters for balancing the optimization between rotation and translation~\cite{kendall2017geometric}.

\noindent\textbf{Adversarial Loss.}
The adversarial training mechanism optimizes the discriminator to distinguish real from rendered image features, while training the feature extractor to fool the discriminator, effectively bridging the domain gaps.
To prevent vanishing gradients, we propose a novel adversarial objective for pose regression, inspired by LSGAN~\cite{mao2017least}:  
\vspace{-10pt}
\begin{align}
    \argmin_{D} \mathcal{L}_\text{Dis}(D) =& \frac{1}{2} \mathbb{E}_{\bs{I} \sim p_{\text{data}}(\bs{I})} 
    \left[ (D(\mathrm{Adj}(\mathcal{F}_{t}(\bs{I}))) - 1)^2 \right] \nonumber\\
    +& \frac{1}{2} \mathbb{E}_{\bs{I'}} \left[ D(\mathrm{Adj'}(\mathcal{F}_{t}(\bs{I'}))^2 \right],\\
    \argmin_{G} \mathcal{L}_\text{Gen}(G) =& \frac{1}{2} \mathbb{E}_{\bs{I'}} 
    \left[ (D(\mathrm{Adj'}(\mathcal{F}_{t}(\bs{I'})) - 1)^2 \right].
\end{align}
Here, $\mathrm{Adj}$ and $\mathrm{Adj'}$ are the adjustment layers, consisting of Conv-ReLU-BN layers. The feature extractor $\varphi$ acts as the generator $G$, while $D$ is the discriminator, composed of several convolutional layers with ReLU activations. More details are in supplemental materials.  % (Sec.~\ref{supp: Adversarial Discriminator}).  

\subsubsection{Branch-2: Training while Synthesizing Data}  
\label{branch-2}
% \vspace{-5pt}
With the proposed appearance-varying 3DGS, more posed images are generated to enrich the training data for better generalizability. Specifically, our data synthesis is categorized into two dimensions: \textit{pose augmentation} and \textit{appearance augmentation}, as illustrated in Fig.~\ref{fig:teaser}. For pose augmentation, given a training pose $\bs{P}$, a perturbed pose $\bs{P}_\text{syn}$ is generated around $\bs{P}$ by the translation noise of $\delta t$ and rotation noise of $\delta r$. 
For appearance augmentation, we randomly adjust the appearance of rendered images using random blending weights $\omega$, and then render the synthetic image $\bs{I}_\text{syn}$ using the Gaussian Splats trained in Sec.~\ref{Pre-Processing with 3DGS}. The novel image-pose pair $(\bs{I}_\text{syn}, \bs{P}_\text{syn})$, online generated every 20 epochs during training until the validation MSE loss and median errors cease to decrease, serves as additional supervision for the training. Given the estimated pose of the synthesized image denoted by $\widehat{\bs{P}_\text{syn}}$, the loss function $\mathcal{L}_\text{pose}^2(\widehat{\bs{P}_\text{syn}}, \bs{P}_\text{syn})$ is same as $\mathcal{L}_\text{pose}^1$.

\subsubsection{Overall Objective}
\label{overall objective}
The total loss for the pose regressor is: 
\begin{equation}
    \mathcal{L}_\text{total} = \beta_1\mathcal{L}_\text{pose}^1 + \beta_2\mathcal{L}_\text{pose}^2 + \beta_3 (\mathcal{L}_\text{Gen} + \mathcal{L}_\text{Dis}),
\label{eq: loss_total}
\end{equation}
where $\beta_1, \beta_2, \beta_3$ are loss weights. The total loss will optimize the pose regressor, adjustment layers, and discriminator. Only the pose regressor will be deployed in the inference phase, while the other two components are discarded.

%% file: sec/5_Experiment.tex
\begin{figure*}[t]
    \centering
    \includegraphics[width=1\linewidth]{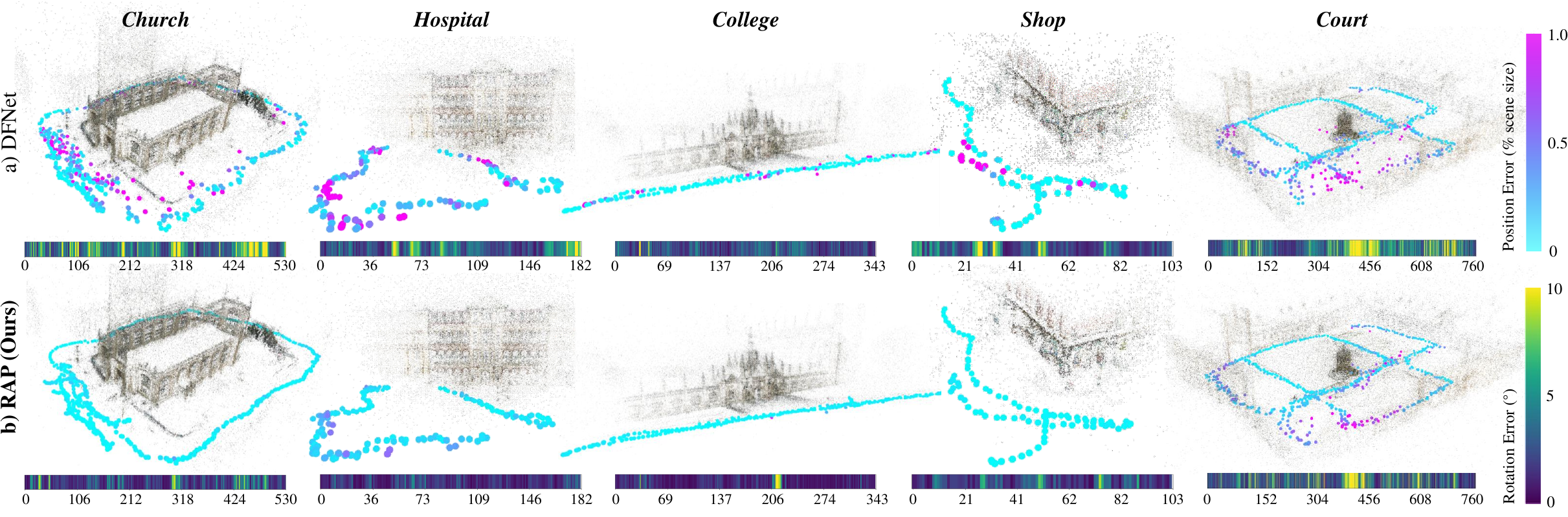}
    \vspace{-18pt}
    \caption{\textbf{Qualitative comparison of camera pose estimation errors between a) DFNet~\cite{chen2022dfnet} and b) our RAP framework} across five scenes on the Cambridge Landmarks dataset~\cite{kendall2015posenet}. Our RAP framework estimates trajectories that more closely follow the ground truth, with significantly reduced rotation and position errors compared to DFNet~\cite{chen2022dfnet}.
    }
    % \caption{\textbf{Qualitative comparison of camera pose estimation errors between a) DFNet~\cite{chen2022dfnet} and b) our RAP framework} across four scenes on the Cambridge Landmarks dataset~\cite{kendall2015posenet}. The rotation errors are represented by the color bar on the upper right (in degrees), while position errors (as a percentage of scene size) are shown based on the lower right color bar. Our RAP framework estimates trajectories that more closely follow the ground truth, with significantly reduced rotation and position errors compared to DFNet~\cite{chen2022dfnet}.
    % }
    \label{fig: Cam_vis}
    \vspace{-10pt}
\end{figure*}

\section{Experiments}
\label{Experiments}

\subsection{Evaluation Setup}
\label{Evaluation Setup}
\noindent\textbf{Datasets.} 
We follow previous works~\cite{chen2022dfnet, lin2024learning} to mainly use four scenes in the Cambridge Landmarks dataset~\cite{kendall2015posenet} with spatial extents around 875 m$^2$.  
Moreover, we evaluate our method on MARS~\cite{li2024multiagent}, a self-driving dataset featuring challenges like moving objects, lighting changes, and motion blur. To investigate the robustness of our model under extreme lighting changes, such as the transition from day to night, we also prepared a subset of the Aachen Day-Night dataset~\cite{sattler2018benchmarking}. The training data includes images captured using various camera models with differing resolutions, which renders direct evaluation with APR methods infeasible. Thus, we standardized the camera models through center cropping and built a COLMAP~\cite{schonberger2016structure} model as pose annotations, including 13 nighttime images for evaluation and 246 daytime images for training 3DGS and RAP. We also employ the 7-Scenes dataset~\cite{shotton2013scene}, which provides seven indoor scenes with volumes spanning 1 m$^3$--18 m$^3$, and follow the original training and testing splits with more accurate SfM pose annotations~\cite{brachmann2023accelerated,chen2024neural}. Although it is an indoor dataset, it is still non-trivial as it includes texture-less surfaces, object occlusions, and motion blur.

\begin{table}[t]
\renewcommand{\arraystretch}{1.15} 
    \caption{\textbf{Median translation (cm) and rotation ($^\circ$) errors on the Cambridge Landmarks dataset~\cite{kendall2015posenet}.} The best results$^\dag$ in pure APR and PPR are highlighted in \hl{\textbf{bold}}.}
    \label{table:cambridge}
    \vspace{-6pt}
    % \centering
    \setlength{\tabcolsep}{2.5pt}
    \scalebox{0.65}{
    \begin{tabular}{c|c|cccc|c|c}
    \toprule
    & \textbf{Methods}                      & \textbf{\textit{College}}      & \textbf{\textit{Hospital}}     & \textbf{\textit{Shop}}        & \textbf{\textit{Church}}       & \textbf{Average$^\ddag$}               & \textbf{\textit{Court}}         \\
    \midrule
    \multirow{8}{*}{\textbf{\makecell{Pure\\APR}}}
    & PN~\cite{kendall2015posenet}          & 166/4.86                       & 262/4.90                       & 141/7.18                      & 245/7.95                       & 204/6.23                        & 683/3.50                        \\
    & MapNet~\cite{brahmbhatt2018geometry}  & 107/1.89                       & 194/3.91                       & 149/4.22                      & 200/4.53                       & 163/3.64                        & N/A                             \\
    % & MSPN~\cite{blanton2020extending}    & 1.73/3.65                      & 2.55/4.05                      & 2.02/7.49                     & 2.67/6.18                      & 2.47/5.34                       & N/A                             \\
    & MS-Trans.~\cite{shavit2021learning}   & 83/1.47                        & 181/2.39                       & 86/3.07                       & 162/3.99                       & 128/2.73                        & N/A                             \\
    & PAE~\cite{shavit2022camera}           & 90/1.49                        & 207/2.58                       & 99/3.88                       & 164/4.16                       & 140/3.03                        & N/A                             \\
    & LENS$^\dag$~\cite{moreau2022lens}     & 33/0.50                        & 44/0.90                        & 27/1.60                       & 53/1.60                        & 39/1.20                         & N/A                             \\
    & DFNet~\cite{chen2022dfnet}            & 73/2.37                        & 200/2.98                       & 67/2.21                       & 137/4.03                       & 119/2.90                        & 217/4.11                        \\
    & PMNet~\cite{lin2024learning}          & 68/1.97                        & 103/1.31                       & 58/2.10                       & 133/3.73                       & 90/2.27                         & N/A                             \\
    & \textbf{RAP (Ours)}                   & \hl{\textbf{52}/\textbf{0.90}} & \hl{\textbf{87}/\textbf{1.21}} & \hl{\textbf{33/1.48}}         & \hl{\textbf{53}/\textbf{1.52}} & \hl{\textbf{56}/\textbf{1.28}}  & \hl{\textbf{115}/\textbf{1.68}} \\
    \midrule
    \multirow{3}{*}{\textbf{SCR}}
    & DSAC*~\cite{brachmann2021visual}      & \ul{18}/\ul{0.3}               & 21/\ul{0.4}                    & 5/0.3                         & 15/0.6                         & 15/0.4                          & 34/0.2                          \\
    & ACE~\cite{brachmann2023accelerated}   & 28/0.4                         & 31/0.6                         & 5/0.3                         & 18/0.6                         & 21/0.5                          & 43/0.2                          \\
    & GLACE~\cite{wang2024glace}            & 19/\ul{0.3}                    & \ul{17}/\ul{0.4}               & \ul{4}/\ul{0.2}               & \ul{9}/\ul{0.3}                & \ul{12}/\ul{0.3}                & \ul{19}/\ul{0.1}                \\
    \midrule
    \multirow{10}{*}{\textbf{PPR}}
    & FQN-MN~\cite{germain2022feature}      & 28/0.38                        & 54/0.82                        & 13/0.63                       & 58/2.00                        & 38/0.96                         & 4253/39.16                      \\
    & LENS~\cite{moreau2022lens}            & 34/0.54                        & 45/0.96                        & 28/1.66                       & 54/1.66                        & 40/1.21                         & N/A                             \\
    & CrossFire~\cite{moreau2023crossfire}  & 47/0.7                         & 43/0.7                         & 20/1.2                        & 39/1.4                         & 37/1.0                          & N/A                             \\
    % & NeFeS$_{30}$~\cite{chen2024neural}  & 37/0.64                        & 98/1.61                        & 17/0.60                       & 42/1.38                        & 49/1.06                         & N/A                             \\
    & NeFeS$_{50}$~\cite{chen2024neural}    & 37/0.54                        & 52/0.88                        & 15/0.53                       & 37/1.14                        & 35/0.77                         & N/A                             \\
    & HR-APR~\cite{liu2024hr}               & 36/0.58                        & 53/0.89                        & 13/0.51                       & 38/1.16                        & 35/0.78                         & N/A                             \\
    & MCLoc~\cite{trivigno2024unreasonable} & 31/0.42                        & 39/0.73                        & 12/0.45                       & 26/0.88                        & 27/0.62                         & N/A                             \\
    & DFNet\sub{GS-CPR}~\cite{liu2025gscpr} & 23/0.32                        & 42/0.74                        & 10/0.36                       & 27/0.62                        & 26/0.51                         & N/A                             \\
    & ACE\sub{GS-CPR}~\cite{liu2025gscpr}   & 20/0.29                        & 21/0.40                        & \hl{\textbf{5}}/0.24          & 13/0.40                        & 15/0.33                         & N/A                             \\
    & DFNet$_\textbf{ref (Ours)}$           & 16/0.24                        & 21/0.41                        & 8/0.42                        & 10/0.26                        & 14/0.33                         & 25/\hl{\textbf{0.13}}           \\
    & \textbf{RAP\sub{ref} (Ours)}          & \hl{\textbf{15}/\textbf{0.23}} & \hl{\textbf{18}/\textbf{0.38}} & \hl{\textbf{5}/\textbf{0.23}} & \hl{\textbf{9}/\textbf{0.23}}  & \hl{\textbf{12}/\textbf{0.27}}  & \hl{\textbf{22}}/0.15           \\
    \bottomrule
    \end{tabular}}
    \vspace{-10pt}
    \footnotesize{\par \bigskip
    $^\ddag$Since most methods did not report results on \textit{Court}, it is excluded from the average error calculation. $^\dag$As CoordiNet + LENS~\cite{moreau2022lens} does not provide open-source code, it is unclear whether any post-processing is used.}
    \vspace{-20pt}
\end{table}

\noindent\textbf{Baselines.} We first compare our proposed RAP against common APR-only approaches on the four datasets, where PMNet~\cite{lin2024learning} and DFNet~\cite{chen2022dfnet} are the most related and advanced methods based on data augmentation. We split the remaining methods into two categories based on whether they rely on extra novel view synthesis in test time, including SCR~\cite{brachmann2017dsac,brachmann2023accelerated,wang2024glace} and PPR (Post Pose Refinement)~\cite{germain2022feature, moreau2023crossfire, zhao2024pnerfloc, chen2024neural, liu2024hr, zhou2024nerfect, trivigno2024unreasonable}, which involves rendering images, querying features in novel views by the initial pose, iterative refinement or sequential refinement~\cite{moreau2022coordinet}. 
% MCLoc~\cite{trivigno2024unreasonable} is a post-refinement method using 3DGS, while others use NeRF.

\begin{figure}[t]
    \centering
    \includegraphics[width=1\linewidth]{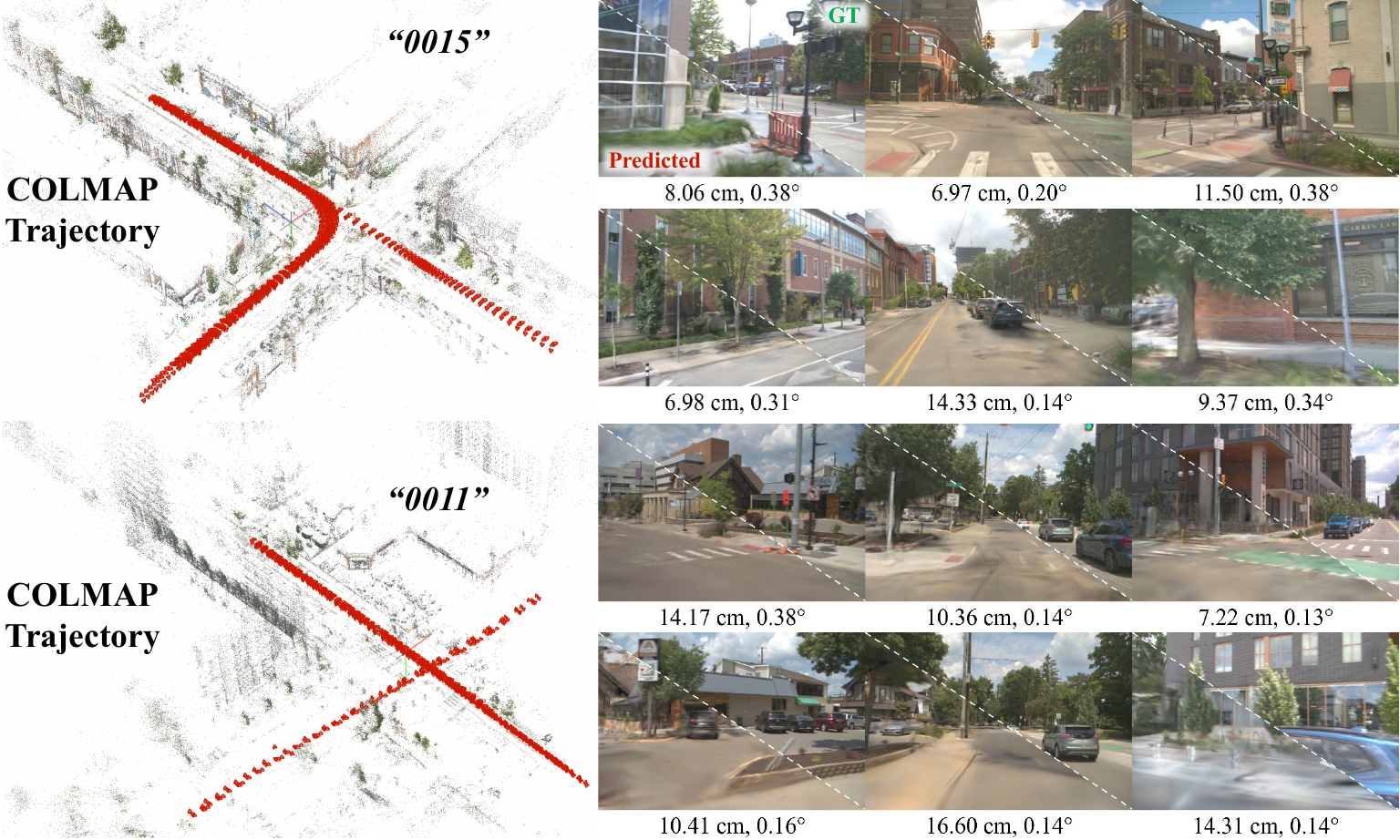}
    \vspace{-18pt}
    % \captionsetup{font=small}
    \caption{\textbf{Visualization of RAP\sub{ref} on MARS~\cite{li2024multiagent}.} In each sub-figure, a diagonal line separates the ``Predicted" (rendered from the refined pose) and ``GT" (ground truth) sections. Smooth alignment along this boundary shows RAP\sub{ref}'s improved pose accuracy.}  %  Our method achieves precise localization even in highly dynamic scenarios with varying weather conditions.
    \label{fig: mars_vis}
    \vspace{-5pt}
\end{figure}

\begin{table}[t]
\renewcommand{\arraystretch}{1.15} 
    \caption{\textbf{Median translation (cm) and rotation ($^\circ$) errors on the MARS dataset~\cite{li2024multiagent}.}}  % We compare our pipeline with DFNet~\cite{chen2022dfnet} and PoseNet~\cite{kendall2015posenet} in translation error (cm) and rotation (°) errors.
    \label{table:MARS}
    \vspace{-5pt}
    \centering
    \scalebox{0.65}{
    \begin{tabular}{c|cccc|c}
    \toprule
    \textbf{Methods}                  & \textbf{\textit{``0011"}}       & \textbf{\textit{``0015"}}       & \textbf{\textit{``0037"}} & \textbf{\textit{``0041"}}       & \textbf{Average}                \\
    \midrule
    PoseNet~\cite{kendall2015posenet} & 149/1.80                        & 136/2.34                        & 123/1.60                  & 75/0.92                         & 121/1.67                        \\
    % DFNet~\cite{chen2022dfnet}        & 298/4.70                        & 528/10.25                       & 535/7.18                  & 642/4.43                        & 530/6.64                        \\
    \textbf{RAP (Ours)}               & \ul{32}/\ul{0.61}               & \ul{37}/\ul{1.08}               & \ul{15}/\ul{0.35}         & \ul{28}/\ul{0.35}               & \ul{28}/\ul{0.60}               \\
    \textbf{RAP\sub{ref} (Ours)}      & \hl{\textbf{8.5}/\textbf{0.13}} & \hl{\textbf{8.2}/\textbf{0.20}} & \hl{\textbf{8.7/0.09}}    & \hl{\textbf{7.6}/\textbf{0.11}} & \hl{\textbf{8.3}/\textbf{0.13}} \\
    \bottomrule
    \end{tabular}}
    % \vspace{-5pt}
\end{table}

\begin{table}[t]
    \caption{\textbf{Median translation (COLMAP~\cite{zheng2015structure} unit) and rotation ($^\circ$) errors on the Aachen Day-Night Dataset~\cite{sattler2018benchmarking}}.}  % Day-time images are used in training, while night-time images are only used for testing. 
    \label{table:aachen}
    \vspace{-5pt}
    \centering
    \setlength{\tabcolsep}{3.5pt}
    \scalebox{0.65}{
    \begin{tabular}{ccccc|cc}
    \toprule
    \multicolumn{5}{c|}{\textbf{APR-Based}} & \multicolumn{2}{c}{\textbf{SCR-Based}}  \\
    \midrule
    PoseNet~\cite{kendall2015posenet} & DFNet~\cite{chen2022dfnet} & \textbf{RAP w/o App.} & \textbf{RAP}        & \textbf{RAP\sub{ref}}          & ACE~\cite{brachmann2023accelerated} & GLACE~\cite{wang2024glace} \\
    \midrule
    217/74.30                         & 174/85.80                  & 134/75.99             & \ul{130}/\ul{13.70} & \hl{\textbf{50}/\textbf{3.93}} & 914/90.50                           & 482/36.4 \\
    \bottomrule
    \end{tabular}}
    \vspace{-10pt}
\end{table}

\begin{figure*}[ht]
    \centering
    \includegraphics[width=1\linewidth]{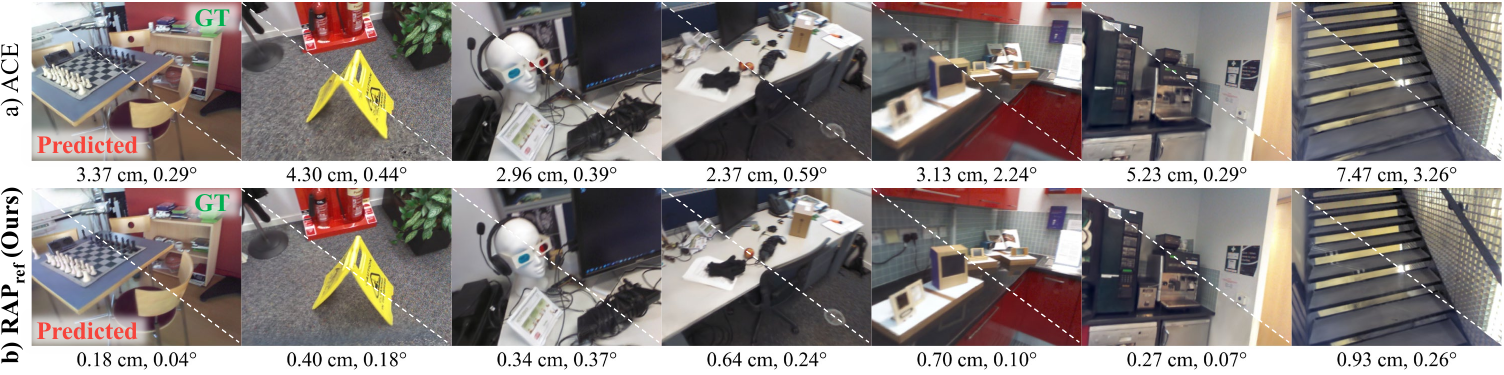}
    \vspace{-18pt}
    \caption{\textbf{Visualization of the localization errors of RAP\sub{ref} on the 7-Scenes dataset~\cite{shotton2013scene}.}}
    \label{fig: 7scenes_vis}
    \vspace{-2pt}
\end{figure*}

\begin{table*}[t]
% \footnotesize
    \caption{\textbf{Quantitative results on the 7-Scenes dataset~\cite{shotton2013scene}.} The best results in pure APR and PPR are highlighted in \hl{\textbf{bold}}. \textbf{DSLAM GT} and \textbf{SfM GT} refer to different sets of ground truth. More visualizations and details are in supplemental materials.}  % (Fig.~\ref{fig: supp_7scenes_rap}) (Sec.~\ref{supp: ground truth pose}).}  % We compare our method against three categories of approaches (single-frame APR, SCR, and PPR) based on median translation (cm) and rotation ($^\circ$) errors.  % Fig.~\ref{fig: supp_7scenes_rap}.
    \label{table:7scenes}
    \vspace{-5pt}
    \centering
    \scalebox{0.65}{
    \begin{tabular}{c|c|ccccccc|c}
    \toprule
    \textbf{Category} & \textbf{Methods}          & \textbf{\textit{Chess}}          & \textbf{\textit{Fire}}           & \textbf{\textit{Heads}}          & \textbf{\textit{Office}}         & \textbf{\textit{Pumpkin}}        & \textbf{\textit{Kitchen}}        & \textbf{\textit{Stairs}}         & \textbf{Average}                 \\
    \cmidrule{1-10}
    \multirow{10}{*}{\textbf{\makecell{Pure\\APR}}}
    & PoseNet (PN)~\cite{kendall2015posenet}      & 32/8.12                          & 47/14.4                          & 29/1.20                          & 48/7.68                          & 47/8.42                          & 59/8.64                          & 47/13.80                         & 44/10.4                          \\
    & MapNet~\cite{brahmbhatt2018geometry}        & 8/3.25                           & 27/11.7                          & 18/13.3                          & 17/51.5                          & 22/4.02                          & 23/4.93                          & 30/12.1                          & 21/7.77                          \\
    & AtLoc+~\cite{wang2020atloc}                 & 10/3.18                          & 26/10.8                          & 14/11.4                          & 17/5.16                          & 20/3.94                          & 16/4.90                          & 29/10.2                          & 19/7.08                          \\
    & MS-Transformer~\cite{shavit2021learning}    & 11/4.66                          & 24/9.60                          & 14/12.2                          & 17/5.66                          & 18/4.44                          & 17/5.94                          & 17/5.94                          & 18/7.28                          \\
    & PAE~\cite{shavit2022camera}                 & 12/4.95                          & 24/9.31                          & 14/12.5                          & 19/5.79                          & 18/4.89                          & 18/6.19                          & 25/8.74                          & 19/7.48                          \\
    & CoordiNet + LENS~\cite{moreau2022lens}      & 4/1.38                           & 11/3.77                          & 8/5.86                           & 8/1.98                           & 9/2.27                           & 10/2.27                          & 15/3.67                          & 9/3.07                           \\
    & DFNet~\cite{chen2022dfnet}                  & 5/1.88                           & 17/6.45                          & 6/\hl{\textbf{3.63}}             & 8/2.48                           & 10/2.78                          & 22/5.45                          & 16/3.29                          & 12/3.71                          \\
    & PMNet~\cite{lin2024learning}                & 4/1.70                           & 10/4.51                          & 7/4.23                           & 7/1.96                           & 14/3.33                          & 14/3.36                          & 16/3.62                          & 10/3.24                          \\
    & \textbf{RAP (Ours, DSLAM GT)}               & {\textbf{3}/\textbf{1.41}}       & {\textbf{7}/\textbf{3.46}}       & {\textbf{6}}/6.02                & {\textbf{5}/1.97}                & {\textbf{6}/\textbf{1.96}}       & {\textbf{7}/\textbf{2.18}}       & {\textbf{10}/\textbf{2.14}}      & {\textbf{6}/\textbf{2.73}}       \\
    & \textbf{RAP (Ours, SfM GT)}                 & \hl{\textbf{2}/\textbf{0.85}}    & \hl{\textbf{6}/\textbf{2.84}}    & \hl{\textbf{4}}/4.52             & \hl{\textbf{4}/\textbf{1.57}}    & \hl{\textbf{3}/\textbf{1.10}}    & \hl{\textbf{5}/\textbf{1.10}}    & \hl{\textbf{10}/\textbf{1.30}}   & \hl{\textbf{5}/\textbf{1.90}}    \\
    \midrule
    \multirow{4}{*}{\textbf{SCR}}
    & DSAC~\cite{brachmann2021visual}             & 0.5/0.17                         & 0.8/0.28                         & 0.5/0.34                         & 1.2/0.34                         & 1.2/0.28                         & 0.7/0.21                         & 2.7/0.78                         & 1.1/0.34                         \\
    & ACE~\cite{brachmann2023accelerated}         & 0.5/0.18                         & 0.8/0.33                         & 0.5/0.33                         & 1.0/0.29                         & 1.0/0.22                         & 0.8/0.20                         & 2.9/0.81                         & 1.1/0.34                         \\
    & GLACE~\cite{wang2024glace}                  & 0.6/0.18                         & 0.9/0.34                         & 0.6/0.34                         & 1.1/0.29                         & 0.9/0.23                         & 0.8/0.20                         & 3.2/0.93                         & 1.2/0.36                         \\
    & \textit{marepo}$^\ddag$~\cite{chen2024map}  & 2.6/1.35                         & 2.5/1.42                         & 2.3/2.21                         & 3.6/1.44                         & 4.2/1.55                         & 5.1/1.99                         & 6.7/1.83                         & 3.9/1.68                         \\
    \midrule
    \multirow{9}{*}{\textbf{PPR}}
    & FQN-MN~\cite{germain2022feature}            & 4.1/1.31                         & 10.5/2.97                        & 9.2/2.45                         & 3.6/2.36                         & 4.6/1.76                         & 16.1/4.42                        & 139.5/34.67                      & 28/7.3                           \\
    & CrossFire~\cite{moreau2023crossfire}        & 1/0.4                            & 5/1.9                            & 3/2.3                            & 5/1.6                            & 3/0.8                            & 2/0.8                            & 12/1.9                           & 4.4/1.38                         \\
    % & pNeRFLoc~\cite{zhao2024pnerfloc}          & 2/0.8                            & 2/0.88                           & 1/0.83                           & 3/1.05                           & 6/1.51                           & 5/1.54                           & 32/5.73                          & 7.3/1.76                         \\
    & DFNet + NeFeS$_{50}$~\cite{chen2024neural}  & 2/0.57                           & 2/0.74                           & 2/1.28                           & 2/0.56                           & 2/0.55                           & 2/0.57                           & 5/1.28                           & 2.4/0.79                         \\
    & HR-APR~\cite{liu2024hr}                     & 2/0.55                           & 2/0.75                           & 2/1.45                           & 2/0.64                           & 2/0.62                           & 2/0.67                           & 5/1.30                           & 2.4/0.85                         \\
    & MCLoc~\cite{trivigno2024unreasonable}       & 2/0.8                            & 3/1.4                            & 3/1.3                            & 4/1.3                            & 5/1.6                            & 6/1.6                            & 6/2.0                            & 4.1/1.43                         \\
    & DFNet + GS-CPR (SfM GT)~\cite{liu2025gscpr} & 0.7/0.20                         & 0.9/0.32                         & 0.6/0.36                         & 1.2/0.32                         & 1.3/0.31                         & 0.9/0.25                         & 2.2/0.61                         & 1.1/0.34                         \\
    & ACE + GS-CPR (SfM GT)~\cite{liu2025gscpr}   & 0.5/0.15                         & 0.6/0.25                         & 0.4/0.28                         & 0.9/0.26                         & 1.0/0.23                         & 0.7/0.17                         & 1.4/0.42                         & 0.8/0.25                         \\
    & \textbf{RAP\sub{ref} (Ours, DSLAM GT)}      & 2.78/1.43                        & 2.07/1.23                        & 1.53/1.87                        & 2.49/1.20                        & 4.47/1.56                        & 4.21/1.83                        & 3.24/1.18                        & 2.97/1.47                        \\
    & \textbf{RAP\sub{ref} (Ours, SfM GT)}        & \hl{\textbf{0.33}/\textbf{0.11}} & \hl{\textbf{0.51}/\textbf{0.21}} & \hl{\textbf{0.39}/\textbf{0.27}} & \hl{\textbf{0.57}/\textbf{0.16}} & \hl{\textbf{0.81}/\textbf{0.20}} & \hl{\textbf{0.45}/\textbf{0.12}} & \hl{\textbf{1.11}/\textbf{0.32}} & \hl{\textbf{0.60}/\textbf{0.20}} \\
    \bottomrule 
    \end{tabular}}
    \vspace{-10pt}
    \footnotesize{\par  \bigskip
    $^\ddag$As \textit{marepo}~\cite{chen2024map} combines SCR and APR, we classify it as SCR.}
    \vspace{-10pt}
\end{table*}

\noindent\textbf{Implementation Details.} 
\label{Implementation}
First, we optimize our 3DGS for each scene without masking moving objects.
We then train our RAP network, which uses an Efficient-B0 backbone~\cite{liu2017efficient} pre-trained on ImageNet~\cite{deng2009imagenet}, optimized with Adam~\cite{Kingma15adam} at a learning rate of $10^{-4}$. Only the features from the third (\texttt{reduction\_3}) and fourth (\texttt{reduction\_4}) layers are used respectively for translation and rotation regression, and both layers are utilized for narrowing the domain gap via the discriminator, which is also optimized with Adam~\cite{Kingma15adam}, using a learning rate of $10^{-4}$ and betas set to (0.5, 0.999).
More details about training are in supplemental materials.  % (Sec.~\ref{supp: Two-Branch Joint APR Training}). 
For generating random views, we apply random normalized perturbations to each training pose: $\delta t = 20$ cm and $\delta r = 10^\circ$ for indoor scenes, and $\delta t = 150$ cm and $\delta r = 4^\circ$ for outdoor scenes. 

To allow for comparison with SCR methods and leverage 3DGS’s efficient rendering for PPR, we extend the APR pipeline with match-based refinement similar to GS-CPR~\cite{liu2025gscpr}, denoted as RAP\sub{ref}. At test time, RAP's initial pose is used to render an RGB-D image via 3DGS. Together with MASt3R~\cite{leroy2024grounding}, we can obtain 2D-3D correspondences to perform RANSAC-PnP~\cite{fischler1981random, gao2003complete}, resulting in a refined pose. More details are in supplementary materials.  % Sec.~\ref{Implementation Details} and Sec.~\ref{supp: metric}.

\subsection{Benchmark Results}

\noindent\textbf{Cambridge Landmarks~\cite{kendall2015posenet}.} In the challenging outdoor Cambridge Landmarks dataset (Table~\ref{table:cambridge}), our RAP reduces both translation and rotation errors across all scenes by over 30\% compared to other APR-only methods. The visualization in Fig.~\ref{fig: Cam_vis} shows that our method produces fewer outliers than DFNet~\cite{chen2022dfnet}. In the three larger-scale scenes with significant appearance diversity (\textit{College}, \textit{Church}, and \textit{Court}), rotation error is even halved compared to DFNet. Table~\ref{table:cambridge} also shows the effectiveness of our RAP\sub{ref} in further reducing pose errors through refinement. 
RAP\sub{ref} outperforms CoordiNet + LENS~\cite{moreau2022lens}, which assumes a continuous trajectory when an Extended Kalman Filter~\cite{EKF} is required for refinement~\cite{moreau2022coordinet}.
RAP\sub{ref} even surpasses ACE~\cite{brachmann2023accelerated} and its post-refinement variant, ACE + GS-CPR~\cite{liu2025gscpr}, despite GS-CPR manually masking dynamic objects when building 3DGS. This demonstrates the strong representation capability of our appearance-varying 3DGS with deblurring.

\noindent\textbf{MARS~\cite{li2024multiagent}.} Autonomous driving scenarios present unique challenges, including moving objects and frequent changes in lighting conditions, as illustrated in Fig.~\ref{fig: mars_vis}. Our RAP demonstrates effective and robust performance across four challenging scenes, as shown in Table~\ref{table:MARS}, achieving an average of 28 cm / 0.60$^\circ$ localization error. This significantly outperforms the baseline\footnote{DFNet~\cite{chen2022dfnet} results are omitted as we were unable to successfully train its NeRF component, likely due to the need for manual scene scaling within 
\([- \pi, \pi]\), which is tedious for diverse outdoor scenes.} PoseNet~\cite{kendall2015posenet}.  % and DFNet~\cite{chen2022dfnet}, which struggles to model such large-scale driving scenarios using NeRF due to noise introduced during the learning process. 
With one-shot refinement, our RAP\sub{ref} further reduces outdoor localization errors to below 10 cm.

\noindent\textbf{Aachen Day-Night~\cite{sattler2018benchmarking}.} Benefiting from appearance augmentation, our RAP significantly reduces the localization error from 134 unit / 75.99$^\circ$ to 130 unit / 13.70$^\circ$, outperforming other APR~\cite{kendall2015posenet, chen2022dfnet} and SCR~\cite{brachmann2023accelerated,wang2024glace} baselines, as shown in Table~\ref{table:aachen}. This demonstrates the effectiveness of appearance diversity in handling extreme lighting changes.

\noindent\textbf{7-Scenes~\cite{shotton2013scene}.} As shown in Table~\ref{table:7scenes}, our RAP reduces translation error by 50\% (10 cm → 5 cm) and rotation error by 41.36\% (3.24° → 1.90°) on average compared to previous state-of-the-art single-frame APR methods.
The only exception is \textit{Heads}, where the rotation error is suboptimal. This scene consists of just two sequences—one for training and one for testing—potentially limiting the effectiveness of our augmentation in capturing scene variability. Meanwhile, RAP\sub{ref} further reduces localization error below 1 cm with one-shot refinement using our 3DGS. It also surpasses ACE~\cite{brachmann2023accelerated} and its post-refinement variant, ACE + GS-CPR~\cite{liu2025gscpr}. Qualitative examples are shown in Fig.~\ref{fig: 7scenes_vis}.

\begin{table}[t]
    \renewcommand{\arraystretch}{1.15} 
    \caption{\textbf{Ablation study.}}  % We systematically investigate the impact of the proposed components; see Sec.~\ref{sec: ablation} for detailed analysis.
    \label{tab:ablation}
    \vspace{-5pt}
    % \small
    \centering
    \scalebox{0.65}{
    \begin{tabular}{lccc}
    \toprule
    \textbf{Setups}\quad\quad\quad  on \textit{Shop}        & \textbf{Trans.} (cm) $\downarrow$  & \textbf{Rot.} ($^\circ$) $\downarrow$    \\
    \midrule
    \rowcolor{gray!25}\label{case: I}I (Baseline): $\varphi$ = \text{VGG16}  &	174 & 5.45  \\
    \label{case: II}II: $\varphi$ = \text{Efficient-B0}  &	103 & 4.64  \\
    \label{case: III}III: II + Pose Aug.  & 75 & 3.52  \\
    \label{case: IV}IV: III + Appearance Aug.  & 60 & 3.14  \\
    \label{case: V}V: IV + Decoder (\text{ConvNet})  & 52 & 2.51  \\
    \label{case: VI}VI: V + Decoder (\text{Transformer}) &	40 & 1.98 \\
    \rowcolor{green!25} \label{case: VII}\textbf{VII (Ours):}  VI + Discriminator   & \textbf{33}  & \textbf{1.48} \\
    \bottomrule
    \end{tabular}}
    \vspace{-5pt}
\end{table}

\subsection{Ablation Study}
\label{sec: ablation}

We conduct ablation studies on the validation set of \textit{Shop} in the Cambridge Landmarks dataset to investigate the impact of all the components in our RAP. 
Setup \hyperref[case: I]{I}, our baseline, consists of the same components as in PoseNet \citep{kendall2015posenet} and has been retrained for our experiments. In Setup \hyperref[case: II]{II}, we replace the feature extraction from VGG16~\cite{simonyan2014very} to Efficient-B0~\cite{liu2017efficient}, which enhances performance due to its superior feature representation, while they both exhibit poor performance due to the lack of data synthesis.
In Setup \hyperref[case: III]{III} and \hyperref[case: IV]{IV}, we explore the effectiveness of the designed pose augmentation and appearance augmentation, which bring notable improvements: translation error reduces from 103~cm to 75 cm, and rotation error from 3.52$^\circ$ to 3.14$^\circ$. 
In Setup \hyperref[case: V]{V} and \hyperref[case: VI]{VI}, we add regular convolutional layers and Pose Transformer between feature extraction and pose regression. Both improve performance due to the increasing parameters, but the Transformer achieves superior results by effectively handling long-term dependencies through attention mechanisms.
Finally, in Setup \hyperref[case: VII]{VII}, our adversarial discriminator effectively reduces the syn-to-real domain gap, allowing the model to learn better pose regression features from synthetic data and further reduce localization error. 

\begin{figure}[t]
    \centering
    \includegraphics[width=1\linewidth]{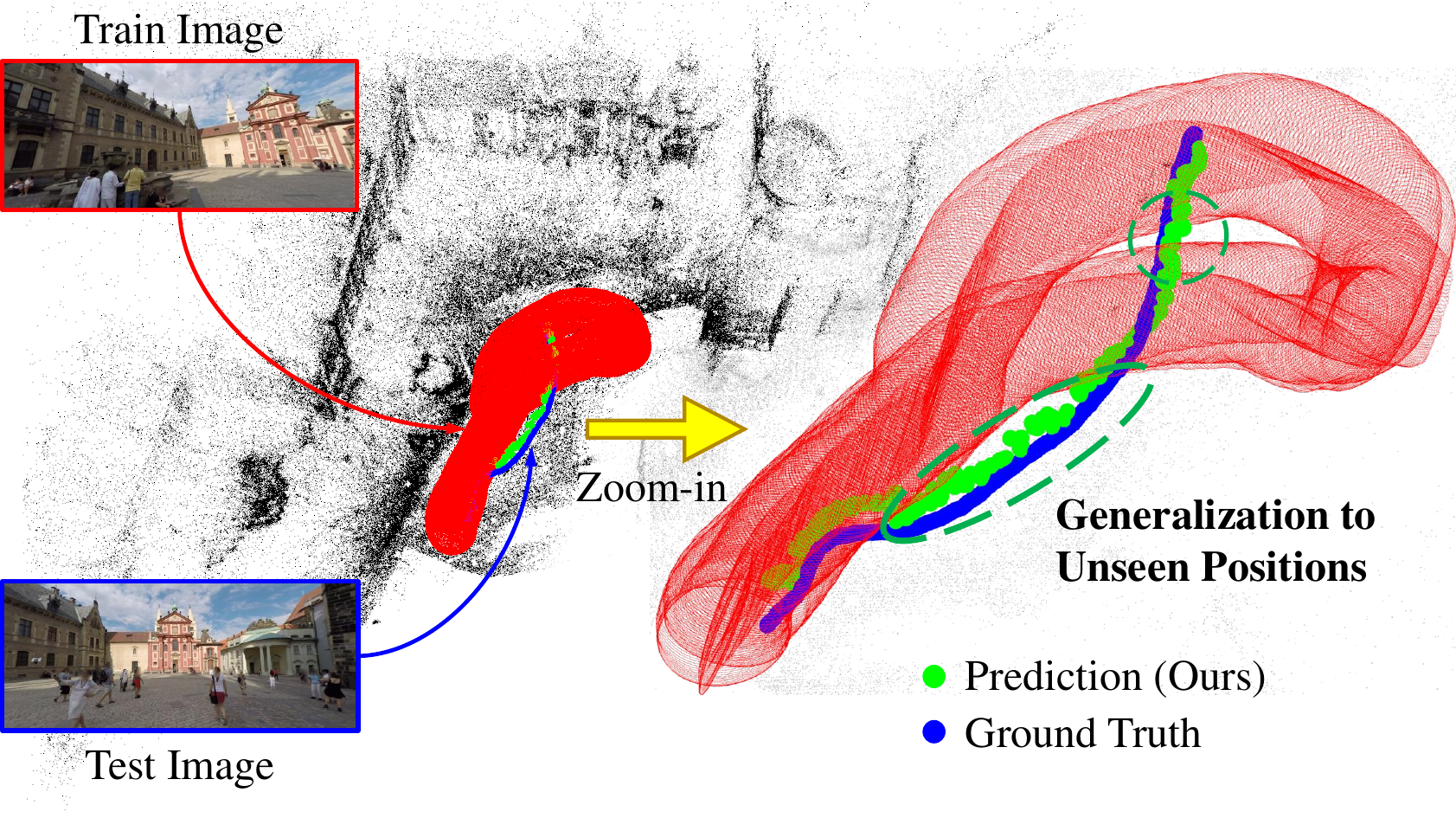}
    \vspace{-20pt}
    \caption{\textbf{Visualization of the training set distribution and results on \textit{ St. George's Basilica Building}~\cite{sattler2019understanding}.} The \textcolor{red}{red hollow spheres}, centered on the real images in the training set, indicate the potential locations of all synthetic images during training.}  %  The \textcolor{blue}{blue dots} and \textcolor{green}{green dots} represent the ground truth and predictions by our RAP, respectively.
    \label{fig: supp_scaling}
    \vspace{-5pt}
\end{figure}

\subsection{Discussion on Data Synthesis}
\noindent\textbf{Emerging Generalizability.} Previously, APR has been understood to implicitly learn image retrieval~\cite{sattler2019understanding}, lacking the ability to successfully interpolate between training samples and generalize beyond them. To investigate how APR training is affected by increasing synthetic data, we trained RAP on the \textit{St. George’s Basilica Building}~\cite{sattler2019understanding} and visualized the results in Fig.~\ref{fig: supp_scaling}. Here, the translation perturbation was set to $\delta t = 350$ cm and the rotation perturbation to $\delta r = 60^\circ$. Notably, the test set contained two regions entirely uncovered by the training set. Despite this, the model still closely predicts the test camera poses, demonstrating generalization ability beyond the original training positions.

We also learn from our experiments that reducing the rotation perturbation, such that the overlap between test and training views remains minimal, leads to high localization error. This is because the translation and rotation parameter space is inherently a $\mathrm{SE}(3)$ manifold. Even if the translation remains fixed, significant rotation changes result in entirely different visual content in the images, naturally preventing the model from estimating poses of such unseen views, which correspond to a large distance on the $\mathrm{SE}(3)$ manifold. Therefore, enabling generalization across a broader range of space is an important direction for future work.

\begin{table}[t]
\caption{\textbf{Exploring the generalization boundaries of the model with synthetic data.} \textcolor{green}{Green}, \textcolor{blue}{blue}, and \textcolor{red}{red} percentages indicate the relative change in localization error (\textbf{Med Err}) compared to the scenario without a void zone.}
\label{tab: margin}
\vspace{-5pt}
\centering
  \renewcommand{\arraystretch}{1.15} % 增加行距
  \setlength{\tabcolsep}{4pt}
  \scalebox{0.65}{
  \begin{tabular}{c|c|ccccc}
    \noalign{\hrule height 1pt}
     \multicolumn{1}{c|}{\multirow{2}{*}{}} & \multicolumn{1}{c|}{\multirow{2}{*}{\textbf{\makecell{w/o Void\\Zone}}}} &\multicolumn{5}{c}{\textbf{w/ Void Zone (cm/$^\circ$)}} \\
     \cline{3-7} &  & \textbf{10/0.5} & \textbf{20/1} & \textbf{30/1.5} & \textbf{50/2} & \textbf{80/2.5}  \\
    \noalign{\hrule height 0.7pt}
    \multicolumn{1}{c|}{\multirow{2}{*}{\makecell{\textbf{Med Err} $\downarrow$ \\ (rel. change)}}}  & 33/1.26 & 30/1.34 & 32/1.32 & 40/1.84 & 39/2.07 & 49/2.20    \\
     & {0\%/0\%} & \textcolor{green}{-9\%/6.3\%} & \textcolor{green}{-3\%/4.7\%} & \textcolor{blue}{21\%/46.0\%} & \textcolor{blue}{18\%/64.3\%} & \textcolor{red}{48\%/74.6\%}   \\
    \hline
    \textbf{Avg Err} $\downarrow$ & 41/1.51 & 38/1.75 & 41/1.63 & 51/2.40  &	48/2.40  & 61/2.84   \\
    \textbf{Max Err} $\downarrow$ & 155/4.52 & 147/6.56 & 219/6.01 & 192/8.38 &	246/9.80 & 242/13.12\\
    \textbf{Min Err} $\downarrow$ & 3/0.17 & 8/0.20 & 4/0.22 & 7/0.30  &	4/0.16  & 8/0.60  \\
    \noalign{\hrule height 1pt}
  \end{tabular}}
  % \vspace{-5pt}
\end{table}

\begin{table}[t]
    \renewcommand{\arraystretch}{1.15} 
    \centering
    \caption{\textbf{Ablation on different pose augmentation policies}.}    
    \vspace{-5pt}
    \scalebox{0.65}{
    \begin{tabular}{c|cccc|c}
        \noalign{\hrule height 1pt}
    \textbf{Methods} & \textbf{\textit{College}} & \textbf{\textit{Hospital}} & \textbf{\textit{Shop}} & \textbf{\textit{Church}} & \textbf{Average}  \\
        \noalign{\hrule height 0.7pt}
    RAP (LENS~\cite{moreau2022lens})          & 73/1.15 & 126/1.87 & 71/3.37 & 128/3.50 & 100/22.47 \\
    \textbf{RAP (Ours)} & \textbf{52}/\textbf{0.90} & \textbf{87}/\textbf{1.21} & \textbf{33/1.48}     & \textbf{53}/\textbf{1.52} & \textbf{56}/\textbf{1.28} \\
        \noalign{\hrule height 1pt}
    \end{tabular}}
    \vspace{-5pt}
    \label{table:policy}
\end{table}

\noindent\textbf{Analyzing Generalization Boundaries.} To evaluate the model's generalizability, we designed an experiment introducing a ``void zone" centered on the test camera, where all real and synthetic data within this zone were excluded. The void zone was progressively expanded to determine the critical threshold at which the localization performance declines most significantly. Specifically, for \textit{Shop}, we used 100\% of the training set to ensure complete scene coverage, with void zone ranges set as [10/0.5, 20/1, 30/1.5, 50/2, 80/2.5, 100/3] (cm/$^\circ$). The results in Table~\ref{tab: margin} demonstrate a stepwise decline in performance. Initially, expanding the void zone has minimal impact on localization accuracy. However, at 30 cm / 1.5°, a sharp decrease in performance marks the model's generalization boundary.

\noindent\textbf{Pose Augmentation Policy.} We conduct experiments using a modified version of RAP, with the pose augmentation approach identical to that of LENS~\cite{moreau2022lens}, as shown in the Table~\ref{table:policy}, where our method obtains superior performance. 
This may be because, although LENS’s pose augmentation policy covers a broader spatial area than the training set, its synthetic data may have lower NVS image quality in many unseen regions, which could negatively affect APR training.

\noindent\textbf{Density of Training Data.} As shown in Table~\ref{tab: quantity}, our method with the proposed augmentation significantly reduces errors as the density of real training data increases from 20\% to 80\%. However, the localization accuracy remains almost unchanged from 80\% to 100\%, as the scene is already sufficiently covered. Notably, using 100\% of the training data without augmentation can result in a significantly higher maximum error in translation, nearly double that with only 20\% of the training data with augmentation, despite its limited spatial coverage. This suggests that our augmentation method successfully prevents overfitting to the training data, improving generalization to the test set. 

\noindent\textbf{Quality of Training Data.} We evaluate the impact of synthetic image quality on model performance in Table~\ref{tab: quality}, using 20\% and 50\% of the real data for pose regression. For \textit{Shop}, it is evident that fewer training samples in 3DGS result in lower-quality rendered views (as indicated by lower PSNR), leading to suboptimal localization performance, particularly for rotation. Surprisingly, localization performance using only 20\% of the data for training suboptimal 3DGS and pose regression surpasses the results obtained with 100\% of the data without augmentation, as shown in Table~\ref{tab: quantity}.
This experiment confirms the need for a robust NVS model and the proposed augmentation method in APR training.

\noindent\textbf{Different 3D Representations.} We evaluate the impact of different 3D representations on the performance in Table~\ref{table:nerfvs3dgs}. Trivially replacing NeRF with 3DGS in existing frameworks degrades performance due to 3DGS's inferior 3D consistency. This shows the effectiveness of our proposed joint training paradigm in RAP, which better utilizes diverse synthetic data to learn appearance-invariant features, rather than naively transferring from NeRF to 3DGS.

%% file: sec/6_Conclusion.tex
\section{Conclusion}

\begin{table}[t]
\caption{\textbf{Impact of the density of real training data.} Our augmentation improves the model's ability to generalize across the entire scene, although this effect has an upper limit.}
\label{tab: quantity}
\vspace{-5pt}
\centering
  \renewcommand{\arraystretch}{1.15} % 增加行距
  \setlength{\tabcolsep}{4pt}
  \scalebox{0.62}{
  \begin{tabular}{l|ccccc|cc}
    \noalign{\hrule height 1pt}
     \multicolumn{1}{c|}{\multirow{2}{*}{\textbf{\makecell{Training\\Pose \%}}}}  & \multicolumn{5}{c|}{\textbf{w/ Appearance \& Pose Aug. (cm/$^\circ$)}} &\multicolumn{2}{c}{\textbf{w/o Aug. (cm/$^\circ$)}} \\
     \cline{2-8} & \textbf{100\%} & \textbf{80\%} & \textbf{60\%} & \textbf{40\%} & \textbf{20\%} & \textbf{100\%} & \textbf{50\%} \\
    \noalign{\hrule height 0.7pt}
    \textbf{Med Err} $\downarrow$ & \underline{33}/\textbf{1.26} & \textbf{32}/\underline{1.27} & 37/1.90 & 57/2.23  & 87/3.65  & 98/3.75  & 104/4.17  \\
    \textbf{Avg Err} $\downarrow$ & \underline{41}/\underline{1.51} & \textbf{40}/\textbf{1.50} & 48/2.17 & 62/2.81  &	91/4.65  & 128/4.49  & 139/5.33  \\
    \textbf{Max Err} $\downarrow$ & \textbf{155}/\underline{4.52} & \underline{158}/\textbf{4.09} & 193/9.39 & 230/11.06 &	231/15.45 & 490/20.73 & 500/21.02 \\
    \textbf{Min Err} $\downarrow$ & \textbf{3}/\textbf{0.17} & 7/0.20 & 7/\underline{0.18} & \underline{6}/0.38  &	12/0.46  & 13/0.63  & 9/0.48  \\
    \noalign{\hrule height 1pt}
  \end{tabular}}
    % \vspace{-5pt}
\end{table}

\begin{table}[t]
\caption{\textbf{Impact of synthetic image quality.} Training with higher-quality synthetic images from advanced NVS models enhances localization performance.}
\label{tab: quality}
\vspace{-5pt}
\centering
  \renewcommand{\arraystretch}{1.15} % 增加行距
  \setlength{\tabcolsep}{5pt}
  \scalebox{0.64}{
  \begin{tabular}{ccc|ccccc}
    \noalign{\hrule height 1pt}
     \multicolumn{3}{c|}{\textbf{3DGS Performance}} & \multicolumn{5}{c}{\textbf{Localization Performance (cm/$^\circ$)}} \\
     \hline
     \makecell{\textbf{\% Images}\\\textbf{(Train)}} & \makecell{\textbf{PSNR} $\uparrow$\\\textbf{(Train)}} & \makecell{\textbf{PSNR} $\uparrow$\\\textbf{(Test)}} & \makecell{\textbf{\% Images}\\\textbf{(Train)}} & \makecell{\textbf{Med}\\\textbf{Err}}$\downarrow$ & \makecell{\textbf{Avg}\\\textbf{Err}}$\downarrow$ & \makecell{\textbf{Max}\\\textbf{Err}}$\downarrow$ & \makecell{\textbf{Min}\\\textbf{Err}}$\downarrow$  \\

    \noalign{\hrule height 0.7pt}
    \textbf{20\%}  & 29.08 & 15.98 & \textbf{20\%}  & 58/3.59 & 68/4.31 & 211/12.19 & 14/0.51 \\
    \textbf{20\%}  & 29.08 & 15.98 & \textbf{50\%}  & 43/2.47 & 55/3.26 & 196/21.11 & 7/0.40 \\
    \textbf{50\%}  & 26.88 & 17.55 & \textbf{50\%}  & 37/1.88 & 48/2.37 & 184/10.17 & 9/0.52 \\
    \textbf{100\%} & 24.60 & 18.30 & \textbf{50\%}  & 35/1.64 & 41/2.12 & 130/11.07 & 4/0.38 \\
    \noalign{\hrule height 1pt}
  \end{tabular}}
  % \vspace{-5pt}
\end{table}

\begin{table}[t]
    % \vspace{-8pt}
    \renewcommand{\arraystretch}{1.15} 
    \centering
    \caption{\textbf{Ablation on different 3D representations.}}
    \vspace{-5pt}
    \scalebox{0.65}{
    \begin{tabular}{c|cccc|c}
        \noalign{\hrule height 1pt}
     \textbf{Methods} & \textbf{\textit{College}} & \textbf{\textit{Hospital}} & \textbf{\textit{Shop}} & \textbf{\textit{Church}} & \textbf{Average}  \\
        \noalign{\hrule height 0.7pt}
    DFNet~\cite{chen2022dfnet} & 73/2.37 & 200/2.98 & 67/2.21 & 137/4.03 & 119/2.90 \\ 
    DFNet$_\text{GS}$ & 102/2.31 & 137/8.08 & 77/3.92 & 123/4.68 & 110/4.75 \\ 
    \textbf{RAP (Ours)} & \textbf{52}/\textbf{0.90} & \textbf{87}/\textbf{1.21} & \textbf{33/1.48}     & \textbf{53}/\textbf{1.52} & \textbf{56}/\textbf{1.28} \\
        \noalign{\hrule height 1pt}
    \end{tabular}}
    \vspace{-5pt}
    \label{table:nerfvs3dgs}
\end{table}

\noindent\textbf{Summary.} We address absolute pose regression with a robust two-branch joint training framework based on Transformer, coupled with an efficient data synthesis pipeline leveraging 3D Gaussian Splats (3DGS) to synthesize numerous posed images with diverse appearances as additional supervision. Our RAP achieves state-of-the-art localization performance, even under challenging appearance variations. Moreover, we thoroughly investigate the impact of synthesizing diverse data and present a novel perspective on APR: generalizability can emerge if the learning gap in APR is effectively addressed together with diverse data. We believe our RAP could be a promising starting point, and the experiments presented in the paper can provide useful insights for future research in this field.

%% file: sec/X_suppl.tex
\clearpage
\setcounter{page}{1}
% \maketitlesupplementary

\section*{Appendix}
\label{sec:appendix}
\renewcommand{\thesection}{\Alph{section}}
\renewcommand{\thefigure}{\Roman{figure}}
\renewcommand{\thetable}{\Roman{table}}

\setcounter{section}{0}
\setcounter{figure}{0}
\setcounter{table}{0}

\section{Pipeline Workflow} 

\begin{itemize}
    \item \textbf{Stage 1: Appearance-Varying 3DGS}  % (Sec.~\ref{Pre-Processing with 3DGS})}
    \begin{itemize}
        \item \textbf{Input:} Sequences of RGB images $\bs{I}$ and corresponding camera poses $\bs{P}$.
        \item \textbf{Output:} 3D appearance-varying Gaussians with deblurring ability.
        \item \textbf{Loss:} $\mathcal{L}_1$, $\mathcal{L}_\text{D-SSIM}$, $\mathcal{L}_\text{LPIPS}$, $\mathcal{L}_{\bs{\mathcal{S}}}$, and optionally, $\mathcal{L}_\text{depth}$.
    \end{itemize}

    \item \textbf{Stage 2: Two-Branch Joint APR Training}  % (Sec.~\ref{Architecture of Pose Regressor})}
    \begin{itemize}
        \item \textbf{Branch-1}
        \begin{itemize}
            \item \textbf{Input:} Real image-pose pair \((\bs{I}, \bs{P})\) and the corresponding synthesized image with the same pose \((\bs{I'}, \bs{P})\). 
            
            \item \textbf{Output:} Adjusted translation features $(\mathrm{Adj}(\mathcal{F}_{t}(\bs{I}),$ $\mathrm{Adj'}(\mathcal{F}_{t}(\bs{I'})))$, adjusted rotation features $(\mathrm{Adj}(\mathcal{F}_{r}(\bs{I})), \mathrm{Adj'}(\mathcal{F}_{r}(\bs{I'})))$, and predicted poses \((\widehat{\bs{P}}, \widehat{\bs{P}'})\)
            \item \textbf{Loss:} Pose loss \(\mathcal{L}_\text{pose}^1\), generator loss \(\mathcal{L}_\text{Gen}\), and adversarial loss \(\mathcal{L}_\text{Dis}\).
        \end{itemize}
    
        \item \textbf{Branch-2}
        \begin{itemize}
            \item \textbf{Input:} Synthesized images with randomly blended appearances and perturbed poses \((\bs{I}_\text{syn}, \bs{P_\text{syn}})\).
            \item \textbf{Output:} Predicted poses \(\widehat{P_\text{syn}}\).
            \item \textbf{Loss:} Pose loss \(\mathcal{L}_\text{pose}^2\).
        \end{itemize}        
    \end{itemize}

    \item \textbf{Stage-3: Post-Refinement}  % (Sec.~\ref{Evaluation Setup})}
    \begin{itemize}
        \item \textbf{Input:} Rendered image from 3DGS using the query image and the initial pose estimated by the trained pose regressor.
        \item \textbf{Output:} Final refined pose.
        \item \textbf{Loss:} RANSAC-PnP~\cite{fischler1981random, gao2003complete} solver on pixel-level matching between the rendered and query images.
    \end{itemize}
\end{itemize}

\section{3D Gaussian Splatting Preliminary} 
\label{supp: 3D Gaussian Splatting Preliminary}
Gaussian Splatting~\cite{kerbl20233d} is a promising approach for real-time novel view synthesis. By representing scenes as a set of 3D Gaussians, it retains the differentiable properties of volumetric radiance fields while offering more efficient optimization and higher-quality rendering compared to NeRF.  The scene is defined through parameters such as position $\bs{\mu} \in \mathbb{R} ^ {K \times 3}$, covariance decomposed as rotation $\bs{q} \in \mathbb{R} ^ {K \times 4}$ and scaling $\bs{s} \in \mathbb{R} ^ {K \times 3}$, anisotropic color $\bs{c} \in \mathbb{R} ^ {K \times 3}$ modeled by sphere harmonics $\bs{\mathcal{Y}} \in \mathbb{R} ^ {K \times 16 \times 3}$, and opacity $\bs{\alpha} \in \mathbb{R} ^ K$. During optimization, the scene representation is optimized by iteratively adjusting parameters through stochastic gradient descent, enabled by a differentiable rasterizer. This process is combined with adaptive density control to dynamically add or remove Gaussians based on the gradient of screen-space points corresponding to the Gaussians and opacity reset to reduce overfitting caused by floaters. The rendering process involves projecting 3D Gaussians onto the 2D image plane, sorting them by depth, and then applying $\alpha$-blending to generate the final image. The render equation is:
\begin{equation}
\bs{C} = \sum_{i=1}^{N} T_i \alpha_i \bs{c}_i, \quad T_i = \prod_{j=1}^{i-1} (1 - \alpha_j),
\end{equation}
where $\bs{C} \in \mathbb{R} ^ 3$ is each pixel's color and $T_i$ is the transmittance. This approach significantly speeds up optimizing and rendering while achieving state-of-the-art visual quality.  

\begin{figure*}[t]
    \centering
    \includegraphics[width=1\linewidth]{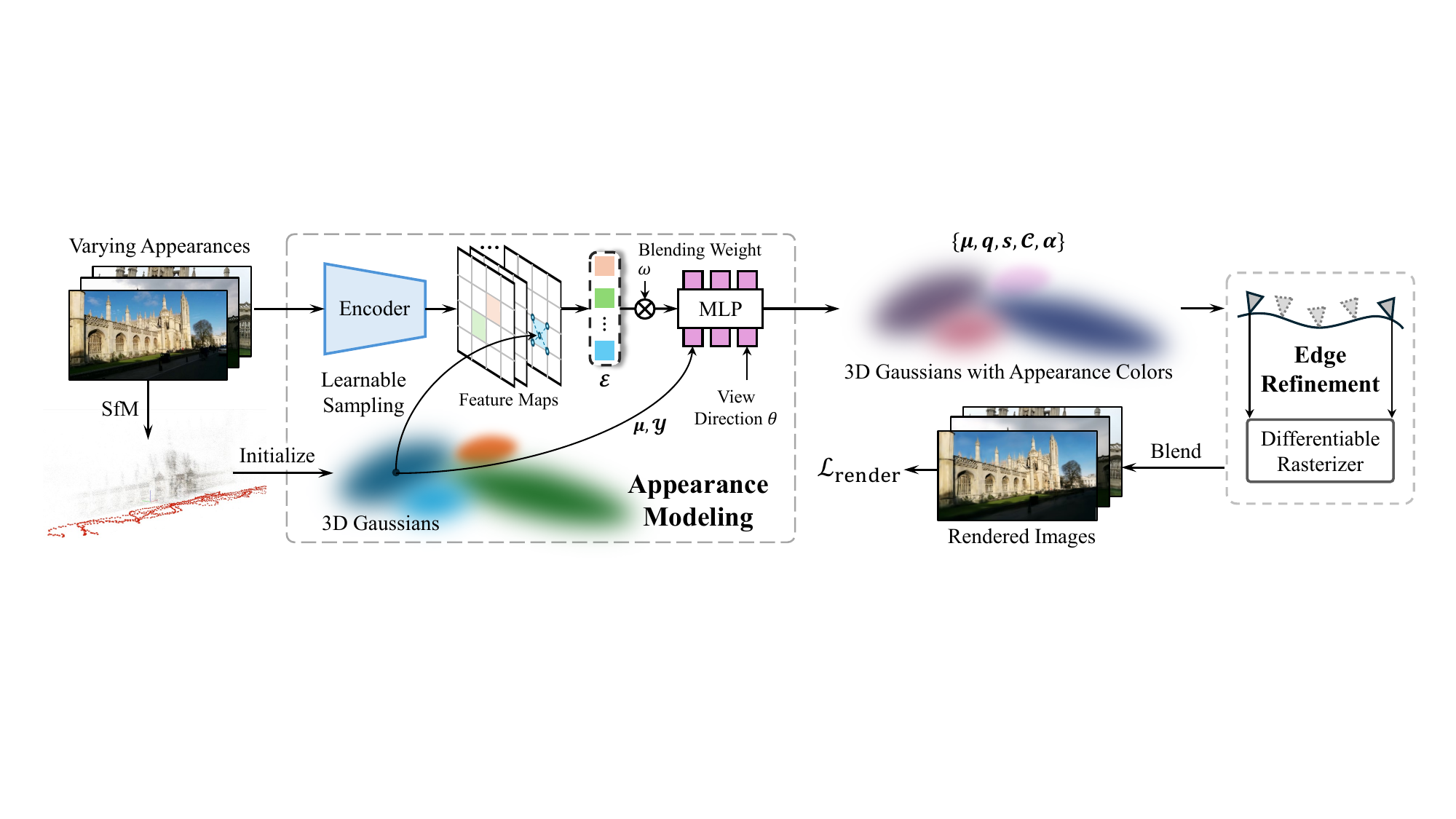}
    \caption{\textbf{Overall illustration of appearance-varying 3DGS.} The framework models varying appearances using 3D Gaussians enhanced with appearance colors. It initializes 3D Gaussians from SfM data, refines their appearance by learnable sampling and blending weights computed via an encoder and MLP, and renders images by a differentiable rasterizer with edge refinement to minimize the rendering loss.}
    \label{fig: appearance-3dgs}
\end{figure*}

\section{MASt3R Preliminary} 
\label{MASt3R Preliminary.}
This section further elaborates on the background knowledge of MASt3R~\cite{leroy2024grounding} mentioned in the paper.  % Sec.~\ref{Implementation}.
MASt3R grounds image matching tasks in 3D space to improve robustness and accuracy in challenging scenarios. Building on the DUSt3R~\cite{wang2024dust3r} framework, MASt3R incorporates a new feature-matching head and a fast reciprocal matching algorithm, significantly enhancing performance for dense correspondences and camera pose estimation. It addresses the limitations of traditional 2D-based methods by leveraging dense 3D pointmaps and a coarse-to-fine matching strategy. Extensive evaluations demonstrate substantial gains in accuracy, computational efficiency, and generalizability, making MASt3R a robust solution for visual localization tasks.  

\section{Architecture Details}
This section provides additional details regarding the network structure of RAP.

\subsection{Feature Extraction}
Our RAP pipeline first downsamples the input real images, adjusting the shorter side to approximately 240–360~px to enhance computational efficiency without losing much information. The downsampled images are then normalized and passed through the backbone network. For feature extraction, we utilize EfficientNet-B0~\cite{liu2017efficient} as the backbone for multi-scale feature extraction. Translation feature $\mathcal{F}_t$ and rotation feature $\mathcal{F}_r$ are extracted from the third (\texttt{reduction\_3}) and fourth (\texttt{reduction\_4}) layers, with the number of feature channels being \(C_t = 40\) and \(C_r = 112\), which then are projected via $1\times1$ convolutions to align with the input channel dimension \(D = 128\) of the proposed \textit{Pose Transformer}.

\begin{figure}[t]
    \centering
    \includegraphics[width=1\linewidth]{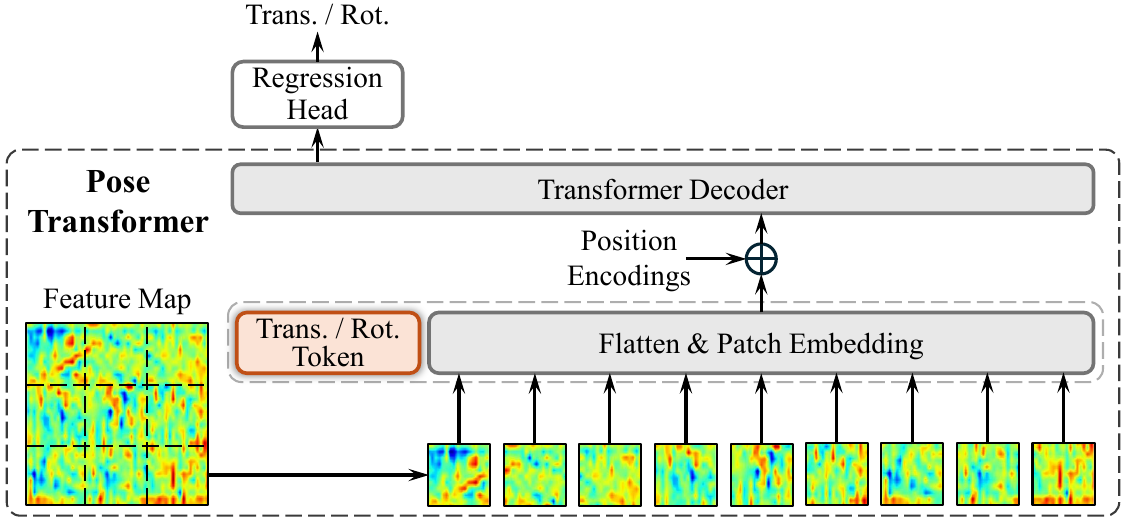}
    % \captionsetup{font=small}
    \caption{\textbf{Structure of the Pose Transformer.} The feature map from the backbone is used as input, where we replace the \texttt{CLS} token in Vision Transformers (ViT)~\cite{dosovitskiy2020image} with translation token \texttt{Trans} or rotation token \texttt{Rot} for the following regression head.}
    \label{fig: supp_transformer}
\end{figure}

\subsection{Pose Transformer}
\label{supp:PoseTransformer}
Relying on fine-grained local features, as done in previous works~\cite{chen2022dfnet, lin2024learning}, can hinder invariant feature learning due to image noise caused by dynamic objects and illumination changes. To overcome this, we leverage Transformer's robust ability to capture long-range dependencies, as illustrated in Fig.~\ref{fig: supp_transformer}.
Taking the Cambridge Landmarks dataset~\cite{kendall2015posenet} as an example, the original image resolution is $854 \times 480$. After downsampling and feature extraction, the resulting \textit{translation feature map} (identical for the rotation feature map) has a shape of $[B, 112, H_t, W_t]$, where $H_t = 15$ and $W_t = 27$. A $1\times1$ convolutional layer is then applied to adjust the number of feature channels to 128, aligning it with the input dimension of the Transformer architecture.
The \textit{translation token} is a learnable parameter with 128 dimensions. The translation feature map is flattened and concatenated with the translation token, forming a matrix of size $[B, 128, H_t \times W_t + 1]$. This matrix, combined with positional encodings, is fed into the Transformer decoder, which consists of six layers, each containing multi-head self-attention with eight heads. Finally, the dimension corresponding to the translation token is extracted and passed to the regression head to predict the translation. For clarity, we describe the process using translation as an example, but the same approach is applied to rotation.  

\subsection{Adversarial Discriminator}
\label{supp: Adversarial Discriminator}
We address the domain gap between synthetic and real images at the feature level by employing an adversarial discriminator. Specifically, the translation features $\mathcal{F}_t(\bs{I})$ and rotation features $\mathcal{F}_r(\bs{I})$ are first processed through the adjustment layers composed of two Conv-ReLU-BN layers, respectively, as $\mathrm{Adj}(\mathcal{F}_{t}(\bs{I}))$ and  $\mathrm{Adj'}(\mathcal{F}_{t}(\bs{I'}))$, which align the channel dimensions to a consistent size of 128. The discriminator, implemented as a sequence of four convolutional layers with LeakyReLU activations and dropout, progressively reduces the spatial dimensions. The output is flattened and passed through a fully connected layer tailored for different datasets. Meanwhile, the MSE loss is applied in the discriminator, bridging the feature-level domain gap by effective adversarial learning.

\begin{table*}[t]
    \caption{\textbf{Metadata showing the number of images in the training and test sets for each scene.} The number of image sequences in each scene is indicated in parentheses. Different appearances across image sequences collected at different times pose challenges to modeling the environment and performing visual localization. More visualization can be found in Fig.~\ref{fig: supp_7scenes_gs}, Fig.~\ref{fig: supple_render_compare_cam}, and Fig.~\ref{fig: supple_render_compare_mars}.}
    \label{table: Benchmarks}
    \vspace{-5pt}
    \renewcommand{\arraystretch}{1.15} % 增加行距
    \centering
    \scalebox{0.8}{
    \begin{tabular}{c|c|ccccccc|c}
    \noalign{\hrule height 1pt}
    \multirow{3}{*}{\textbf{7-Scenes~\cite{shotton2013scene}}} & \textbf{Scenes} & \textbf{\textit{Chess}} (6) & \textbf{\textit{Fire}} (4) & \textbf{\textit{Heads}} (2) & \textbf{\textit{Office}} (10) & \textbf{\textit{Pumpkin}} (8) & \textbf{\textit{Kitchen}} (14) & \textbf{\textit{Stairs}} (6) & \textbf{Total}  \\
    \cline{2-10}
    & Train            & 4000 & 2000 & 1000 & 6000 & 4000 & 7000 & 2000 & 26000 \\
    & Test          & 2000 & 2000 & 1000 & 4000 & 2000 & 5000 & 1000 & 17000 \\
    \noalign{\hrule height 0.7pt}
    \multirow{3}{*}{\textbf{Cambridge~\cite{kendall2015posenet}}} & \textbf{Scenes} & 
    \textbf{\textit{College}} (8) & \textbf{\textit{Hospital}} (9) & \textbf{\textit{Church}} (14) & \textbf{\textit{Shop}} (3) &
    - & - & - & \textbf{Total} \\
    \cline{2-10}
    & Train            & 1220 & 895 & 1487 & 231 & - & - & - & 3833 \\
    & Test          & 343 & 182 & 530 & 103 & - & - & - & 1158 \\
    \noalign{\hrule height 0.7pt}
    \multirow{3}{*}{\textbf{MARS~\cite{li2024multiagent}}} & \textbf{Scenes} & \textbf{\textit{``0011"}} (9) & \textbf{\textit{``0015"}} (5) & \textbf{\textit{``0037"}} (5) & \textbf{\textit{``0041"}} (5) & - & - & - & \textbf{Total} \\
    \cline{2-10}
    & Train            & 792 & 788 & 771 & 819 & - & - & - & 3170 \\
    & Test          & 186 & 172 & 225 & 204 & - & - & - & 787 \\
    \noalign{\hrule height 1pt}
    \end{tabular}}
    \vspace{-5pt}
\end{table*}

\subsection{Regression Head}
The output features from the Transformer are fed into dedicated regression heads. For translation, the regressor outputs a 3-dimensional vector representing \([x, y, z]\) coordinates. For rotation, the regressor outputs a 6-dimensional vector, which is a continuous representation of a rotation matrix~\cite{zhou2019continuity}, which is subsequently converted into a \(3 \times 3\) rotation matrix with Gram-Schmidt orthogonalization~\cite{bjorck1967solving}.

\section{Implementation Details}
\label{Implementation Details}
\subsection{Appearance-Varying 3DGS}
\label{supp: Appearance-Varying 3DGS}
The detailed pipeline of our Appearance-varying 3DGS is demonstrated in Fig.~\ref{fig: appearance-3dgs}. For challenging datasets with moving objects and camera motion blur, we extend the optimization iterations to 90,000 and adjust densification parameters and pruning behaviors according to the scene’s size and complexity. For 7-Scenes~\cite{shotton2013scene}, we use the provided depth information to regularize 3DGS. We sample scenes that need edge refinement twice when optimizing each frame. The loss $\mathcal{L}$ is implemented as:  
\begin{equation}
\mathcal{L} = \gamma_1\mathcal{L}_1 + \gamma_2\mathcal{L}_\text{D-SSIM} + \gamma_3\mathcal{L}_\text{LPIPS} + \gamma_4\mathcal{L}_{\mathcal{S}} + \gamma_5\mathcal{L}_\text{depth},
\label{loss: 3dgs}
\end{equation}
where $\gamma_1=0.8$, $\gamma_2=0.2$, $\gamma_3=0.005$, $\gamma_4=0.001$, and $\gamma_5$ is decayed from 1 to 0.01 if depth regularization is enabled; otherwise, $\gamma_5 = 0$. $\bs{\mathcal{S}}$ is the learnable sampler mentioned in the paper. $\mathcal{L}_{\bs{\mathcal{S}}}$ is computed as:  % Sec.~\ref{Pre-Processing with 3DGS}
\begin{equation}
\mathcal{L}_{\bs{\mathcal{S}}} = \frac{1}{n}\sum\operatorname{ReLU}(|\bs{\mathcal{S}}| - 1).
\end{equation}

\subsection{Two-Branch Joint APR Training}  
\label{supp: Two-Branch Joint APR Training}

The RAP is trained and tested with a batch size of 8--12 on NVIDIA RTX A6000 GPUs. To optimize training and save time, we employ an early stopping mechanism with a patience value of 200, and enable FP16 auto-mixed precision (AMP). The learning rate is reduced by a factor of 0.95 whenever the validation loss plateaus, with this adjustment made every 50 epochs. The loss weights used during training are $\beta_1 = 1, \beta_2 = 1, \beta_3 = 0.7$. We also include an additional VICReg loss~\cite{bardes2022vicreg} with the same weight with $\beta_1$ to mitigate the domain gap between the synthetic and real data. For every $N = 20$ epoch, we randomly generate the same number of views as the training sample size using our appearance and pose augmentations. The model generally converges after approximately 1000 epochs.

For scenes where the camera pose is close to surrounding objects, to reduce interference from augmented poses moving inside 3D Gaussian Splats and rendering low-quality images, while also avoiding manual adjustment of augmentation intensity, we filter out generated images with a BRISQUE~\cite{BRISQUE} quality score $\ge$ 50 during augmentation in these scenes. However, this is not always effective and often weakens the augmentation effect. Enabling the augmentation policy to become learnable might be an interesting direction for future work.

\subsection{Matching-Based Post Refinement}
\label{supp: Matching-Based Post Refinement}

We only use MASt3R's coarse mode to obtain 2D-2D matches between the rendered RGB image and the query image to save time. Then, we back-project the rendered depth map from 3DGS into 3D space. In the following RANSAC-PnP~\cite{fischler1981random, gao2003complete}, we set the projection error to be 2 pixels. All other settings follow the defaults provided in the MASt3R repository~\cite{leroy2024grounding}.

\section{Scene Metadata and Evaluation Metrics}
\label{supp: metric}
All the benchmark metadata are shown in Table~\ref{table: Benchmarks}.
We use a widely accepted metric to assess and compare the localization performance of various methods: the median error in translation and rotation, defined as $a$ cm and $b^\circ$, respectively. In the main manuscript, we also report the mean, maximum, and minimum errors to statistically compare the performance distribution across different methods.  

\begin{figure}[t]
    \centering
    \includegraphics[width=\linewidth]{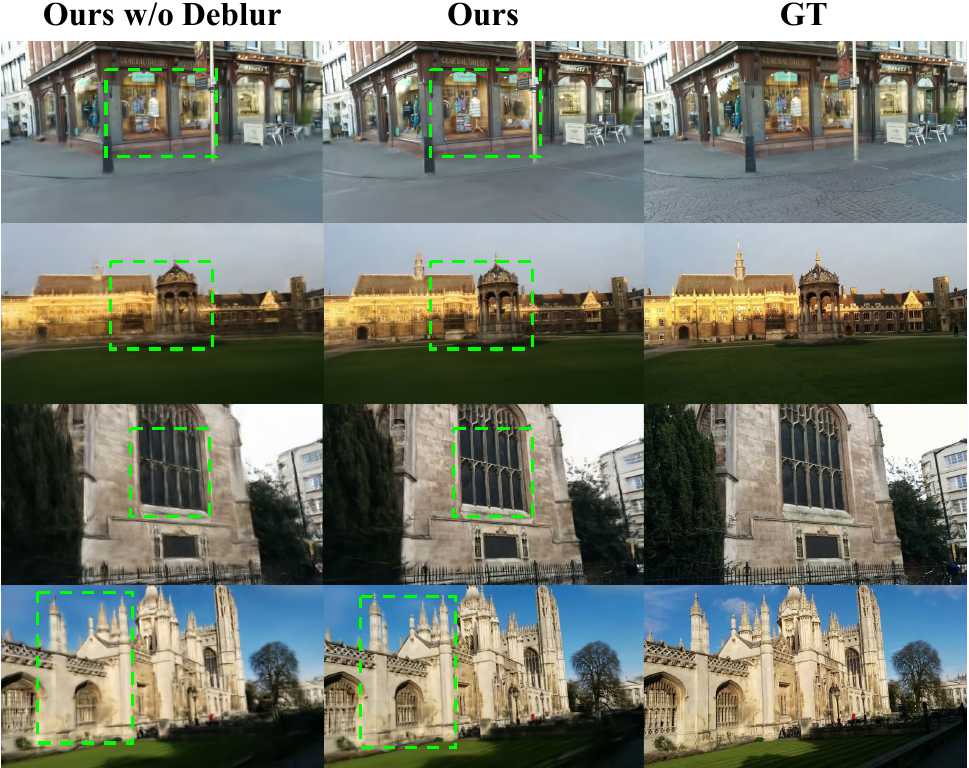}
    \vspace{-15pt}
    \caption{\textbf{Effectiveness of deblurring.} The images in the second column, generated by 3DGS with deblurring ability, exhibit clearer and sharper edges than those produced without deblurring.}
    \label{fig: supp_cam_deblur}
    \vspace{-5pt}
\end{figure}

\begin{figure*}[t]
    \centering
    \includegraphics[width=\linewidth]{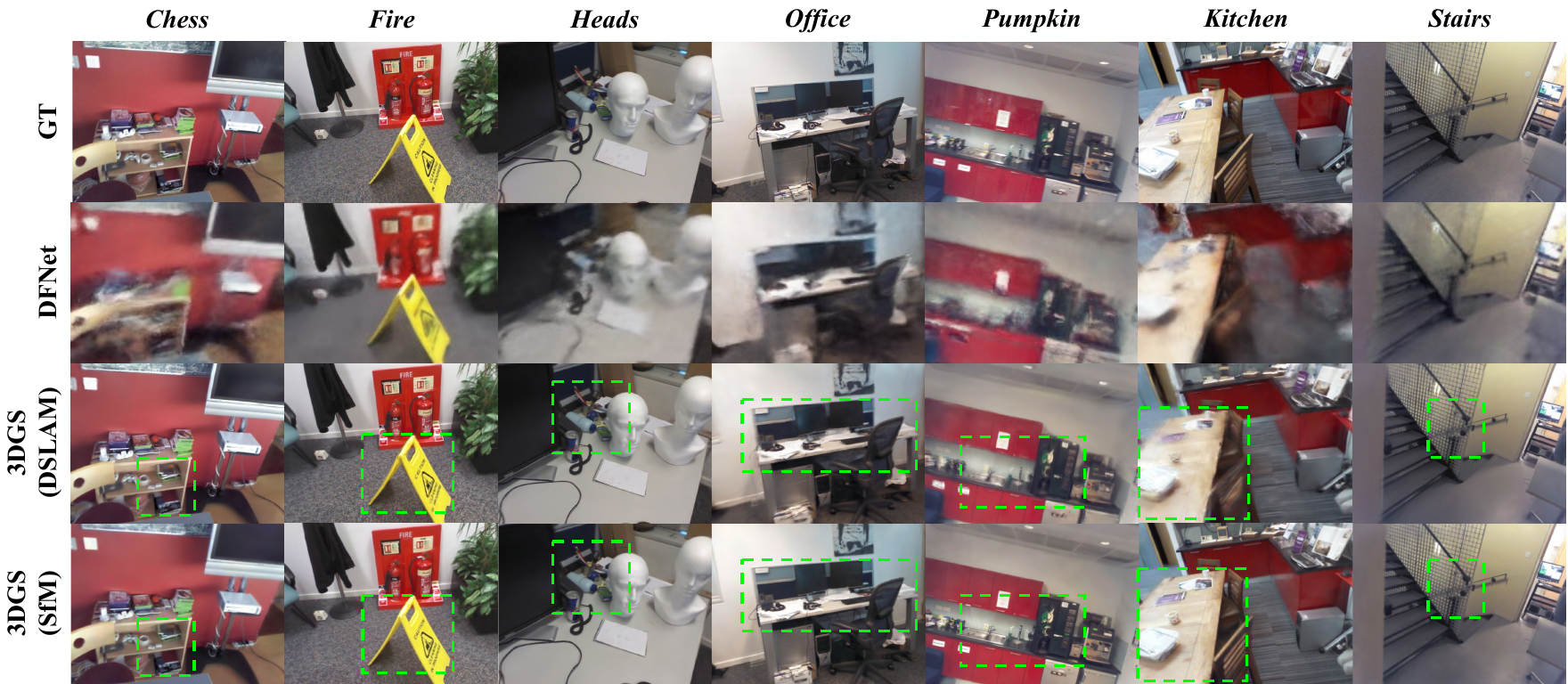}
    \caption{\textbf{Qualitative comparison of different NVS settings.} Our 3DGS achieves the highest visual quality with SfM ground truth poses, whereas DSLAM poses introduce noticeable blurriness, and the NeRF-based method delivers the worst results.}
    \label{fig: supp_7scenes_gs}
\end{figure*}

\section{Cambridge Landmarks~\cite{kendall2015posenet}}

\subsection{Effectiveness of Deblurring}
Visual localization benchmarks are typically collected from video sequences, where motion blur between adjacent frames is inevitable. This negatively affects both the optimization of 3DGS and localization performance. To address this, we incorporate a deblurring module when optimizing 3DGS to mitigate these effects. As shown in Fig.~\ref{fig: supp_cam_deblur}, the deblurring module enhances modeling object edge details and removes artifacts, resulting in higher-quality data synthesis for APR training and post refinement.  %  mentioned in Sec.~\ref{Pre-Processing with 3DGS},

\subsection{3DGS with Controllable Appearances} 
\begin{figure*}[t]
    \centering
    \includegraphics[width=1\linewidth]{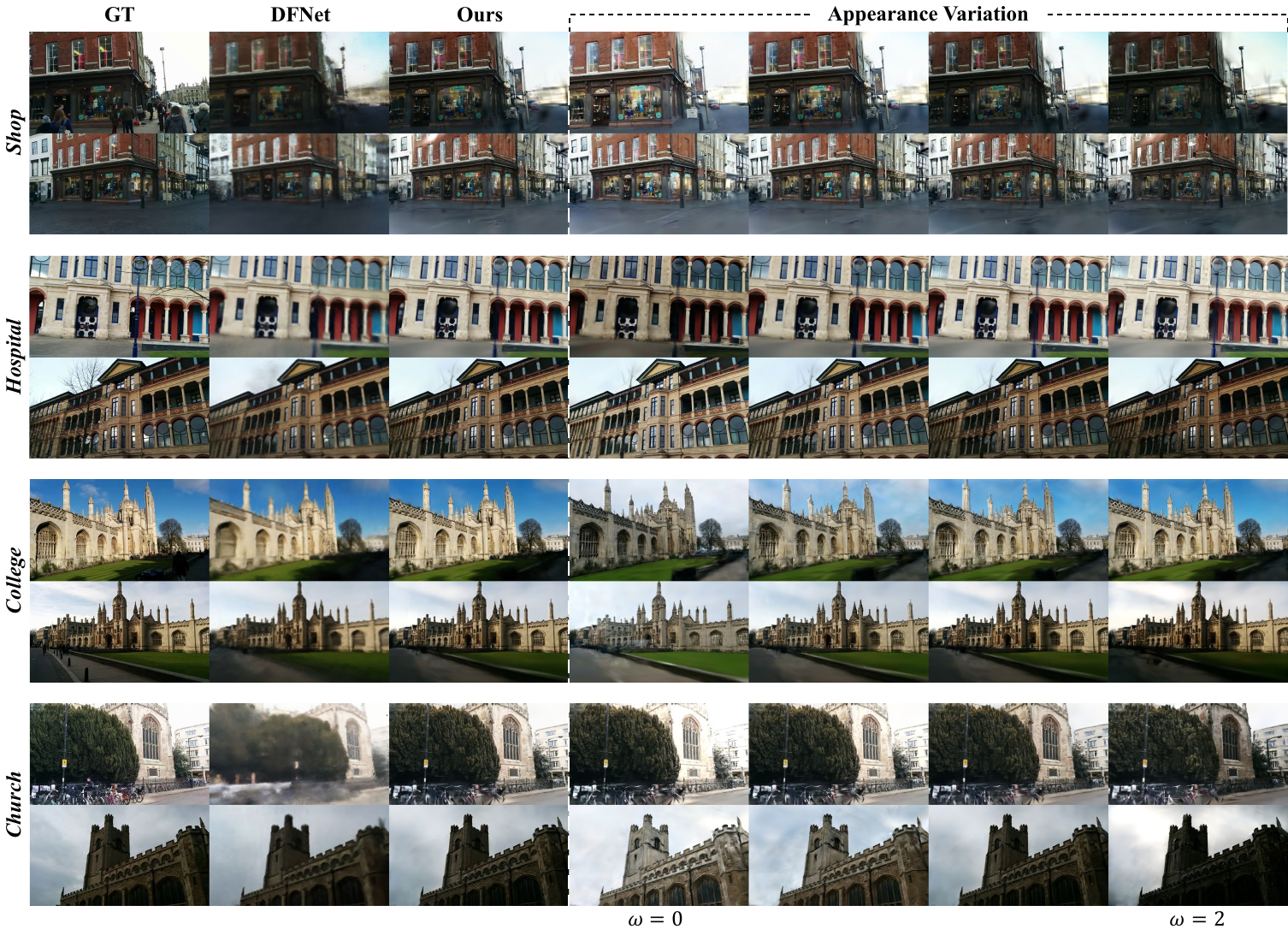}
    \vspace{-20pt}
    \caption{\textbf{Synthetic images with varying appearances on the Cambridge Landmarks dataset~\cite{kendall2015posenet}.} The appearances of synthetic images can be arbitrarily generated using different blending weights $\omega$, ranging from 0 to 2.}
    \label{fig: supple_render_compare_cam}
    \vspace{-12pt}
\end{figure*}

Figure~\ref{fig: supple_render_compare_cam} presents images synthesized by our appearance-varying 3DGS on the Cambridge Landmarks dataset. The images rendered by our 3DGS exhibit finer details, such as sharper edges and textures, compared to those produced by DFNet~\cite{chen2022dfnet}, a NeRF-based method. These improvements contribute to better performance in both translation and rotation regression. Additionally, our 3DGS can model environments with varying lighting conditions using multiple image sequences, enabling seamless interpolation between them. This allows for synthesizing more diverse images, aiding RAP in learning robust and invariant features and further enhancing pose regression performance. 

\begin{table}[t]
    \caption{\textbf{Quantitative comparison of image quality between 3DGS using DSLAM~\cite{newcombe2011kinectfusion} and SfM~\cite{schonberger2016structure} poses.} SfM poses produce more realistic synthetic images with better consistency, as indicated by higher PSNR values that reflect higher image quality.}
    \label{table:PSNR}
    \renewcommand{\arraystretch}{1.15}
    \centering
    \scalebox{0.8}{
    \setlength{\tabcolsep}{4pt}{
    \begin{tabular}{c|ccccccc}
    \noalign{\hrule height 1pt}
    \textbf{PSNR} $\uparrow$ & \textbf{\textit{Chess}}  & \textbf{\textit{Fire}}  & \textbf{\textit{Heads}} & \textbf{\textit{Office}} & \textbf{\textit{Pumpkin}}  & \textbf{\textit{Kitchen}}  & \textbf{\textit{Stairs}}   \\
    \hline
    \textbf{DSLAM}           & 19.98 & 19.04 & 17.23 & 20.89 & 19.20 & 18.92 & 18.93  \\
    \textbf{SfM}             & \textbf{26.52} & \textbf{24.79} & \textbf{20.51} & \textbf{26.35} & \textbf{24.87} & \textbf{24.66} & \textbf{22.98}  \\
    \noalign{\hrule height 1pt}
    \end{tabular}}}
\end{table}

\begin{figure*}[t]
    \centering
    \includegraphics[width=0.9\linewidth]{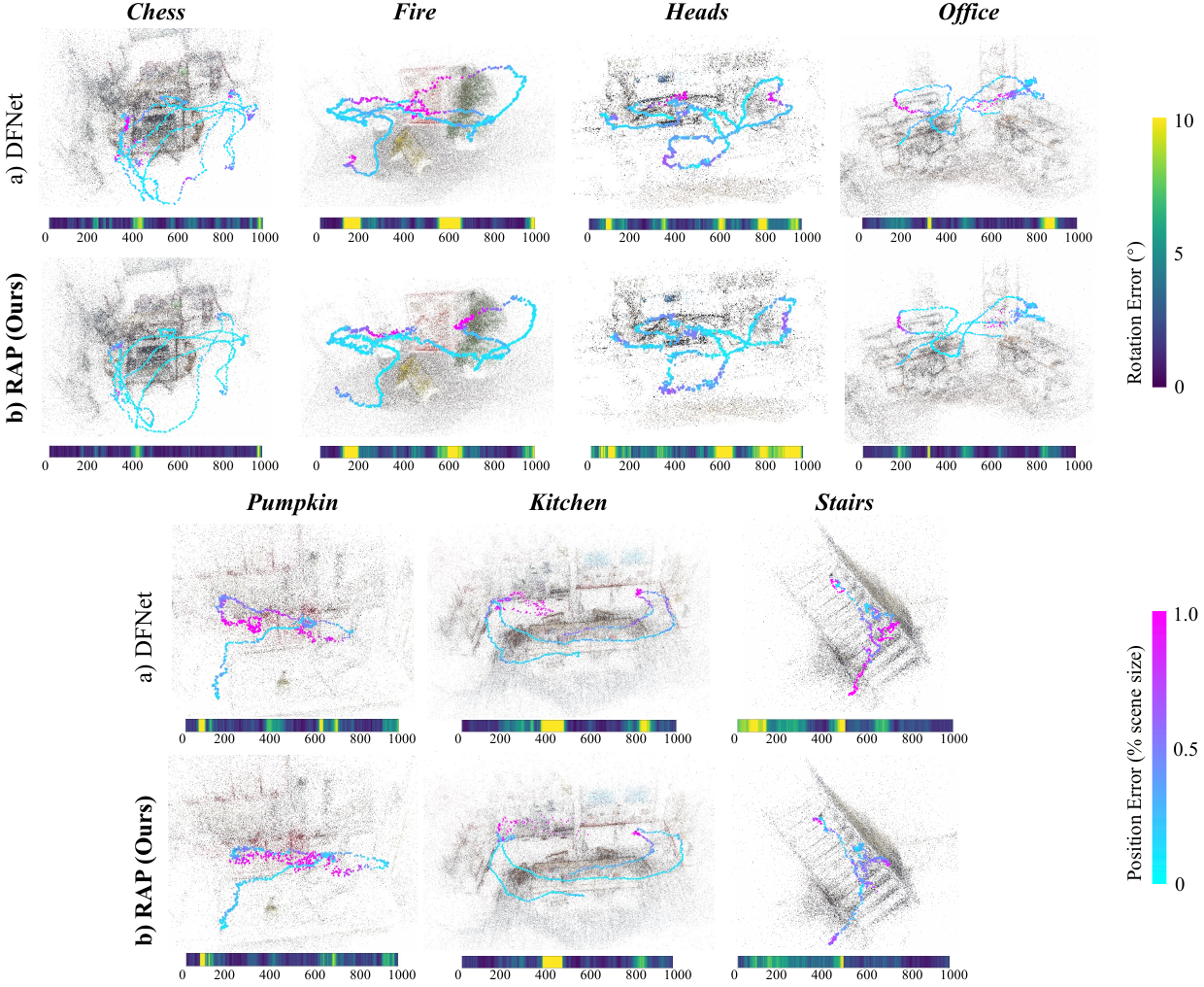}
    \caption{\textbf{Visualization of estimated camera poses on the 7-Scenes dataset~\cite{shotton2013scene}.} Translation and rotation errors are indicated by the color of the error bars. Our RAP framework more closely follows the ground truth trajectory with fewer outliers compared to DFNet~\cite{chen2022dfnet}. The sequences visualized are: \textit{Chess-seq-03}, \textit{Fire-seq-04}, \textit{Heads-seq-01}, \textit{Office-seq-07}, \textit{Pumpkin-seq-01}, \textit{Kitchen-seq-14}, and \textit{Stairs-all}.}
    \label{fig: supp_7scenes_rap}
\end{figure*}

\subsection{Additional Visualization of RAP$_\textbf{ref}$} 
\begin{figure*}[t]
    \centering
    \includegraphics[width=1\linewidth]{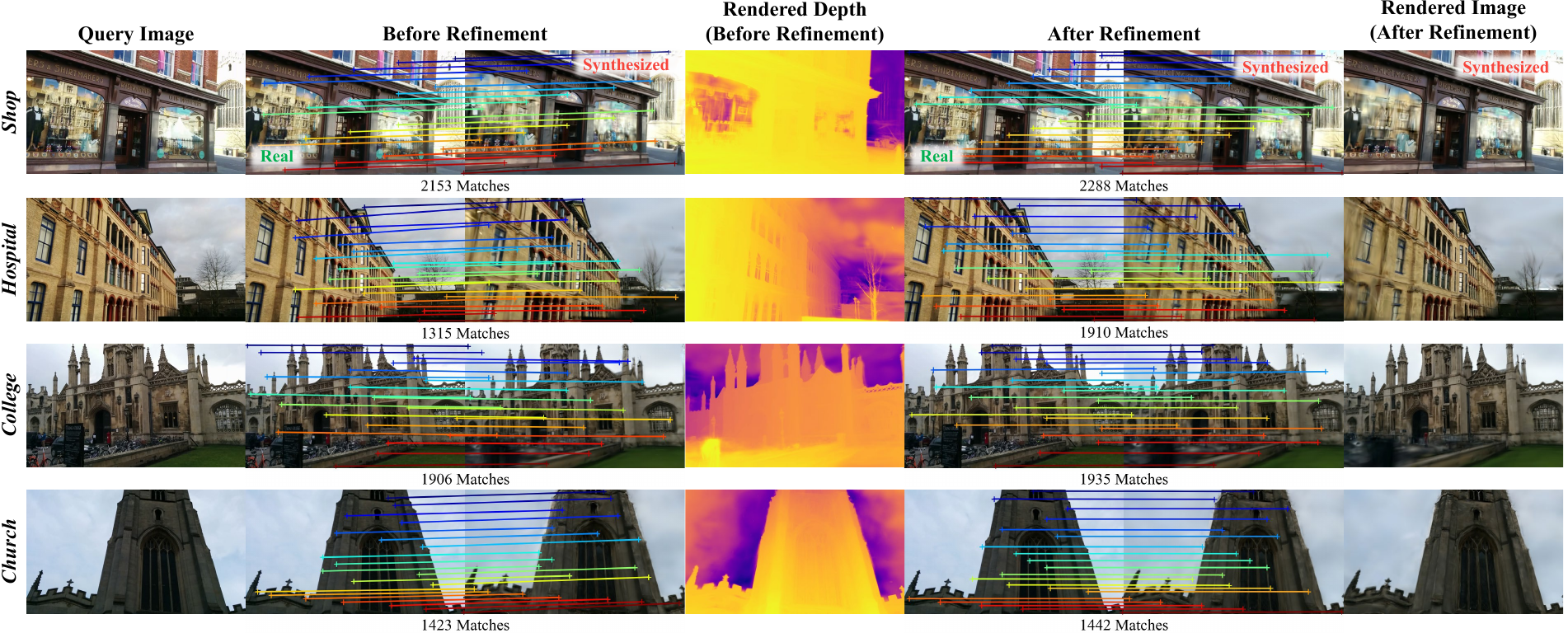}
    \vspace{-15pt}
    \caption{\textbf{Visualization of the post-refinement pipeline on the Cambridge Landmarks dataset~\cite{kendall2015posenet}.} Starting with the query image, we first obtain its initial pose from RAP, render it using 3DGS, and generate matches. The lines before refinement are not sufficiently horizontal due to inaccuracies in the initial pose. Next, we back-project the rendered depth to 3D and use RANSAC-PnP~\cite{fischler1981random, gao2003complete} to compute a refined pose, which is then tested by rendering and matching again. The matches after refinement are horizontal, indicating that the refined poses are more accurate.}
    \label{fig: supp_cam_post}
    \vspace{-20pt}
\end{figure*}

\begin{figure*}[t]
    \centering
    \includegraphics[width=1\linewidth]{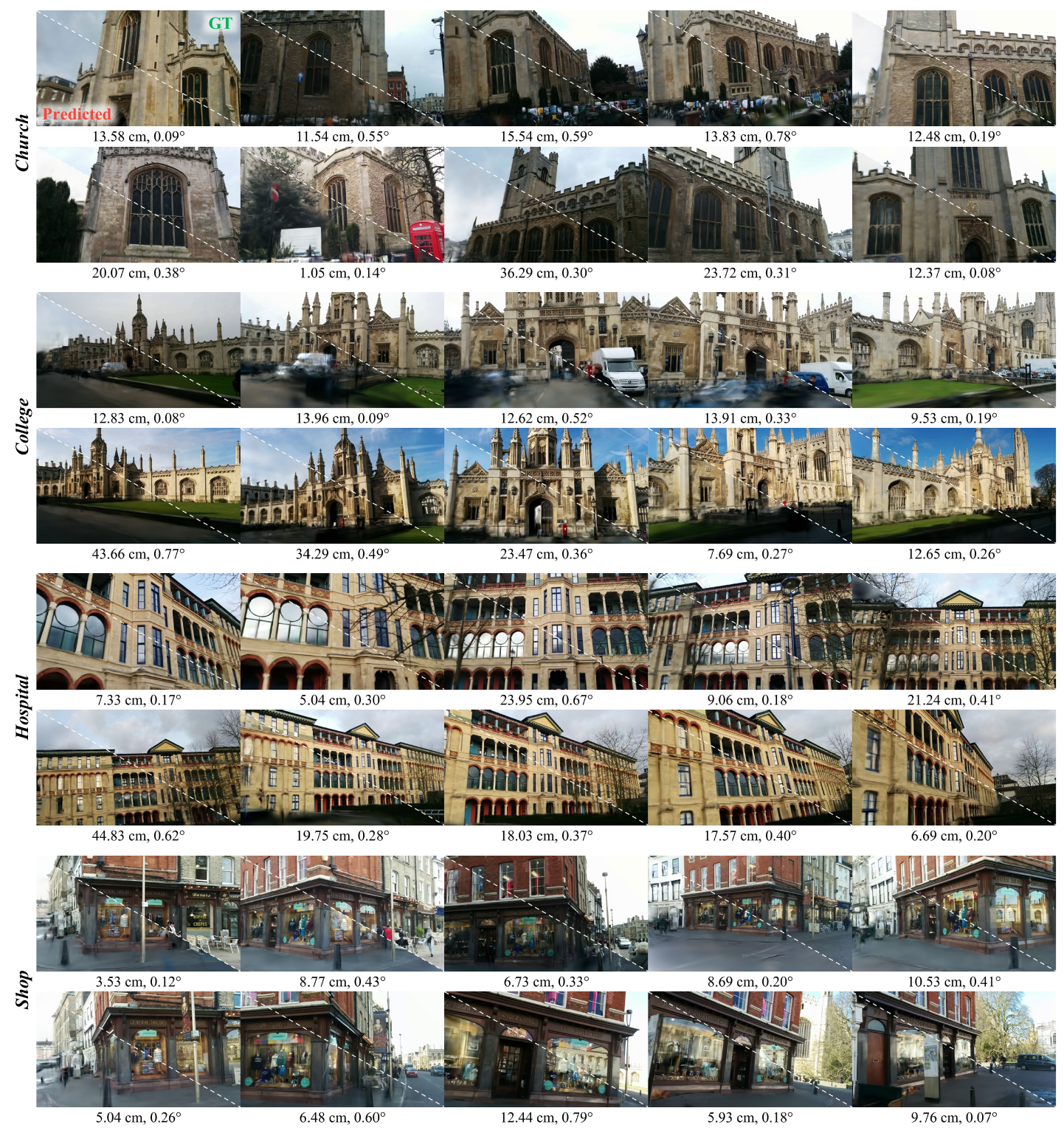}
    \caption{\textbf{Visualization of localization errors on the Cambridge Landmarks dataset~\cite{kendall2015posenet}.} In each sub-figure, a diagonal boundary separates the ``Predicted" (rendered from the refined pose) and ``GT" (ground truth) sections. Smooth alignment along this boundary demonstrates RAP\sub{ref}'s improved pose accuracy.}
    \label{fig: Cambridge_vis_supp}
\end{figure*}

The same process in RAP\sub{ref} on the Cambridge Landmarks dataset is shown in Fig.~\ref{fig: supp_cam_post}, with results in Fig.~\ref{fig: Cambridge_vis_supp}. Compared to indoor scenes, localization errors in outdoor scenes are significantly larger. This is attributed to inherent limitations in scene scale and image resolution. Even when the visual matching between the real and synthesized images appears nearly perfect to humans, as indicated by the image continuity near the diagonal in Fig.~\ref{fig: Cambridge_vis_supp}, errors can still occur within a single pixel in the image coordinate system. For fairness, we use the same resolution as DFNet~\cite{chen2022dfnet}.  

\section{MARS~\cite{li2024multiagent}}
\subsection{Ground Truth Pose Details}
The GPS/IMU poses provided in the dataset are inaccurate, so we use COLMAP~\cite{schonberger2016structure} poses as the ground truth and compute the scaling factor relative to the GPS locations to calculate the metric translation error.  

\subsection{3DGS with Controllable Appearances}
\begin{figure*}[t]
    \centering
    \begin{minipage}{\linewidth}
    \includegraphics[width=1\linewidth]{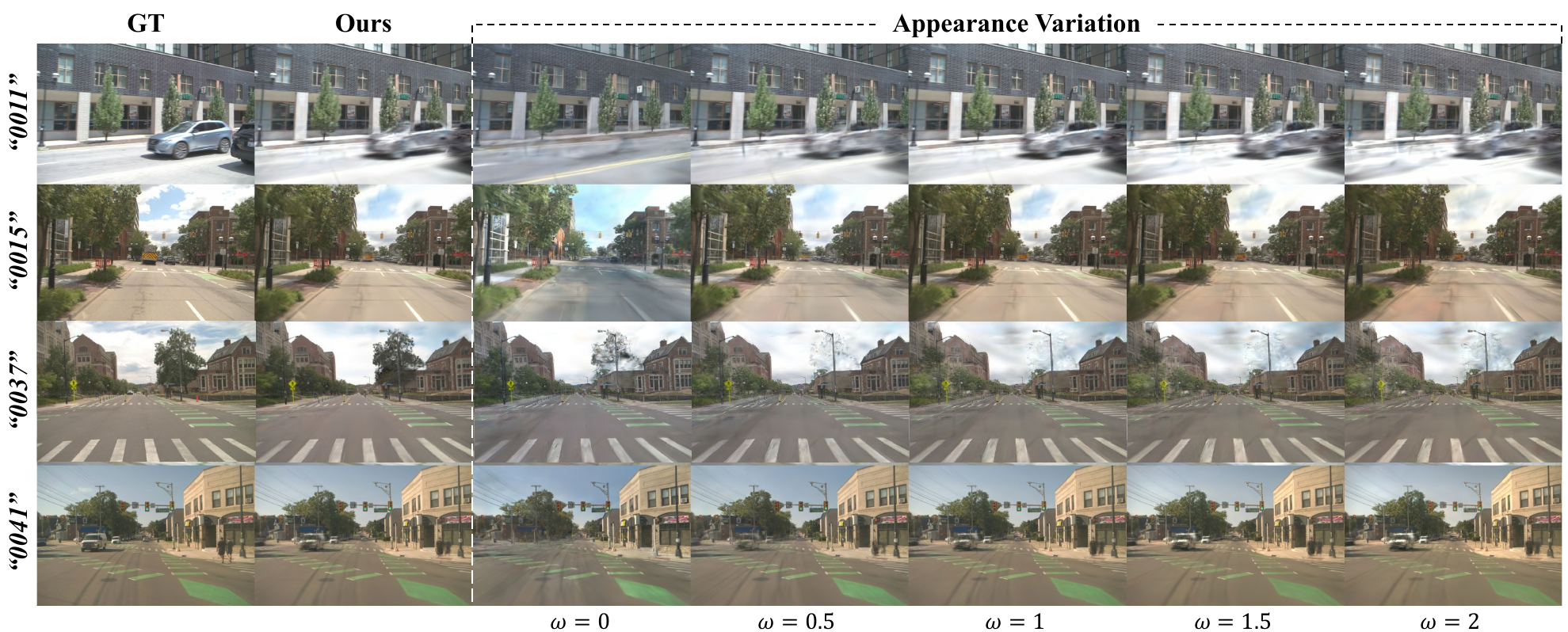}
    \caption{\textbf{Synthetic images with varying appearances on the MARS dataset~\cite{li2024multiagent}.} The appearances of the synthetic images can be arbitrarily generated using different blending weights $\omega$, ranging from 0 to 2.}
    \label{fig: supple_render_compare_mars}
    \end{minipage}
\end{figure*}

The main challenges in driving scenarios include dynamic objects, such as vehicles and pedestrians, and dynamic environments with varying weather conditions. Fig.~\ref{fig: supple_render_compare_mars} shows that our appearance-varying 3DGS successfully models variations in ambient lighting. Notably, it can also capture dynamic elements in the scene, such as vehicles on the road.

\subsection{Additional Visualization of RAP$_\textbf{ref}$} 
\begin{figure*}[t]
    \centering
    \includegraphics[width=1\linewidth]{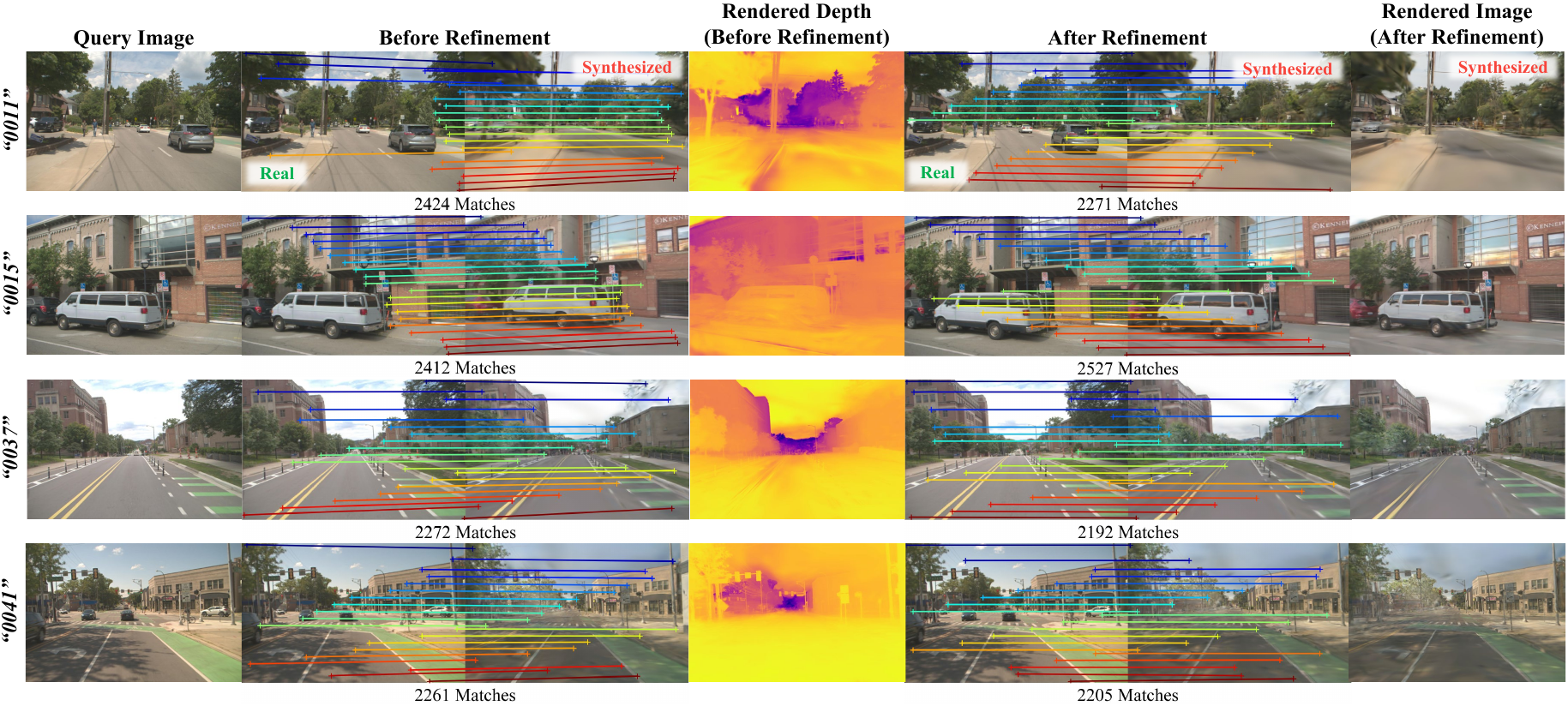}
    \caption{\textbf{Visualization of the post-refinement pipeline on the MARS dataset~\cite{li2024multiagent}.} Starting with the query image, we first obtain its initial pose from RAP, render it using 3DGS, and generate matches. The lines before refinement are not sufficiently horizontal due to inaccuracies in the initial pose. Next, we back-project the rendered depth to 3D and use RANSAC-PnP~\cite{fischler1981random, gao2003complete} to compute a refined pose, which is then tested by rendering and matching again. The matches after refinement are horizontal, indicating that the refined poses are more accurate. Moreover, the rendered depth maps illustrate that our appearance-varying 3DGS successfully reconstructs the scene's geometric information, a critical factor in ensuring accurate 2D-3D correspondences.}
    \label{fig: supp_mars_post}
\end{figure*}

\begin{figure*}[t]
    \centering
    \includegraphics[width=1\linewidth]{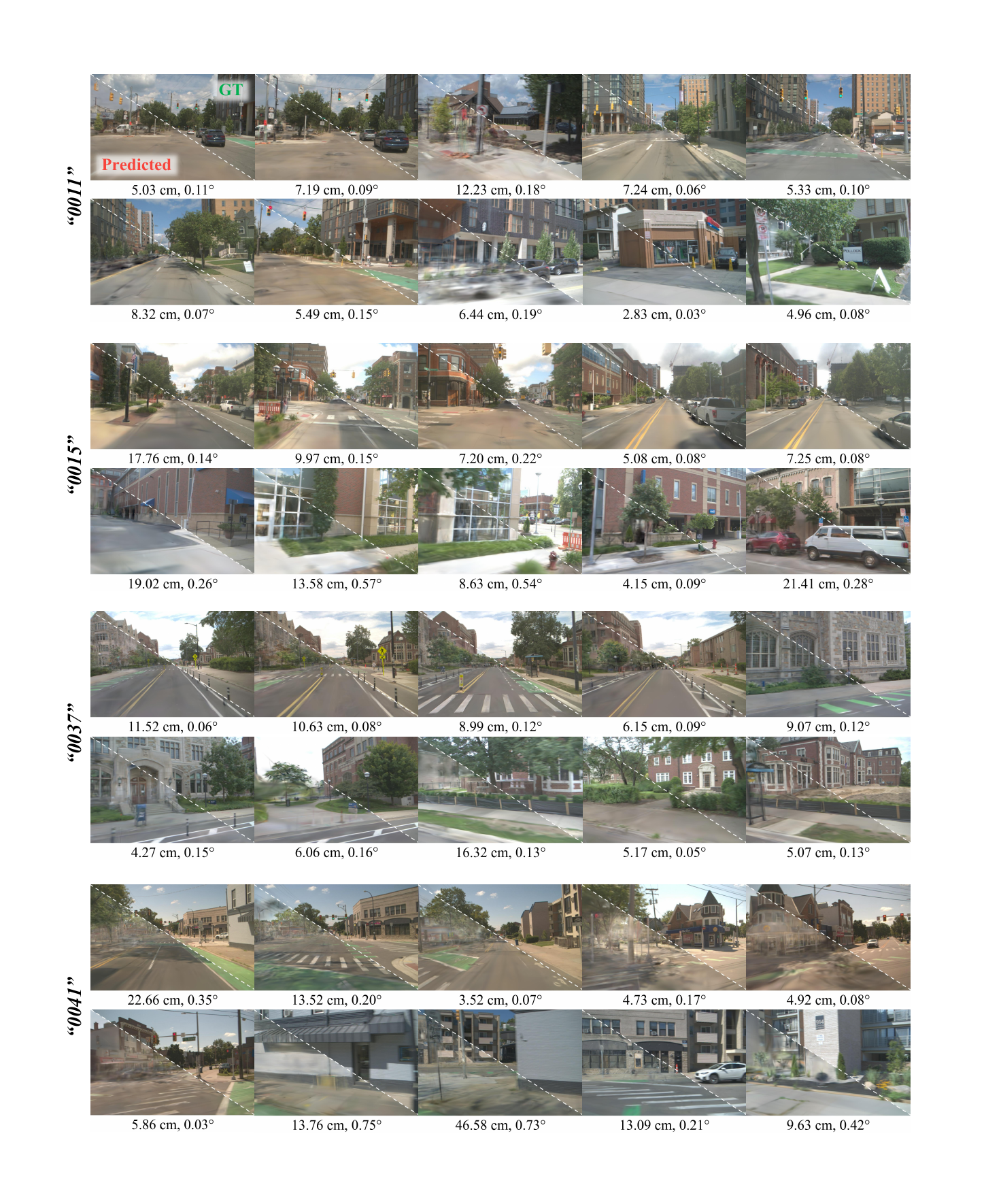}
    \caption{\textbf{Visualization of the localization errors on the MARS dataset~\cite{li2024multiagent}.} In each sub-figure, a diagonal boundary separates the ``Predicted" (rendered from the refined pose) and ``GT" (ground truth) sections. Smooth alignment along this boundary demonstrates RAP\sub{ref}'s improved pose accuracy.}
    \label{fig: MARS_vis_supp}
\end{figure*}

Figure~\ref{fig: supp_mars_post} and Fig.~\ref{fig: MARS_vis_supp} illustrate the same post-refinement process and results of RAP\sub{ref} on the MARS dataset. Our model successfully handles the challenges of varying appearances in autonomous driving scenarios. As shown in Fig.~\ref{fig: MARS_vis_supp}, although the ground truth images and rendered images along the diagonal often exhibit differences in appearance, this does not compromise localization accuracy, as evidenced by the continuity of the images.

\begin{figure*}[t]
    \centering
    \includegraphics[width=0.7\linewidth]{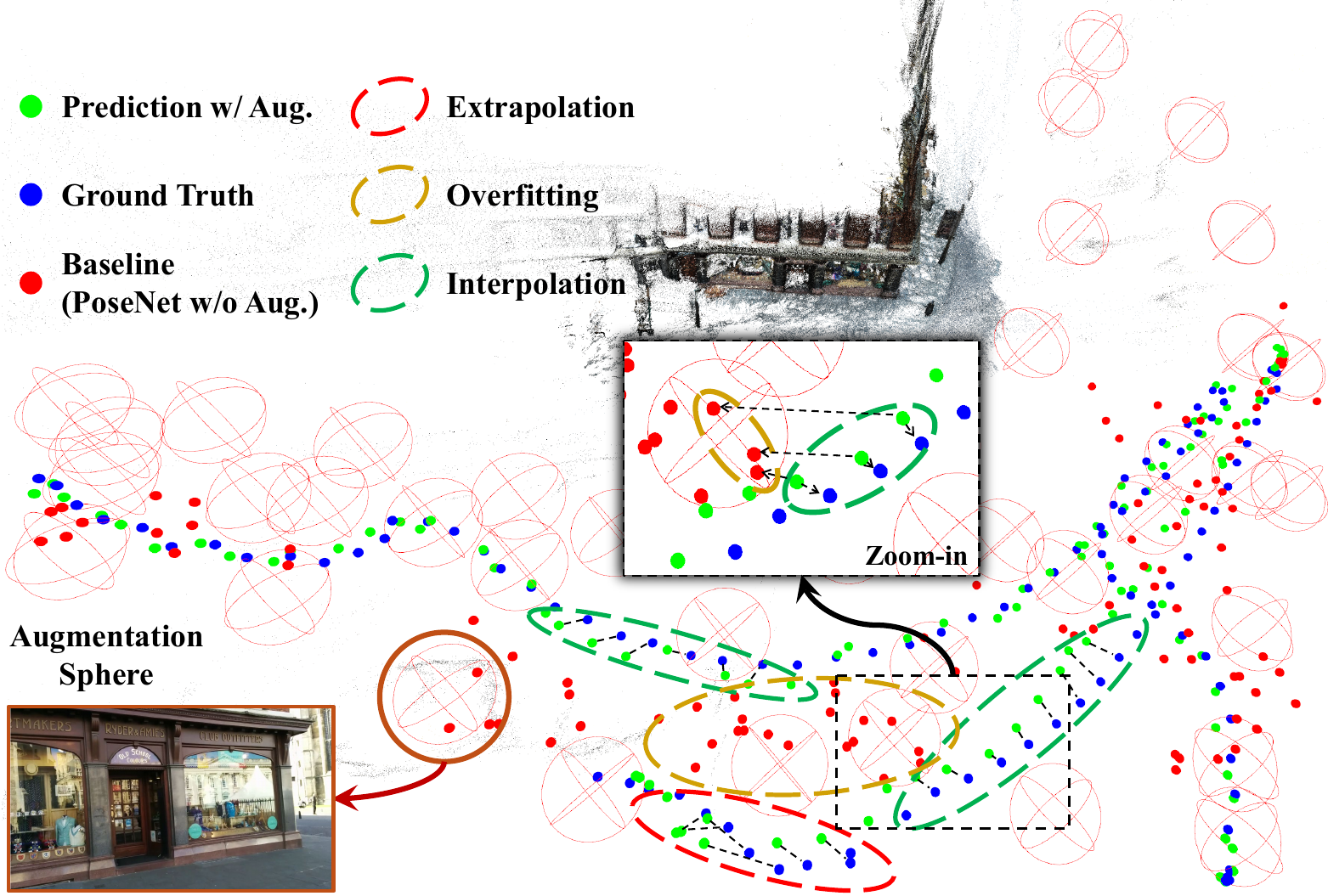}
    % \captionsetup{font=small}
    \caption{\textbf{Visualization of camera locations.} The \textcolor{red}{red hollow spheres}, centered on the real images in the training set, indicate the potential locations of all synthetic images during training. The \textcolor{blue}{blue dots}, \textcolor{green}{green dots}, and \textcolor{red}{red dots} represent the ground truth, predictions by our RAP, and predictions by the baseline, respectively.}
    \label{fig: scaling}
    \vspace{-5pt}
\end{figure*}

\section{Aachen Day-Night~\cite{sattler2018benchmarking}}
\subsection{Subset Details}
We used only a subset of the Aachen dataset due to compatibility issues with the pose ground truth. The full dataset contains images captured using various camera models with different resolutions, and its COLMAP ground truth assigns different camera intrinsics to each image. However, our APR network does not take the camera focal length as input. When the focal length is not consistent, it can lead to ambiguities. For example, close-up shots with a small focal length may appear similar to distant shots with a large focal length, despite different translations. Resolving this would require rerunning COLMAP with a unified camera model, which is computationally expensive for such a large scene, so we opted to use only a subset.

\subsection{Visualization on Appearance Variation}
\begin{figure}
    \centering
    \begin{subfigure}{0.09\textwidth}
        \includegraphics[width=\linewidth]{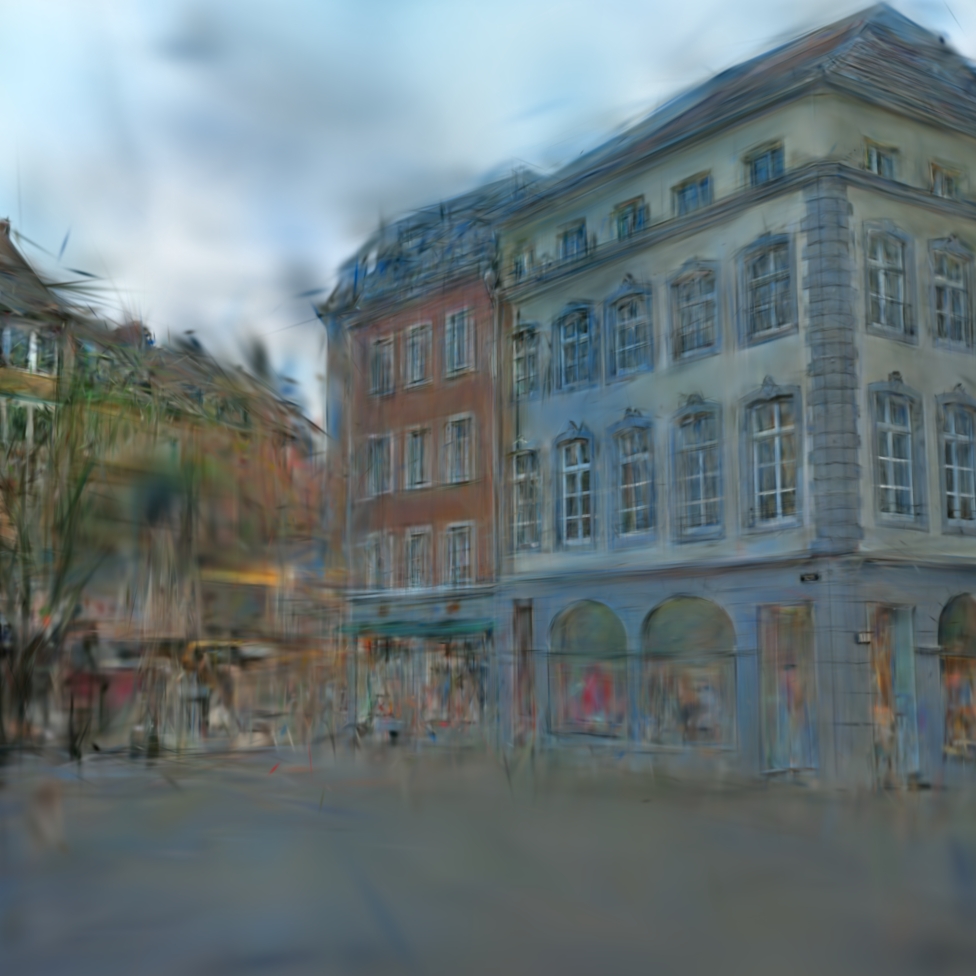}
        \caption{$\omega = 0$}
    \end{subfigure}
    \begin{subfigure}{0.09\textwidth}
        \includegraphics[width=\linewidth]{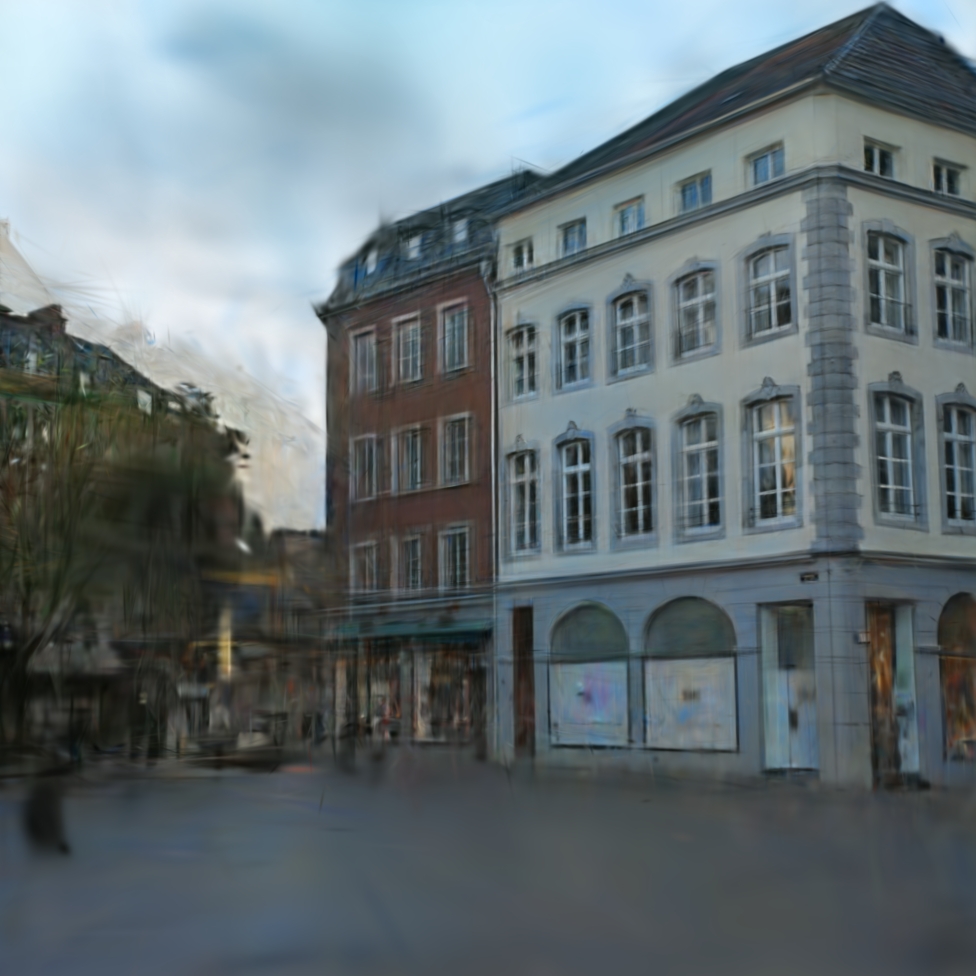}
        \caption{$\omega = 0.5$}
    \end{subfigure}
    \begin{subfigure}{0.09\textwidth}
        \includegraphics[width=\linewidth]{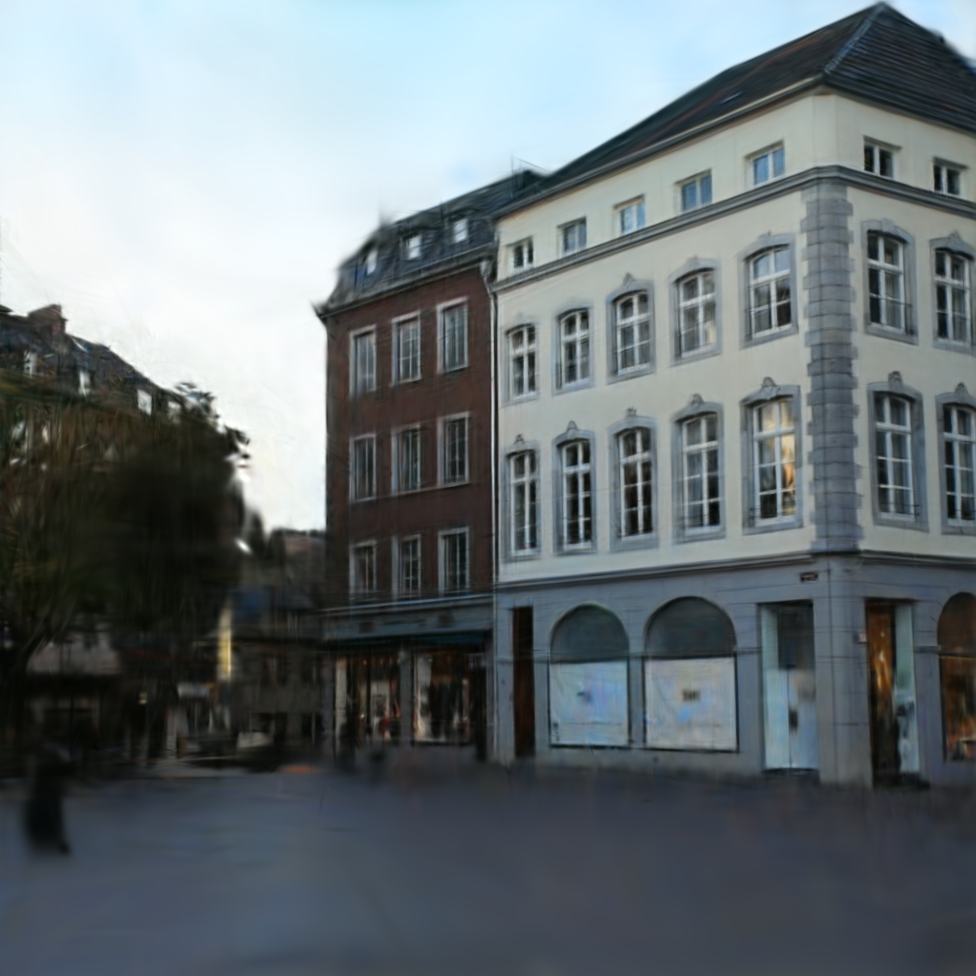}
        \caption{$\omega = 1$}
    \end{subfigure}
    \begin{subfigure}{0.09\textwidth}
        \includegraphics[width=\linewidth]{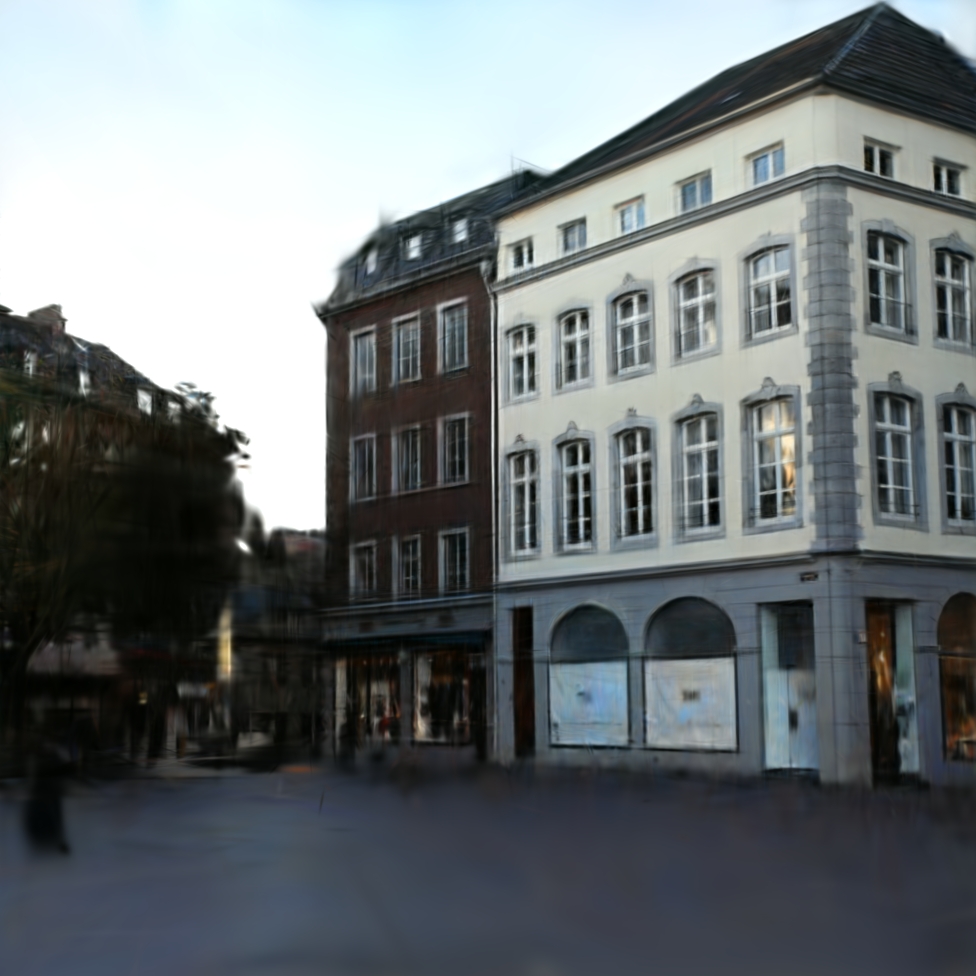}
        \caption{$\omega = 1.5$}
    \end{subfigure}
    \begin{subfigure}{0.09\textwidth}
        \includegraphics[width=\linewidth]{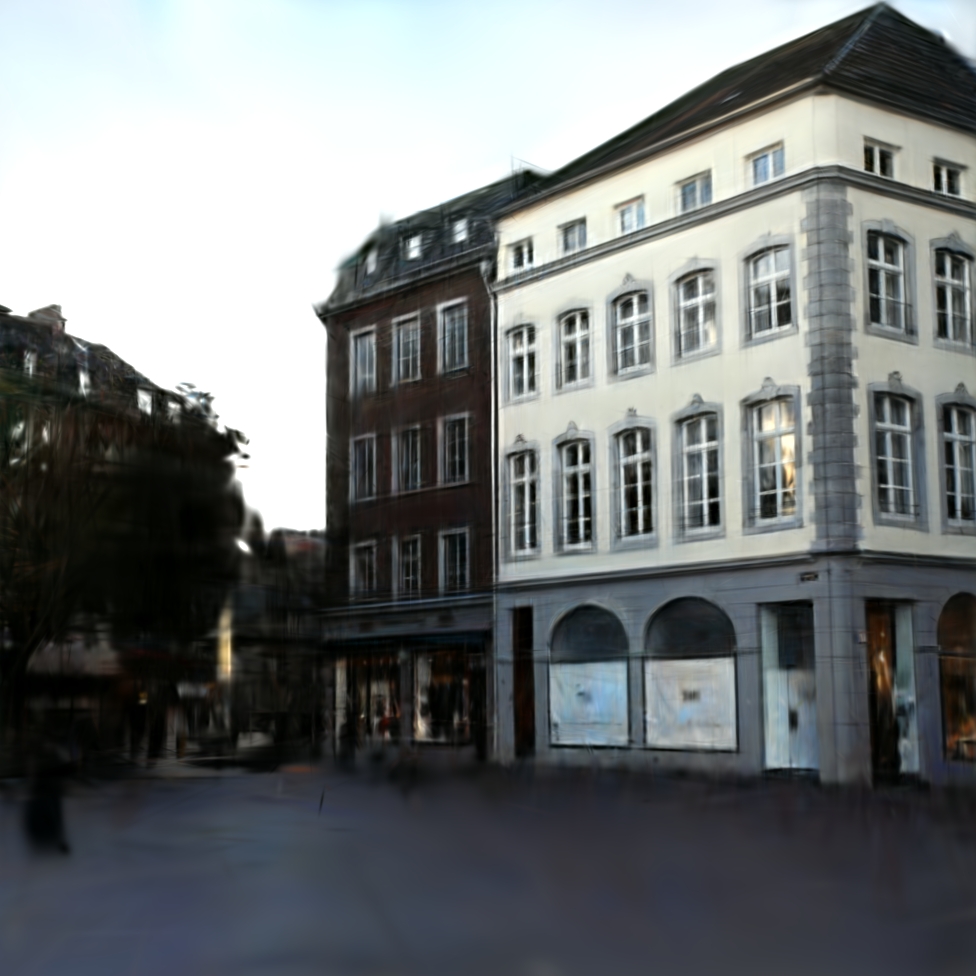}
        \caption{$\omega = 2$}
    \end{subfigure}
\caption{\textbf{Appearance variation in Aachen~\cite{sattler2018benchmarking}.}}
\label{aachen app var}
\end{figure}

Appearance variation in Aachen~\cite{sattler2018benchmarking} is shown in Fig.~\ref{aachen app var}.

\section{7-Scenes~\cite{shotton2013scene}}

\subsection{Visualization of Rendering Quality}
Figure~\ref{fig: supp_7scenes_gs} shows the image rendering results of our method compared to the DFNet~\cite{chen2022dfnet} method across various scenes in the 7-Scenes dataset. DFNet consistently exhibits blurred edges and artifacts in all scenes, primarily due to the low resolution of voxel density sampling in NeRF. When the sampling points are insufficient, edge details become fuzzy. In contrast, our method leverages the explicit 3DGS approach, successfully addressing this issue. The image quality achieved by our method is significantly better than that of NeRF-based methods. Furthermore, our deblurring technique ensures that object edges are clear and sharp, further enhancing the overall rendering quality.

\subsection{Ground Truth Pose Details} 
\label{supp: ground truth pose}
In addition to evaluating performance using SfM ground truth poses of the 7-Scenes dataset, which enable synthesizing higher-quality images~\cite{chen2024neural}, we also provide results based on DSLAM~\cite{newcombe2011kinectfusion} ground truth poses of the same dataset. As shown in Fig.~\ref{fig: supp_7scenes_gs}, SfM poses yield more accurate results, while DSLAM poses introduce noticeable artifacts along object edges. The quantitative results in Table~\ref{table:PSNR} compare the image quality metric (PSNR) for the two sets of poses, demonstrating a significant improvement in image quality with SfM poses. This enhancement further boosts the performance of APR. Furthermore, the table in the paper shows that RAP using SfM poses achieves lower localization errors, although RAP using DSLAM poses already delivers state-of-the-art performance.  % Table~\ref{table:7scenes}

\subsection{Additional Visualization of RAP} 
We present qualitative comparisons on the subsets of the 7-Scenes dataset in Fig.~\ref{fig: supp_7scenes_rap}, comparing our RAP with DFNet~\cite{chen2022dfnet}. In \textit{Fire-seq-04}, \textit{Pumpkin-seq-01}, and \textit{Kitchen-seq-14}, our RAP avoids collapsing in certain regions, unlike DFNet, which generates a significant number of outliers. This demonstrates our RAP's strong generalizability. 

\subsection{Additional Visualization of RAP$_\textbf{ref}$} 
\begin{figure*}[t]
    \centering
    \includegraphics[width=1\linewidth]{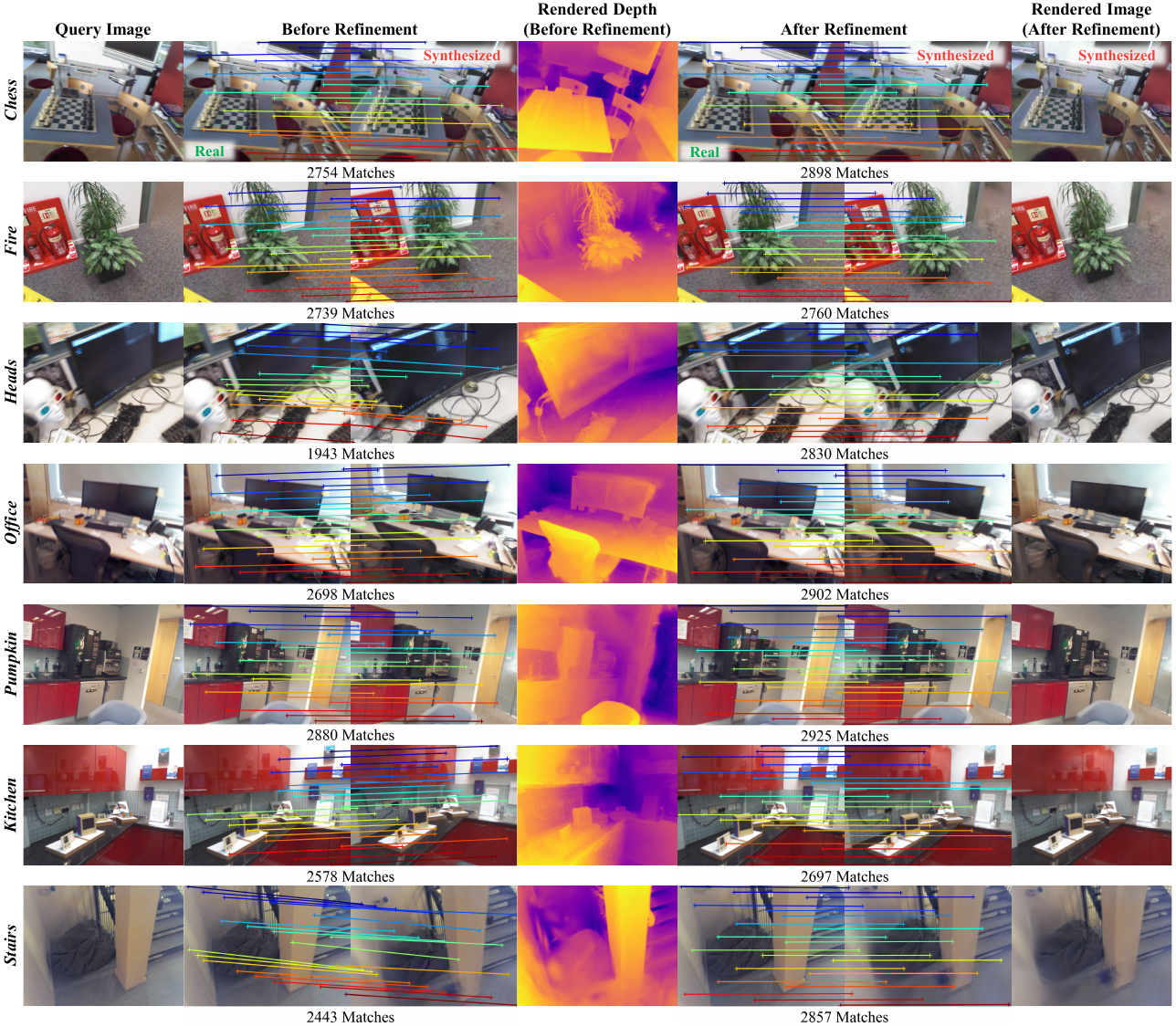}
    \caption{\textbf{Visualization of the post-refinement pipeline on the 7-Scenes dataset~\cite{shotton2013scene}.} Starting with the query image, we first obtain its initial pose from RAP, render it using 3DGS, and generate matches. The lines before refinement are not sufficiently horizontal due to inaccuracies in the initial pose. Next, we back-project the rendered depth to 3D and use RANSAC-PnP~\cite{fischler1981random, gao2003complete} to compute a refined pose, which is then tested by rendering and matching again. The matches after refinement are horizontal, indicating that the refined poses are more accurate. Moreover, the rendered depth maps illustrate that our appearance-varying 3DGS successfully reconstructs the scene's geometric information, a critical factor in ensuring accurate 2D-3D correspondences.}
    \label{fig: supp_7scenes_post}
\end{figure*}

\begin{figure*}[t]
    \centering
    \includegraphics[width=1\linewidth]{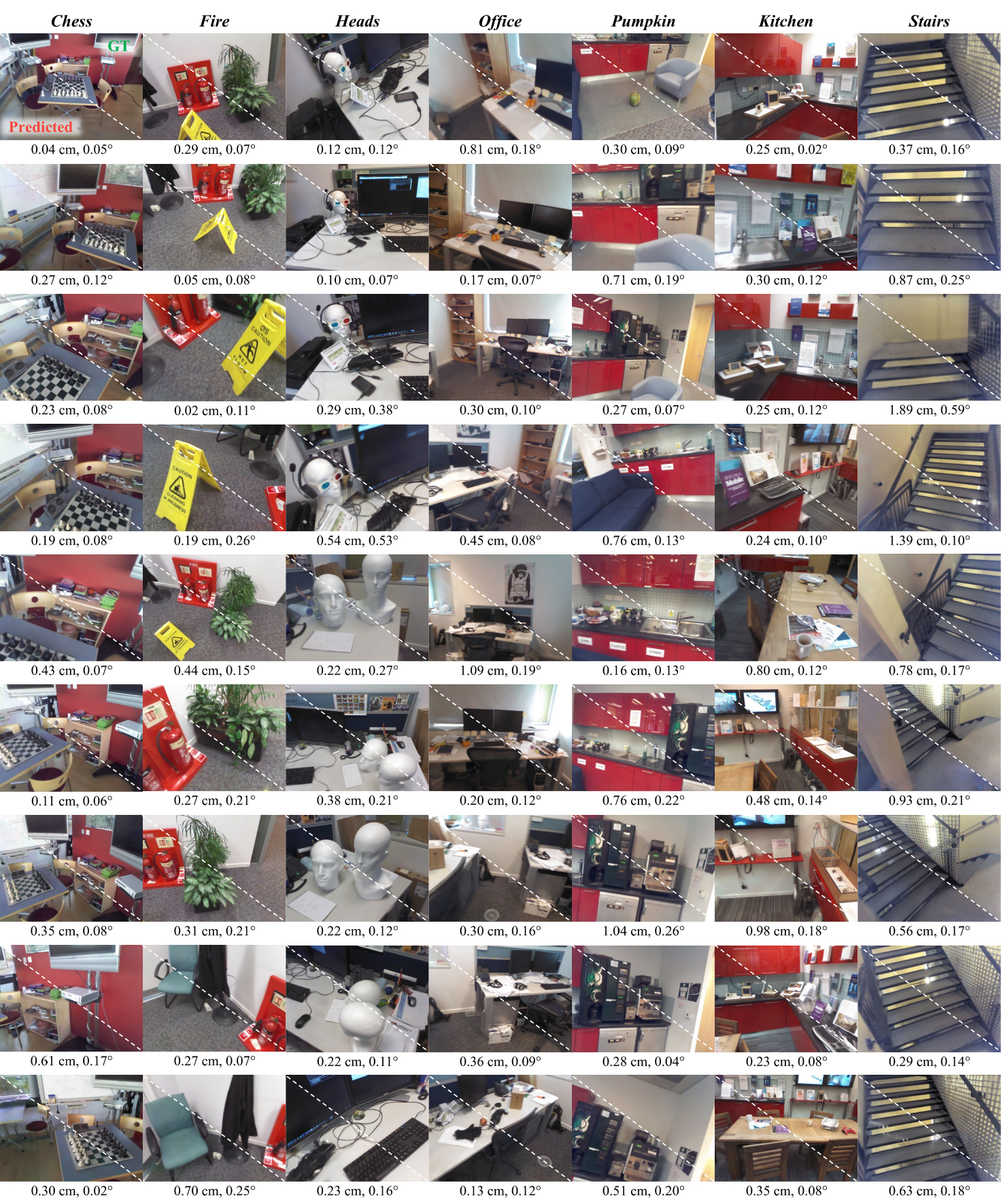}
    \caption{\textbf{Visualization of localization errors on the 7-Scenes dataset~\cite{shotton2013scene}.} In each sub-figure, a diagonal boundary separates the ``Predicted" (rendered from the refined pose) and ``GT" (ground truth) sections. Smooth alignment along this boundary demonstrates RAP\sub{ref}'s improved pose accuracy.}
    \label{fig: 7scenes_vis_supp}
\end{figure*}

Figure~\ref{fig: supp_7scenes_post} illustrates the intermediate steps involved in post-refinement. Specifically, after obtaining the initial pose estimation of the query image from RAP, we render the corresponding image and depth map through 3DGS. Then, we use MASt3R~\cite{leroy2024grounding} to calculate the pixel correspondences between the two images. As shown in Fig.~\ref{fig: supp_7scenes_post}, the matching lines before refinement are not sufficiently horizontal, indicating inaccuracies in the initial pose estimation. Next, we derive 2D-3D correspondences from the depth map and optimize the pose using a RANSAC-PnP~\cite{fischler1981random, gao2003complete} solver. The final column of images demonstrates that the refined pose produces a rendered image almost indistinguishable from the original query image, with the matching lines now highly horizontal, demonstrating the improved accuracy of the pose estimation. As shown in Fig.~\ref{fig: 7scenes_vis_supp}, the errors of our RAP\sub{ref} are even less than 1 centimeter.

\section{Emerging Generalization in APR}
We experimented on \textit{Shop} using only 20\% of the real training set with our synthetic data, as shown in Fig.~\ref{fig: scaling}. We see that the training set with synthetic data, represented by the red hollow spheres, does not fully cover the test set spatially. Despite this, the model still closely predicts the test camera poses, demonstrating generalization ability beyond the original training positions.

\begin{table}[t]
\caption{\textbf{Inference efficiency.} Measured on \textit{Heads} with input images of size $320\times240$.}
\label{effi}
\centering
\scalebox{0.8}{
\begin{tabular}{llc}
\toprule
\textbf{Method}                      & \textbf{PyTorch Mode}                 & \textbf{Avg FPS} $\uparrow$ \\
\midrule
ACE~\cite{brachmann2023accelerated}  & With C++                              & 50                          \\
\midrule
\multirow{5}{*}{\textbf{RAP (Ours)}} & Eager                                 & 105                         \\
                                     & Compiled                              & 154                         \\
                                     & Compiled \texttt{reduce-overhead}     & 187                         \\
                                     & Compiled AMP                          & 192                         \\
                                     & Compiled \texttt{reduce-overhead} AMP & 279                         \\
\bottomrule
\end{tabular}}
% \vspace{-12pt}
\end{table}

\section{Inference Efficiency}
As shown in Table~\ref{effi}, our Python prototype of RAP achieves approximately 279 FPS with the \texttt{reduce-overhead} mode of \texttt{torch.compile}\footnote{Timing measured using the function provided in \url{https://pytorch.org/tutorials/intermediate/torch_compile_tutorial.html}; dataloader time is excluded.} and AMP enabled on a laptop equipped with an NVIDIA RTX 4060 GPU running at 30 W and an Intel Core i9-13900H CPU, demonstrating real-time inference performance on compact devices. For RAP\sub{ref}, the post-refinement time per frame is 0.5~s on the same device, including RAP inference, 3DGS rendering with \texttt{gsplat}~\cite{ye2024gsplat}, MASt3R matching, and RANSAC-PnP~\cite{fischler1981random, gao2003complete} solving using OpenCV. During this process, the GPU power consumption can reach 70–80~W. Please see our code for additional implementation details.

\begin{figure}[t]
    \centering
    \includegraphics[width=1\linewidth]{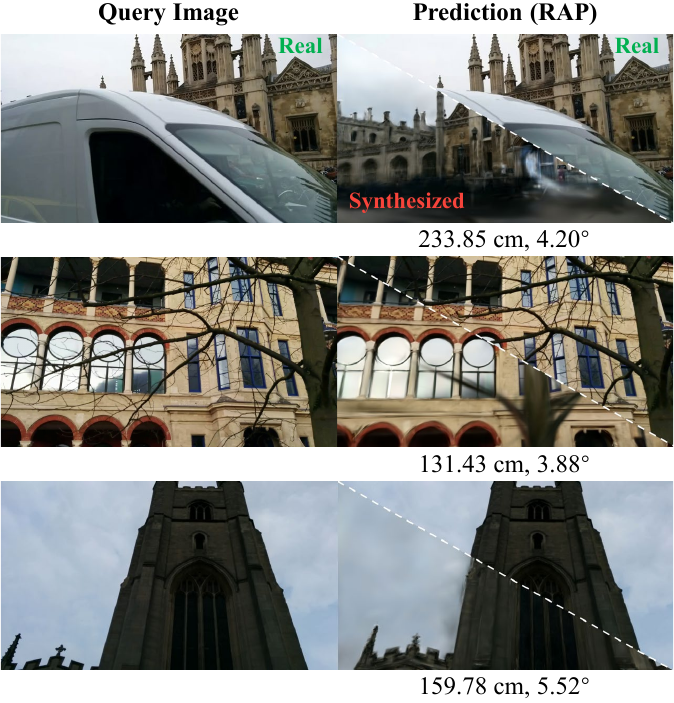}
    % \captionsetup{font=small}
    \caption{\textbf{Failure cases.} The primary reason for localization failure is occlusion, as shown in the first two rows. Additionally, textureless regions in the query image, such as the sky, can also result in significant errors.}
    % of RAP$_\text{ref}$}.}
    \label{fig: fail}
\end{figure}

\section{Failure Cases}
Fig.~\ref{fig: fail} presents several failure cases encountered during evaluation. Occlusions pose the most significant challenge for APR, particularly when dynamic objects are present. For example, in the second-row images, tree branches—absent in the 3DGS-synthesized image—appear during the inference stage, disrupting feature extraction. Additionally, textureless patterns in the image can degrade APR performance. For instance, in the third row, the stark contrast between the featureless sky and the building's underexposed color creates ambiguities, posing challenges for feature extraction, potentially misleading the regression head, and impacting localization accuracy. 

\section{Limitations and Future Work}
Like other APR approaches, our method has yet to surpass geometry-based techniques in accuracy, and per-scene training remains time-consuming. We also observe accuracy loss when training on the metric scale in large scenes. Additionally, current APR methods do not account for camera intrinsics and are sensitive to input image resolution, leading to accuracy degradation when the testing resolution differs from training. Future directions include efficiently training stronger APR models with geometric priors, leveraging temporal information, and integrating powerful vision foundation models~\cite{oquab2023dinov2}. Generalizing to dynamic environments with fewer training samples is also a promising research avenue.

%% file: main.bib
@String(AAAI = {AAAI})

@String(AAAI  = {AAAI})

@inproceedings{kendall2015posenet,
  title={Posenet: A convolutional network for real-time 6-dof camera relocalization},
  author={Kendall, Alex and Grimes, Matthew and Cipolla, Roberto},
  booktitle={IEEE international conference on computer vision},
  pages={2938--2946},
  year={2015}
}

@inproceedings{kendall2017geometric,
  title={Geometric loss functions for camera pose regression with deep learning},
  author={Kendall, Alex and Cipolla, Roberto},
  booktitle={IEEE/CVF Conference on Computer Vision and Pattern Recognition},
  pages={5974--5983},
  year={2017}
}

@inproceedings{shavit2021learning,
  title={Learning multi-scene absolute pose regression with transformers},
  author={Shavit, Yoli and Ferens, Ron and Keller, Yosi},
  booktitle={IEEE/CVF International Conference on Computer Vision},
  pages={2733--2742},
  year={2021}
}

@inproceedings{chen2022dfnet,
  title={Dfnet: Enhance absolute pose regression with direct feature matching},
  author={Chen, Shuai and Li, Xinghui and Wang, Zirui and Prisacariu, Victor A},
  booktitle={European Conference on Computer Vision},
  pages={1--17},
  year={2022},
  organization={Springer}
}

@inproceedings{tan2019efficientnet,
  title={Efficientnet: Rethinking model scaling for convolutional neural networks},
  author={Tan, Mingxing and Le, Quoc},
  booktitle={International Conference on Machine Learning},
  pages={6105--6114},
  year={2019},
}

@inproceedings{shotton2013scene,
  title={Scene coordinate regression forests for camera relocalization in RGB-D images},
  author={Shotton, Jamie and Glocker, Ben and Zach, Christopher and Izadi, Shahram and Criminisi, Antonio and Fitzgibbon, Andrew},
  booktitle={IEEE/CVF Conference on Computer Vision and Pattern Recognition},
  pages={2930--2937},
  year={2013}
}

@inproceedings{moreau2022coordinet,
  title={CoordiNet: uncertainty-aware pose regressor for reliable vehicle localization},
  author={Moreau, Arthur and Piasco, Nathan and Tsishkou, Dzmitry and Stanciulescu, Bogdan and de La Fortelle, Arnaud},
  booktitle={IEEE/CVF Winter Conference on Applications of Computer Vision},
  pages={2229--2238},
  year={2022}
}

@inproceedings{moreau2022lens,
  title={LENS: Localization enhanced by NeRF synthesis},
  author={Moreau, Arthur and Piasco, Nathan and Tsishkou, Dzmitry and Stanciulescu, Bogdan and de La Fortelle, Arnaud},
  booktitle={Conference on Robot Learning},
  pages={1347--1356},
  year={2022}
}

@inproceedings{sattler2018benchmarking,
  title={Benchmarking 6dof outdoor visual localization in changing conditions},
  author={Sattler, Torsten and Maddern, Will and Toft, Carl and Torii, Akihiko and Hammarstrand, Lars and Stenborg, Erik and Safari, Daniel and Okutomi, Masatoshi and Pollefeys, Marc and Sivic, Josef and others},
  booktitle={IEEE/CVF Conference on Computer Vision and Pattern Recognition},
  pages={8601--8610},
  year={2018}
}

@inproceedings{sattler2019understanding,
  title={Understanding the limitations of cnn-based absolute camera pose regression},
  author={Sattler, Torsten and Zhou, Qunjie and Pollefeys, Marc and Leal-Taixe, Laura},
  booktitle={IEEE/CVF Conference on Computer Vision and Pattern Recognition},
  pages={3302--3312},
  year={2019}
}

@inproceedings{dusmanu2019d2,
  title={D2-net: A trainable cnn for joint description and detection of local features},
  author={Dusmanu, Mihai and Rocco, Ignacio and Pajdla, Tomas and Pollefeys, Marc and Sivic, Josef and Torii, Akihiko and Sattler, Torsten},
  booktitle={IEEE/CVF Conference on Computer Vision and Pattern Recognition},
  pages={8092--8101},
  year={2019}
}

@inproceedings{sarlin2019coarse,
  title={From coarse to fine: Robust hierarchical localization at large scale},
  author={Sarlin, Paul-Edouard and Cadena, Cesar and Siegwart, Roland and Dymczyk, Marcin},
  booktitle={IEEE/CVF Conference on Computer Vision and Pattern Recognition},
  pages={12716--12725},
  year={2019}
}

@inproceedings{sarlin2020superglue,
  title={Superglue: Learning feature matching with graph neural networks},
  author={Sarlin, Paul-Edouard and DeTone, Daniel and Malisiewicz, Tomasz and Rabinovich, Andrew},
  booktitle={IEEE/CVF Conference on Computer Vision and Pattern Recognition},
  pages={4938--4947},
  year={2020}
}

@inproceedings{taira2018inloc,
  title={InLoc: Indoor visual localization with dense matching and view synthesis},
  author={Taira, Hajime and Okutomi, Masatoshi and Sattler, Torsten and Cimpoi, Mircea and Pollefeys, Marc and Sivic, Josef and Pajdla, Tomas and Torii, Akihiko},
  booktitle={IEEE/CVF Conference on Computer Vision and Pattern Recognition},
  pages={7199--7209},
  year={2018}
}

@inproceedings{noh2017large,
  title={Large-scale image retrieval with attentive deep local features},
  author={Noh, Hyeonwoo and Araujo, Andre and Sim, Jack and Weyand, Tobias and Han, Bohyung},
  booktitle={IEEE international conference on computer vision},
  pages={3456--3465},
  year={2017}
}

@inproceedings{zheng2015structure,
  title={Structure from motion using structure-less resection},
  author={Zheng, Enliang and Wu, Changchang},
  booktitle={IEEE International Conference on Computer Vision},
  pages={2075--2083},
  year={2015}
}

@inproceedings{chen2024neural,
  title={Neural Refinement for Absolute Pose Regression with Feature Synthesis},
  author={Chen, Shuai and Bhalgat, Yash and Li, Xinghui and Bian, Jia-Wang and Li, Kejie and Wang, Zirui and Prisacariu, Victor Adrian},
  booktitle={IEEE/CVF Conference on Computer Vision and Pattern Recognition},
  pages={20987--20996},
  year={2024}
}

@article{simonyan2014very,
  title={Very deep convolutional networks for large-scale image recognition},
  author={Simonyan, Karen and Zisserman, Andrew},
  journal={arXiv preprint arXiv:1409.1556},
  year={2014}
}

@inproceedings{brachmann2017dsac,
  title={Dsac-differentiable ransac for camera localization},
  author={Brachmann, Eric and Krull, Alexander and Nowozin, Sebastian and Shotton, Jamie and Michel, Frank and Gumhold, Stefan and Rother, Carsten},
  booktitle={IEEE/CVF Conference on Computer Vision and Pattern Recognition},
  pages={6684--6692},
  year={2017}
}

@article{brachmann2021visual,
  title={Visual camera re-localization from RGB and RGB-D images using DSAC},
  author={Brachmann, Eric and Rother, Carsten},
  journal={IEEE Transactions on Pattern Analysis and Machine Intelligence},
  volume={44},
  number={9},
  pages={5847--5865},
  year={2021},
}

@InProceedings{Kingma15adam,
  author    = {Kingma, Diederik and Ba, Jimmy},
  booktitle = {International Conference on Learning Representations},
  title     = {Adam: A Method for Stochastic Optimization},
  year      = {2015},
  optmonth  = {12},
}

@inproceedings{lindenberger2023lightglue,
  title={Lightglue: Local feature matching at light speed},
  author={Lindenberger, Philipp and Sarlin, Paul-Edouard and Pollefeys, Marc},
  booktitle={IEEE/CVF International Conference on Computer Vision},
  pages={17627--17638},
  year={2023}
}

@inproceedings{chen2024map,
  title={Map-Relative Pose Regression for Visual Re-Localization},
  author={Chen, Shuai and Cavallari, Tommaso and Prisacariu, Victor Adrian and Brachmann, Eric},
  booktitle={IEEE/CVF Conference on Computer Vision and Pattern Recognition},
  pages={20665--20674},
  year={2024}
}

@article{gao2003complete,
  title={Complete solution classification for the perspective-three-point problem},
  author={Gao, Xiao-Shan and Hou, Xiao-Rong and Tang, Jianliang and Cheng, Hang-Fei},
  journal={IEEE Transactions on Pattern Analysis and Machine Intelligence},
  volume={25},
  number={8},
  pages={930--943},
  year={2003},
}

@article{fischler1981random,
  title={Random sample consensus: a paradigm for model fitting with applications to image analysis and automated cartography},
  author={Fischler, Martin A and Bolles, Robert C},
  journal={Communications of the ACM},
  volume={24},
  number={6},
  pages={381--395},
  year={1981},
}

@inproceedings{moreau2023crossfire,
  title={Crossfire: Camera relocalization on self-supervised features from an implicit representation},
  author={Moreau, Arthur and Piasco, Nathan and Bennehar, Moussab and Tsishkou, Dzmitry and Stanciulescu, Bogdan and de La Fortelle, Arnaud},
  booktitle={IEEE/CVF International Conference on Computer Vision},
  pages={252--262},
  year={2023}
}

@article{bjorck1967solving,
  title={Solving linear least squares problems by Gram-Schmidt orthogonalization},
  author={Bj{\"o}rck, {\AA}ke},
  journal={BIT Numerical Mathematics},
  volume={7},
  number={1},
  pages={1--21},
  year={1967},
}

@article{kerbl20233d,
  title={3D Gaussian Splatting for Real-Time Radiance Field Rendering},
  author={Kerbl, Bernhard and Kopanas, Georgios and Leimk{\"u}hler, Thomas and Drettakis, George},
  journal={ACM Transaction on Graphics},
  volume={42},
  number={4},
  year={2023}
}

@article{leroy2024grounding,
  title={Grounding Image Matching in 3D with MASt3R},
  author={Leroy, Vincent and Cabon, Yohann and Revaud, J{\'e}r{\^o}me},
  journal={arXiv preprint arXiv:2406.09756},
  year={2024}
}

@inproceedings{wang2024dust3r,
  title={Dust3r: Geometric 3d vision made easy},
  author={Wang, Shuzhe and Leroy, Vincent and Cabon, Yohann and Chidlovskii, Boris and Revaud, Jerome},
  booktitle={IEEE/CVF Conference on Computer Vision and Pattern Recognition},
  pages={20697--20709},
  year={2024}
}

@article{zhou2024nerfect,
  title={The NeRFect match: Exploring NeRF features for visual localization},
  author={Zhou, Qunjie and Maximov, Maxim and Litany, Or and Leal-Taix{\'e}, Laura},
  journal={arXiv preprint arXiv:2403.09577},
  year={2024}
}

@inproceedings{brachmann2023accelerated,
  title={Accelerated coordinate encoding: Learning to relocalize in minutes using rgb and poses},
  author={Brachmann, Eric and Cavallari, Tommaso and Prisacariu, Victor Adrian},
  booktitle={IEEE/CVF Conference on Computer Vision and Pattern Recognition},
  pages={5044--5053},
  year={2023}
}

@inproceedings{wang2024glace,
  title={GLACE: Global Local Accelerated Coordinate Encoding},
  author={Wang, Fangjinhua and Jiang, Xudong and Galliani, Silvano and Vogel, Christoph and Pollefeys, Marc},
  booktitle={IEEE/CVF Conference on Computer Vision and Pattern Recognition},
  pages={21562--21571},
  year={2024}
}

@INPROCEEDINGS{liu2024hr,
  author={Liu, Changkun and Chen, Shuai and Zhao, Yukun and Huang, Huajian and Prisacariu, Victor and Braud, Tristan},
  booktitle={2024 IEEE International Conference on Robotics and Automation}, 
  title={HR-APR: APR-agnostic Framework with Uncertainty Estimation and Hierarchical Refinement for Camera Relocalisation}, 
  year={2024},
  volume={},
  number={},
  pages={8544-8550},}

@inproceedings{germain2022feature,
  title={Feature query networks: Neural surface description for camera pose refinement},
  author={Germain, Hugo and DeTone, Daniel and Pascoe, Geoffrey and Schmidt, Tanner and Novotny, David and Newcombe, Richard and Sweeney, Chris and Szeliski, Richard and Balntas, Vasileios},
  booktitle={IEEE/CVF Conference on Computer Vision and Pattern Recognition},
  pages={5071--5081},
  year={2022}
}

@inproceedings{trivigno2024unreasonable,
  title={The Unreasonable Effectiveness of Pre-Trained Features for Camera Pose Refinement},
  author={Trivigno, Gabriele and Masone, Carlo and Caputo, Barbara and Sattler, Torsten},
  booktitle={IEEE/CVF Conference on Computer Vision and Pattern Recognition},
  pages={12786--12798},
  year={2024}
}

@inproceedings{wang2020atloc,
  title={Atloc: Attention guided camera localization},
  author={Wang, Bing and Chen, Changhao and Lu, Chris Xiaoxuan and Zhao, Peijun and Trigoni, Niki and Markham, Andrew},
  booktitle={AAAI Conference on Artificial Intelligence},
  volume={34},
  pages={10393--10401},
  year={2020}
}

@InProceedings{lin2024learning,
    author = {Jingyu Lin and Jiaqi Gu and Bojian Wu and Lubin Fan and Renjie Chen and Ligang Liu and Jieping Ye},
    title = {Learning Neural Volumetric Pose Features for Camera Localization},
    booktitle = {European Conference on Computer Vision},
    year = {2024}
}

@inproceedings{zhao2024pnerfloc,
  title={PNeRFLoc: Visual Localization with Point-Based Neural Radiance Fields},
  author={Zhao, Boming and Yang, Luwei and Mao, Mao and Bao, Hujun and Cui, Zhaopeng},
  booktitle={AAAI Conference on Artificial Intelligence},
  volume={38},
  pages={7450--7459},
  year={2024}
}

@inproceedings{zhou2019continuity,
  title={On the continuity of rotation representations in neural networks},
  author={Zhou, Yi and Barnes, Connelly and Lu, Jingwan and Yang, Jimei and Li, Hao},
  booktitle={IEEE/CVF Conference on Computer Vision and Pattern Recognition},
  pages={5745--5753},
  year={2019}
}

@inproceedings{mildenhall2020nerf,
  title={NeRF: Representing Scenes as Neural Radiance Fields for View Synthesis},
  author={Mildenhall, Ben and Srinivasan, Pratul P and Tancik, Matthew and Barron, Jonathan T and Ramamoorthi, Ravi and Ng, Ren},
  booktitle={European Conference on Computer Vision},
  pages={405--421},
  year={2020},
}

@inproceedings{newcombe2011kinectfusion,
  title={Kinectfusion: Real-time dense surface mapping and tracking},
  author={Newcombe, Richard A and Izadi, Shahram and Hilliges, Otmar and Molyneaux, David and Kim, David and Davison, Andrew J and Kohi, Pushmeet and Shotton, Jamie and Hodges, Steve and Fitzgibbon, Andrew},
  booktitle={2011 10th IEEE international symposium on mixed and augmented reality},
  pages={127--136},
  year={2011},
}

@inproceedings{shavit2022camera,
  title={Camera pose auto-encoders for improving pose regression},
  author={Shavit, Yoli and Keller, Yosi},
  booktitle={European Conference on Computer Vision},
  pages={140--157},
  year={2022},
}

@article{sattler2016efficient,
  title={Efficient \& effective prioritized matching for large-scale image-based localization},
  author={Sattler, Torsten and Leibe, Bastian and Kobbelt, Leif},
  journal={IEEE Transactions on Pattern Analysis and Machine Intelligence},
  volume={39},
  number={9},
  pages={1744--1756},
  year={2016},
}

@inproceedings{liu2017efficient,
  title={Efficient global 2d-3d matching for camera localization in a large-scale 3d map},
  author={Liu, Liu and Li, Hongdong and Dai, Yuchao},
  booktitle={IEEE International Conference on Computer Vision},
  pages={2372--2381},
  year={2017}
}

@inproceedings{deng2009imagenet,
  title={Imagenet: A large-scale hierarchical image database},
  author={Deng, Jia and Dong, Wei and Socher, Richard and Li, Li-Jia and Li, Kai and Fei-Fei, Li},
  booktitle={IEEE/CVF International Conference on Computer Vision and Pattern Recognition},
  pages={248--255},
  year={2009},
}

@inproceedings{mao2017least,
  title={Least squares generative adversarial networks},
  author={Mao, Xudong and Li, Qing and Xie, Haoran and Lau, Raymond YK and Wang, Zhen and Paul Smolley, Stephen},
  booktitle={IEEE/CVF International Conference on Computer Vision and Pattern Recognition},
  pages={2794--2802},
  year={2017}
}

@inproceedings{chessa2023detection,
  title={Detection and Localization of Changes in Immersive Virtual Reality},
  author={Chessa, Manuela and Bassano, Chiara and Solari, Fabio},
  booktitle={International Conference on Image Analysis and Processing},
  pages={121--132},
  year={2023},
}

@inproceedings{Geiger2012KITTI,
  author    = {Andreas Geiger and Philip Lenz and Raquel Urtasun},
  title     = {Are we ready for autonomous driving? The KITTI vision benchmark suite},
  booktitle = {IEEE/CVF International Conference on Computer Vision and Pattern Recognition},
  year      = {2012},
  pages     = {3354--3361},
}

@inproceedings{biswas2012depth,
  title={Depth camera based indoor mobile robot localization and navigation},
  author={Biswas, Joydeep and Veloso, Manuela},
  booktitle={2012 IEEE International Conference on Robotics and Automation},
  pages={1697--1702},
  year={2012},
}

@inproceedings{schonberger2016structure,
  title={Structure-from-motion revisited},
  author={Schonberger, Johannes L and Frahm, Jan-Michael},
  booktitle={IEEE/CVF Conference on Computer Vision and Pattern Recognition},
  pages={4104--4113},
  year={2016}
}

@inproceedings{snavely2006photo,
  title={Photo tourism: exploring photo collections in 3D},
  author={Snavely, Noah and Seitz, Steven M and Szeliski, Richard},
  booktitle={ACM Transactions on Graphics},
  volume={25},
  pages={835--846},
  year={2006},
}

@inproceedings{dosovitskiy2020image,
  title={An Image is Worth 16x16 Words: Transformers for Image Recognition at Scale},
  author={Dosovitskiy, Alexey and Beyer, Lucas and Kolesnikov, Alexander and Weissenborn, Dirk and Zhai, Xiaohua and Unterthiner, Thomas and Dehghani, Mostafa and Minderer, Matthias and Heigold, Georg and Gelly, Sylvain and others},
  booktitle={International Conference on Learning Representations},
  year={2020}
}

@article{oquab2023dinov2,
  title={DINOv2: Learning Robust Visual Features without Supervision},
  author={Oquab, Maxime and Darcet, Timoth{\'e}e and Moutakanni, Th{\'e}o and Vo, Huy and Szafraniec, Marc and Khalidov, Vasil and Fernandez, Pierre and Haziza, Daniel and Massa, Francisco and El-Nouby, Alaaeldin and others},
  journal={Transactions on Machine Learning Research Journal},
  pages={1--31},
  year={2024}
}

@inproceedings{li2024multiagent,
  title={Multiagent Multitraversal Multimodal Self-Driving: Open MARS Dataset},
  author={Li, Yiming and Li, Zhiheng and Chen, Nuo and Gong, Moonjun and Lyu, Zonglin and Wang, Zehong and Jiang, Peili and Feng, Chen},
  booktitle={IEEE/CVF Conference on Computer Vision and Pattern Recognition},
  pages={22041--22051},
  year={2024}
}

@inproceedings{martinbrualla2020nerfw,
  author = {Martin-Brualla, Ricardo
            and Radwan, Noha
            and Sajjadi, Mehdi S. M.
            and Barron, Jonathan T.
            and Dosovitskiy, Alexey
            and Duckworth, Daniel},
  title = {{NeRF in the Wild: Neural Radiance Fields for
           Unconstrained Photo Collections}},
  booktitle = {IEEE/CVF Conference on Computer Vision and Pattern Recognition},
  year={2021}
}

@inproceedings{Chen_Baatz_Koser,   title={City-scale landmark identification on mobile devices},  url={http://dx.doi.org/10.1109/cvpr.2011.5995610},  DOI={10.1109/cvpr.2011.5995610},  booktitle={IEEE/CVF Conference on Computer Vision and Pattern Recognition},  author={Chen, David M. and Baatz, Georges and Koser, Kevin and Tsai, Sam S. and Vedantham, Ramakrishna and Pylvanainen, Timo and Roimela, Kimmo and Chen, Xin and Bach, Jeff and Pollefeys, Marc and Girod, Bernd and Grzeszczuk, Radek},  year={2011},  month={Jun},  language={en-US}  }

@inproceedings{Liu_Li_Dai_2017,   title={Efficient Global 2D-3D Matching for Camera Localization in a Large-Scale 3D Map},  url={http://dx.doi.org/10.1109/iccv.2017.260},  DOI={10.1109/iccv.2017.260},  booktitle={IEEE/CVF Conference on Computer Vision and Pattern Recognition},  author={Liu, Liu and Li, Hongdong and Dai, Yuchao},  year={2017},  month={Oct},  language={en-US}  }

@inproceedings{Martin-Brualla_Radwan_Sajjadi_Barron_Dosovitskiy_Duckworth_2021,   title={NeRF in the Wild: Neural Radiance Fields for Unconstrained Photo Collections},  url={http://dx.doi.org/10.1109/cvpr46437.2021.00713},  DOI={10.1109/cvpr46437.2021.00713},  booktitle={IEEE/CVF Conference on Computer Vision and Pattern Recognition},  author={Martin-Brualla, Ricardo and Radwan, Noha and Sajjadi, Mehdi S. M. and Barron, Jonathan T. and Dosovitskiy, Alexey and Duckworth, Daniel},  year={2021},  month={Jun},  language={en-US}  }

@inproceedings{Lin2024VastGaussianV3,
  title={VastGaussian: Vast 3D Gaussians for Large Scene Reconstruction},
  author={Jiaqi Lin and Zhihao Li and Xiao Tang and Jianzhuang Liu and Shiyong Liu and Jiayue Liu and Yangdi Lu and Xiaofei Wu and Songcen Xu and Youliang Yan and Wenming Yang},
  booktitle={IEEE/CVF Conference on Computer Vision and Pattern Recognition},
  year={2024},
  pages={5166-5175},
}

@inproceedings{dahmani2024swag,
  title={Swag: Splatting in the wild images with appearance-conditioned gaussians},
  author={Dahmani, Hiba and Bennehar, Moussab and Piasco, Nathan and Roldao, Luis and Tsishkou, Dzmitry},
  booktitle={European Conference on Computer Vision},
  pages={325--340},
  year={2025},
}

@inproceedings{zhang2024gaussianwild3dgaussian,
    title={Gaussian in the Wild: 3D Gaussian Splatting for Unconstrained Image Collections}, 
    author={Dongbin Zhang and Chuming Wang and Weitao Wang and Peihao Li and Minghan Qin and Haoqian Wang},
    booktitle={European Conference on Computer Vision},
    year={2024}, 
}

@inproceedings{Chen_deblurgs2024,
   author       = {Wenbo, Chen and Ligang, Liu},
   title        = {Deblur-GS: 3D Gaussian Splatting from Camera Motion Blurred Images},
   booktitle    = {ACM in Computer Graphics and Interactive Techniques},
   year         = {2024},
   volume       = {7},
   number       = {1},
   numpages     = {13},
}

@inproceedings{
  liu2025gscpr,
  title={{GS}-{CPR}: Efficient Camera Pose Refinement via 3D Gaussian Splatting},
  author={Changkun Liu and Shuai Chen and Yash Sanjay Bhalgat and Siyan HU and Ming Cheng and Zirui Wang and Victor Adrian Prisacariu and Tristan Braud},
  booktitle={The Thirteenth International Conference on Learning Representations},
  year={2025}
}

@inproceedings{brahmbhatt2018geometry,
  title={Geometry-aware learning of maps for camera localization},
  author={Brahmbhatt, Samarth and Gu, Jinwei and Kim, Kihwan and Hays, James and Kautz, Jan},
  booktitle={IEEE/CVF Conference on Computer Vision and Pattern Recognition},
  pages={2616--2625},
  year={2018}
}

@inproceedings{Kendall_Cipolla_2017,   title={Geometric Loss Functions for Camera Pose Regression with Deep Learning},  url={http://dx.doi.org/10.1109/cvpr.2017.694},  DOI={10.1109/cvpr.2017.694},  booktitle={2017 IEEE Conference on Computer Vision and Pattern Recognition},  author={Kendall, Alex and Cipolla, Roberto},  year={2017},  month={Jul},  language={en-US}  }

@article{Shavit_Ferens_Keller_2021,   title={Paying Attention to Activation Maps in Camera Pose Regression},  journal={arXiv preprint arXiv:2103.11477},  author={Shavit, Yoli and Ferens, Ron and Keller, Yosi},  year={2021}}

@inproceedings{Chen_Wang_Prisacariu_2021,   title={Direct-PoseNet: Absolute Pose Regression with Photometric Consistency},  url={http://dx.doi.org/10.1109/3dv53792.2021.00125},  DOI={10.1109/3dv53792.2021.00125},  booktitle={2021 International Conference on 3D Vision (3DV)},  author={Chen, Shuai and Wang, Zirui and Prisacariu, Victor},  year={2021},  month={Dec},  language={en-US}  }

@article{ye2024gsplat,
    title={gsplat: An Open-Source Library for {Gaussian} Splatting}, 
    author={Vickie Ye and Ruilong Li and Justin Kerr and Matias Turkulainen and Brent Yi and Zhuoyang Pan and Otto Seiskari and Jianbo Ye and Jeffrey Hu and Matthew Tancik and Angjoo Kanazawa},
    year={2024},
    eprint={2409.06765},
    journal={arXiv preprint arXiv:2409.06765},
}

@inproceedings{li2024memorize,
      title={Memorize What Matters: Emergent Scene Decomposition from Multitraverse}, 
      author={Yiming Li and Zehong Wang and Yue Wang and Zhiding Yu and Zan Gojcic and Marco Pavone and Chen Feng and Jose M. Alvarez},
      booktitle={Advances in Neural Information Processing Systems},
      year={2024}
}

@inproceedings{EKF,
author = {Simon J. Julier and Jeffrey K. Uhlmann},
title = {{New extension of the Kalman filter to nonlinear systems}},
volume = {3068},
booktitle = {Signal Processing, Sensor Fusion, and Target Recognition VI},
pages = {182 -- 193},
year = {1997},
}

@article{goodfellow2014generative,
  title={Generative adversarial nets},
  author={Goodfellow, Ian and Pouget-Abadie, Jean and Mirza, Mehdi and Xu, Bing and Warde-Farley, David and Ozair, Sherjil and Courville, Aaron and Bengio, Yoshua},
  journal={Advances in neural information processing systems},
  volume={27},
  year={2014}
}

@ARTICLE{BRISQUE,
  author={Mittal, Anish and Moorthy, Anush Krishna and Bovik, Alan Conrad},
  journal={IEEE Transactions on Image Processing}, 
  title={No-Reference Image Quality Assessment in the Spatial Domain}, 
  year={2012},
  volume={21},
  number={12},
  pages={4695-4708}}

@inproceedings{bardes2022vicreg,
  title={VICReg: Variance-Invariance-Covariance Regularization For Self-Supervised Learning},
  author={Bardes, Adrien and Ponce, Jean and Lecun, Yann},
  booktitle={International Conference on Learning Representations},
  year={2022}
}
